\PassOptionsToPackage{table,dvipsnames}{xcolor}
\documentclass[11pt,a4paper]{google}
\usepackage[authoryear,round]{natbib}
\usepackage{cafa_layout}
\usepackage{afterpage}
\newcommand{\PaperLogoPath}{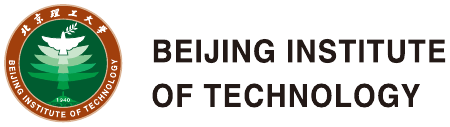}
\newcommand{\PaperLogoHeight}{27pt}
\newcommand{\PaperDate}{2026-11-13}
\newcommand{\PaperCorrespondingEmail}{qiianruii@gmail.com}
\newcommand{\PaperAuthorEmails}{{\{huizhang.ai.2025, qiianruii\}@gmail.com}}
\newcommand{\PaperCodeURL}{https://github.com/huizhang-ai/CAFA}
\newcommand{\PaperCodeLabel}{\expandafter\nolinkurl\expandafter{\PaperCodeURL}}
\newcommand{\PaperModelsURL}{https://huggingface.co/collections/Gen-Verse/open-agentrl}

\newtheorem{proposition}{Proposition}
\newtheorem{corollary}{Corollary}
\newcommand{\E}{\mathbb{E}}
\newcommand{\1}{\mathbf{1}}
\newcommand{\R}{\mathbb{R}}

\newcommand{\Scache}{S_{\mathrm{cache}}}
\newcommand{\Smean}{S_{\mathrm{mean}}}
\newcommand{\Sdraw}{S_{\mathrm{draw}}}
\newcommand{\Qc}{Q_{\mathrm{cache}}}
\newcommand{\Qm}{Q_{\mathrm{mean}}}
\newcommand{\Qd}{Q_{\mathrm{mean+cov}}}
\newcommand{\Fc}{F_{\mathrm{cache}}}
\newcommand{\Fm}{F_{\mathrm{mean}}}
\newcommand{\Fd}{F_{\mathrm{mean+cov}}}
\newcommand{\Tc}{T_{\mathrm{cache}}}
\newcommand{\Td}{T_{\mathrm{mean+cov}}}
\newcommand{\TV}{\mathrm{TV}}

\newcommand{\QwenIndependentInputs}{2728}
\newcommand{\QwenSharedInputs}{2303}
\newcommand{\QwenQueryReduction}{15.6}
\newcommand{\QwenExpectedRatio}{0.8925}
\newcommand{\QwenExpectedLower}{0.8846}
\newcommand{\QwenExpectedUpper}{0.8994}

\newcommand{\HoldoutQwenSaving}{55.8}

\newcommand{\HoldoutSmolSaving}{90.1}

\newcommand{\AutoStageFiveQwenTree}{1059.50}
\newcommand{\AutoStageFiveQwenSaving}{52.5}
\newcommand{\AutoStageFiveQwenIncrement}{7.4}
\newcommand{\AutoStageFiveQwenTreeSaving}{59.0}
\newcommand{\AutoStageFiveQwenTreeRange}{1012--1097}

\newcommand{\AutoHoldoutQwenRange}{55.5--56.2}

\newcommand{\AutoStageFiveSmolTree}{195.50}
\newcommand{\AutoStageFiveSmolSaving}{88.9}
\newcommand{\AutoStageFiveSmolIncrement}{19.1}
\newcommand{\AutoStageFiveSmolTreeSaving}{90.9}
\newcommand{\AutoStageFiveSmolTreeRange}{179--204}

\newcommand{\AutoHoldoutSmolRange}{89.3--91.1}
\newcommand{\AutoSevenShortBefore}{4329}
\newcommand{\AutoSevenShortAfter}{4027}
\newcommand{\AutoSevenShortRemoved}{302}
\newcommand{\AutoSevenShortShared}{4057}
\newcommand{\AutoSevenLongBefore}{4674}
\newcommand{\AutoSevenLongAfter}{4525}
\newcommand{\AutoSevenLongRemoved}{149}
\newcommand{\AutoSevenLongShared}{4455}  
\newcommand{\TotalSavingRange}{7.5--21.4}
\newcommand{\OffpeakSavingRange}{9.2--12.1}
\newcommand{\RevisionEvidence}{77b616340eb3}
\graphicspath{{./}}
\title{Scoring What Matters: Exact Feedback Reduction for Language Model Comparison}
\author[1]{Hui Zhang}
\author[2]{Haoda Yue}
\author[3]{Bicheng Deng}
\author[4]{Yuanchao Zhu}
\author[5]{Jun He}
\author[3 \CorrespondingAuthorSymbol]{Rui Qian}
\affil[1]{Beijing Institute of Technology}
\affil[2]{Nanyang Technological University}
\affil[3]{Fudan University}
\affil[4]{\mbox{University of Electronic Science and Technology of China}}
\affil[5]{\mbox{Beijing Technology and Business~University}}
\correspondingauthor{\href{mailto:\PaperCorrespondingEmail}{\PaperCorrespondingEmail}}

\begin{abstract}
Comparing language-model checkpoints requires estimating performance differences, 
yet evaluation pipelines can pay to score responses whose contributions cancel from 
those differences. We introduce Contrast-Aware Feedback Acquisition (CAFA), a 
sampler-agnostic method that identifies which scores a requested linear report needs 
before acquiring feedback. CAFA groups identical scoring inputs and aggregates their 
report coefficients, removing queries beyond ordinary caching while preserving the 
fully scored result. For a fixed response bank and an otherwise unrestricted scalar scorer, 
we characterize necessary and sufficient query sets for exact recovery of sample means, 
either alone or jointly with ordinary sample covariances. On controlled updates, 
CAFA removes 55.8\% and 90.1\% of the scoring inputs remaining after caching for 
Qwen2.5-1.5B-Instruct and SmolLM2-1.7B-Instruct, respectively, averaged over four 
generation repetitions. Across three online group relative policy optimization (GRPO) 
trajectories, Independent Star sampling with CAFA achieves 46.1--51.3\% lower recorded 
generation-plus-feedback cost than independent autoregressive sampling with caching at 
a common empirical precision threshold (RMS MCSE $\leq 0.04$). This is a post hoc comparison 
of completed runs on a fixed budget grid. All 504 evaluated report cells reproduce 
the fully scored means and covariances to numerical precision. We provide a modular 
implementation supporting existing response banks, versioned caches and external scorers.
\end{abstract}

\begin{document}
\hypersetup{pdftitle={Scoring What Matters: Exact Feedback Reduction for Language Model Comparison},pdfauthor={Hui Zhang, Haoda Yue, Rui Qian, Bicheng Deng, Yuanchao Zhu, and Jun He},pdfsubject={Language model comparison},pdfkeywords={language model comparison, feedback acquisition}}
\enlargethispage{20pt}
\maketitle\par
\begingroup
\setlength{\intextsep}{4pt}
\begin{figure}[H]
\captionsetup{skip=6pt}
\centering\includegraphics[trim=0 10.3bp 0 1.5bp,clip,width=\linewidth]{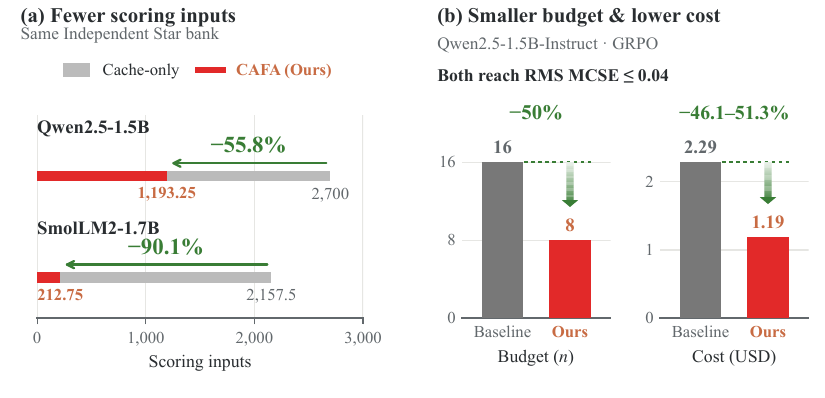}
\caption{Fewer inputs, smaller budget, lower cost. Left: Cache-only versus CAFA (Ours) on identical Independent Star response banks and draw budgets. CAFA preserves the fully scored statistics across four repetitions in Appendix Table~\ref{tab:controlled}. Right: complete workflows for online GRPO with Qwen2.5-1.5B-Instruct. Independent AR + Cache-only (Baseline, $n=16$) and Independent Star + CAFA (Ours, $n=8$) both reach RMS MCSE $\leq 0.04$ across all six matched seed--repetition cells. Costs include generation and feedback per 128-prompt comparison. The panels use separate experiments in Sections~\ref{sec:controlled} and~\ref{sec:grpo-main}.}\label{fig:highlight}
\end{figure}
\endgroup
\clearpage

\begin{figure}[H]
\centering\includegraphics[width=\linewidth]{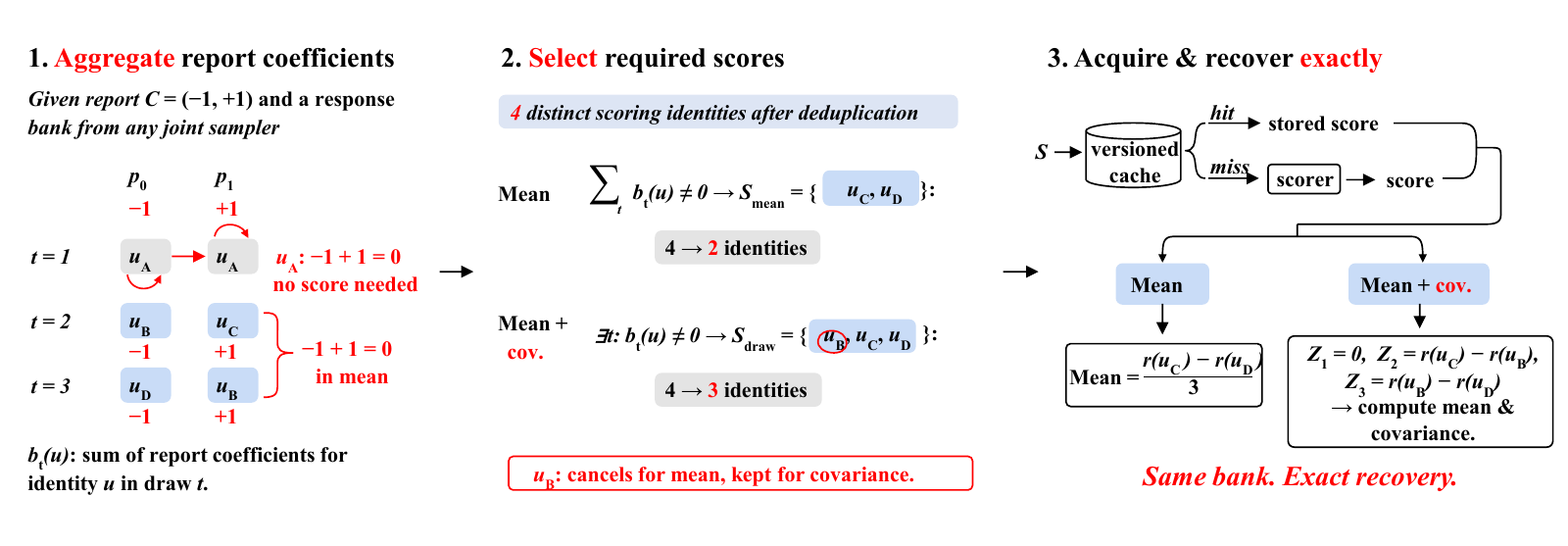}
\caption{CAFA core components on a fixed response bank. The report $C=(-1,+1)$ requests one gain from three equally weighted draws. Six response occurrences reduce to four scoring identities after deduplication. Coefficient aggregation removes $u_A$ within the first draw and cancels $u_B$ across draws in the mean. The mean requires $\{u_C,u_D\}$, while recovery of the mean and covariance requires $\{u_B,u_C,u_D\}$. Required scores are reused from a versioned cache or acquired from the scorer, and the report is recovered with the original draw weights.}\label{fig:overview}\label{fig:mechanism}\label{fig:method}
\end{figure}

\afterpage{\afterpage{%
\begin{figure}[H]
\centering\includegraphics[width=\linewidth]{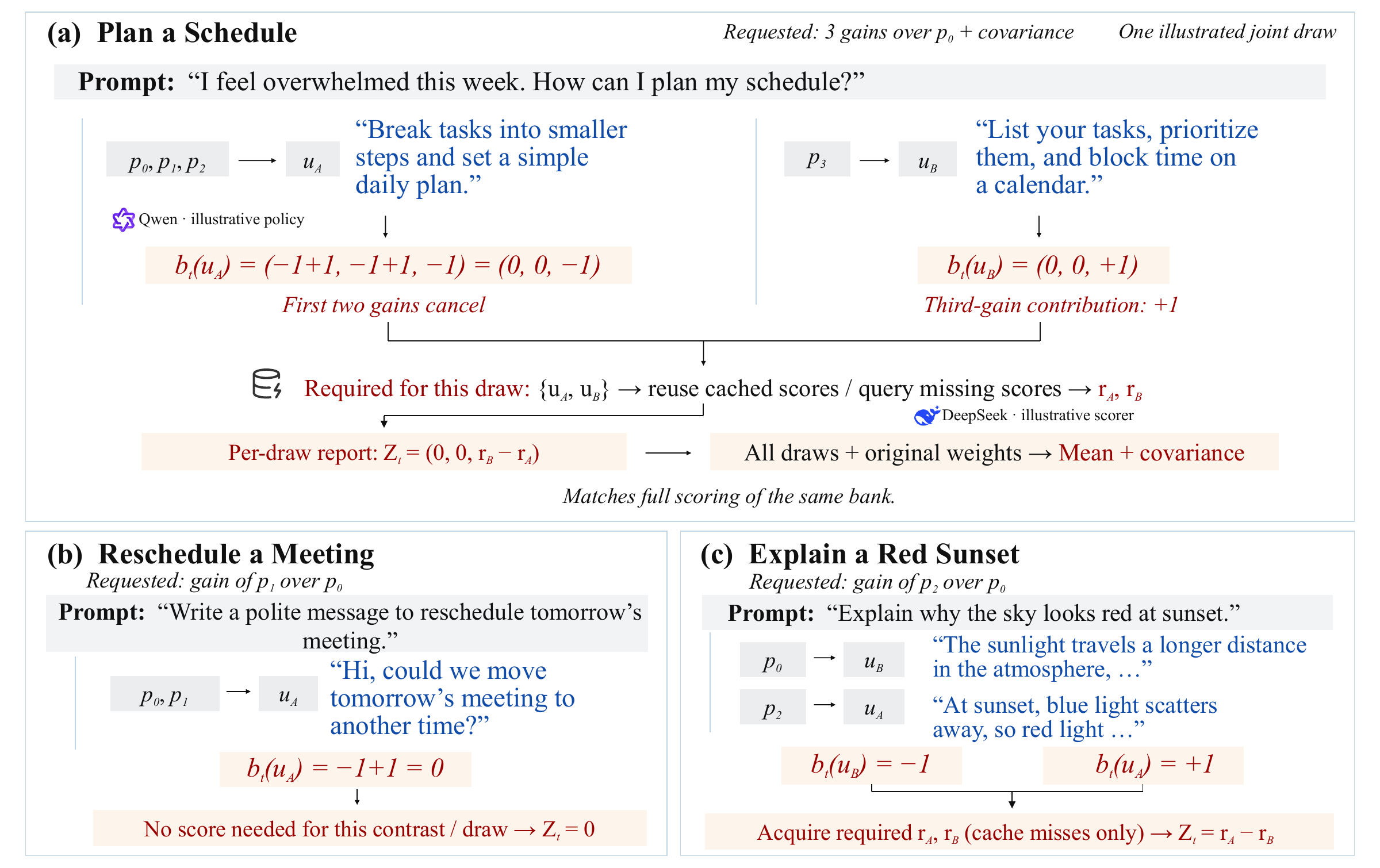}
\caption{CAFA data flow on illustrative prompts. Each panel uses its own prompt and scoring identities, with $r_A=r(u_A)$ and $r_B=r(u_B)$. (a) Three gains over $p_0$ are requested jointly with their covariance. When $p_0,p_1,p_2$ share $u_A$ and $p_3$ produces $u_B$, the draw contributes $Z_t=(0,0,r_B-r_A)$. Both identities are required, and all draws are combined to recover the mean and covariance. (b) Matching responses cancel in a single-gain report, giving $Z_t=0$ without a score query for that draw. (c) Two distinct responses require their scores and contribute $Z_t=r_A-r_B$. In each case, only uncached required scores trigger new calls.}\label{fig:pipeline}
\end{figure}%
}}

\section{Introduction}\label{sec:intro}
Model development requires repeated checkpoint comparisons. Instruction tuning \citep{ouyang2022instruct}, preference optimization \citep{rafailov2023dpo} and reinforcement learning methods, including proximal policy optimization \citep{schulman2017ppo} and group relative policy optimization \citep{shao2024deepseekmath}, produce checkpoints that must be compared to measure progress and select a model. When each response is scored by a paid model-based judge or a human rater, evaluation cost depends on which scores the comparison actually needs.

A report specifies the outputs and statistics to compute. Suppose a base model and three updates produce the same response to a prompt. Scoring the four response occurrences requires four score acquisitions. Caching reduces this to one. The three gains over the base require zero score acquisitions because the common score cancels in every difference. Adding the absolute base score to the report requires that score. The required scores depend on the report and can be identified before scoring. Figure~\ref{fig:overview} illustrates this dependence on a three-draw bank.

Independent generation can produce few exact matches between complete responses. Shared-token autoregressive coupling \citep{benz2026coupled} and tree-coupled feedback sharing \citep{wen2026tcab} increase sharing while preserving each checkpoint's output distribution. Contrast-Aware Feedback Acquisition (CAFA) then aggregates coefficients over identical scoring inputs to recover the requested means and covariances with only the necessary scores. Throughout, \emph{Cache-only} denotes scoring after cache deduplication. \emph{CAFA (Ours)} denotes the proposed scoring strategy, regardless of sampler. Independent autoregressive (AR), shared-token AR, Independent Star and total-variation (TV) tree denote sampling methods. Figure~\ref{fig:highlight}a shows the scoring inputs removed by CAFA on identical Independent Star banks. Figure~\ref{fig:highlight}b compares complete workflows at a common empirical precision threshold, including both generation and feedback cost.

CAFA exactly recovers the sample means obtained by fully scoring a fixed bank and, when requested, the ordinary sample covariances of its per-draw report vectors. For an otherwise unrestricted scorer, every unknown score in the corresponding support is necessary. A response can cancel from the mean while affecting covariance, so the requested statistics are part of the report. We evaluate the method on fixed controlled updates to isolate reductions in score acquisitions on identical samples, and on online group relative policy optimization (GRPO) trajectories to assess the cost and statistical precision of checkpoint comparisons, as illustrated in Figure~\ref{fig:highlight}b. We make the following contributions:

\begingroup
\noindent(i) We introduce CAFA, which aggregates coefficients over exact scoring identities for any joint sampler and supports both mean-only and mean-with-covariance reports, including general linear reports with absolute anchors.

\noindent(ii) We establish exact recovery and query necessity for both modes, show that invertible coordinate changes preserve the required support and that adding outputs takes the union of supports, and characterize when immediate CAFA attains the batch support.

\noindent(iii) We separate reductions attributable to CAFA on fixed banks from workflow improvements due to coupling and CAFA. Controlled updates measure feedback savings beyond caching. Online GRPO trajectories measure cost at a common observed precision threshold. All 504 report cells reproduce the fully scored statistics to numerical precision.

\noindent(iv) We provide an installable implementation with interfaces for existing response banks, external scorers and Independent Star generation from local checkpoints, as detailed in Appendix~\ref{app:software}.
\par\endgroup

\FloatBarrier
\section{Problem Setup}\label{sec:setup}

\paragraph{Policies, Prompts and Scores.}
Fix a base language-model policy $p_0$ and updated policies $p_1,\ldots,p_J$. Each policy is defined by its decoding settings and stopping rule, with any length cap treated as part of that rule. Distinct prompts $x_i$, $i=1,\ldots,P$, carry positive weights $w_i$ summing to one ($w_i=1/P$ in all our analyses). A scorer $r$ maps a prompt and a response to a real number. The scorer is fixed or versioned, with its identity defined by the rubric, judge deployment and settings. For a stochastic judge, we store and reuse the first score for each input in a versioned cache. All recovery and MCSE statements condition on this fixed scoring function. They exclude variability from repeated judge calls. Write $\mu_j(x_i)=\E_{Y\sim p_j(\cdot\mid x_i)}[r(x_i,Y)]$ and $\mu(x_i)=(\mu_0(x_i),\ldots,\mu_J(x_i))^\top$. Population means and MCSEs assume finite second moments, as guaranteed by the bounded experimental scores. Fixed-bank recovery does not require boundedness.

\Needspace{4\baselineskip}

\paragraph{The Requested Report.}
A comparison is a linear report
\begin{equation}\label{eq:target}
 \theta=\sum_i w_i\,C\,\mu(x_i),\qquad C\in\mathbb{R}^{d\times(J+1)} .
\end{equation}
Each row of $C$ defines one requested output. Column $j$ contains the coefficients applied to the mean score of model $j$.
The main case is a contrast, $C\1=0$. For instance $C=[-\1,I_J]$ requests the $J$ gains over the base, and a single row $e_j^\top-e_0^\top$ requests one of them. Appending $e_0^\top$ as an additional row adds the absolute mean of the base model. The resulting anchored report is general linear, with a nonzero row sum. The report also specifies the requested statistics: either the mean alone or the mean together with the ordinary sample covariance of the per-draw report vectors. The latter requires at least two draws per prompt and is used to estimate the Monte Carlo covariance of the reported mean.

\paragraph{Joint Samples and Two Identities.}
At prompt $x_i$ an evaluator draws $n_i$ joint samples $Y_{i,t}=(Y_{i,t,0},\ldots,Y_{i,t,J})$ whose coordinate $j$ has law $p_j(\cdot\mid x_i)$. The coordinates may be dependent. Write $\boldsymbol r(Y_{i,t})=(r(x_i,Y_{i,t,0}),\ldots,r(x_i,Y_{i,t,J}))^\top$. The fully scored estimator is $\widehat\theta=\sum_i w_i\,n_i^{-1}\sum_t C\,\boldsymbol r(Y_{i,t})$. The scoring identity $\kappa(x_i,y)$ is determined by prompt $x_i$ and the exact decoded text of response $y$. Responses with the same scoring identity share a score. Probability identity instead uses the original token sequence under the stated policy. Sequences with identical decoded text can share a score while remaining distinct probability events. Neither identity uses text normalization, truncation or semantic~matching.

\Needspace{4\baselineskip}

\paragraph{What Is Counted.}
We distinguish three quantities related to evaluation cost. The required support contains the distinct scoring identities needed for a report, determined from the bank before observing scores. New paid calls count uncached required scores acquired in an execution. Recorded cost sums generation and feedback costs at the stated tariffs. The goal is to obtain the scores on the required support and recover the same sample statistics as full scoring of the bank.
\section{Contrast-Aware Feedback Acquisition (CAFA)}\label{sec:method}

Given a report with coefficient matrix $C$ that requests means alone or means with covariance, and a joint response bank, CAFA groups identical scoring inputs, sums their coefficients, selects the required support, acquires its uncached scores and recovers the report with the original draw weights. Figure~\ref{fig:overview} shows these core operations. Figure~\ref{fig:pipeline} traces the data flow for a joint three-gain report and two single-gain reports.

\subsection{Coupled Generation}\label{sec:coupling}
CAFA accepts any joint sampler with the specified marginals, including independent sampling. We illustrate coupled generation with \emph{Independent Star}. For each joint draw, sample a root $X\sim p_0$. For each update $j$, draw an independent uniform $U_j$ and copy the root, $Y_j=X$, if $U_j\le\min\{1,p_j(X)/p_0(X)\}$. Otherwise draw $Y_j$ from the positive residual $\rho_j\propto(p_j-p_0)_+$. The copied mass $\min(p_0,p_j)$ and residual mass $(p_j-p_0)_+$ sum to $p_j$, giving the required marginal. Since $\Pr(Y_j\ne X)=\TV(p_0,p_j)$, $(X,Y_j)$ is a maximal coupling of $p_0$ and $p_j$ at the token-sequence level. Simultaneous agreement across all updates is a separate property. Sequence probabilities are evaluated on complete original token sequences under the sampling temperature and stopping rule. Every residual proposal is included in generation cost. A joint draw that exhausts its residual-proposal budget without acceptance is recorded as incomplete. Appendix~\ref{app:B-samplers} specifies acceptance probabilities, random streams and the baseline samplers.

\subsection{Cancel before Scoring}\label{sec:cancel}
For each prompt $x_i$, draw $t$ and scoring identity $u$, sum the report columns of the policies whose responses have identity $u$:
\begin{equation}\label{eq:coefficients}
 b_{i,t,u}=\sum_{\makebox[0pt]{$\scriptstyle j:\,\kappa(x_i,Y_{i,t,j})=u$}}C_{:j},\qquad
 A_{i,u}=\sum_{t=1}^{n_i}b_{i,t,u}.
\end{equation}
Here $b_{i,t,u}$ is the net coefficient vector for identity $u$ within one draw, and $A_{i,u}$ is its sum over all draws for prompt $x_i$.
Only columns with a non-zero entry in $C$ are active, so a single-gain report excludes responses from the other updates. Then
\begin{equation}\label{eq:cancel}
 \bar Z_i=\frac{1}{n_i}\sum_{u}A_{i,u}\,r(u),\qquad \widehat\theta=\sum_i w_i\bar Z_i ,
\end{equation}
which is the fully scored estimator with terms grouped by scoring identity. Coefficients are aggregated exactly before division by $n_i$, and a coefficient vector is treated as zero only when all $d$ components are zero. The identities that remain are
\begin{equation}\label{eq:scoring-supports}
 \Smean=\{(i,u):A_{i,u}\ne0\}\ \subseteq\ \Sdraw=\{(i,u):\exists t,\ b_{i,t,u}\ne0\}\ \subseteq\ \Scache ,
\end{equation}
where $\Scache$ collects every identity observed on an active column. We write $\Qc=|\Scache|$, $\Qm=|\Smean|$ and $\Qd=|\Sdraw|$ for the corresponding numbers of required scoring inputs. The operation removes score queries, never draws: a draw in which all policies agree still counts towards $n_i$.

\subsection{Recover the Requested Report}\label{sec:recover}
Recovering the mean requires the scores on $\Smean$. Reporting uncertainty may require additional scores: with $Z_{i,t}=\sum_u b_{i,t,u}r(u)$, the ordinary sample covariance
$\widehat\Sigma_i=(n_i-1)^{-1}\sum_t(Z_{i,t}-\bar Z_i)(Z_{i,t}-\bar Z_i)^\top$
and the Monte Carlo covariance of the weighted mean, $\widehat V=\sum_i w_i^2\widehat\Sigma_i/n_i$, are recovered from the per-draw vectors determined by the scores on $\Sdraw$. For fixed-prompt MCSEs, joint draws are independent and identically distributed within each prompt, and generation streams are independent across distinct prompts. Thus $\widehat V$ estimates the generation covariance conditional on the prompts and fixed scoring function. The algebraic recovery identities themselves do not require independence. For example, two mirror draws for the same prompt return $(A,B)$ and $(B,A)$, with the base listed first. Their differences, $r(B)-r(A)$ and $r(A)-r(B)$, cancel, so the sample mean is zero without either score. The ordinary sample variance is $2[r(B)-r(A)]^2$, however, and still requires both scores for unrestricted unknown $r(A)$ and $r(B)$, as shown in Appendix Figure~\ref{fig:mirror-draws}.

Algorithm~\ref{alg:acquire} summarizes the fixed-bank procedure. Coefficient aggregation determines the required inputs before acquisition, and the selected mode determines whether per-draw vectors are reconstructed. Multi-output reports are planned jointly, and changed reports or budgets are re-planned with cached scores. Appendix~\ref{app:B-algorithm} gives the complete procedure in Algorithm~\ref{alg:acquire-details}, its missing-score handling and the windowed variant.

\begin{algorithm}[!htbp]
\caption{CAFA for a fixed bank.}\label{alg:acquire}
\begin{algorithmic}[1]
\Require response bank $\{Y_{i,t,j}\}$, report matrix $C$, mode $s\in\{\text{mean},\text{mean+cov}\}$, scorer $r$ and versioned cache $\mathcal K_v$; $n_i\ge2$ for mean+cov
\State Aggregate $b_{i,t,u}\gets\sum_{j:\kappa(x_i,Y_{i,t,j})=u}C_{:j}$ over active columns.
\State Sum $A_{i,u}\gets\sum_{t=1}^{n_i}b_{i,t,u}$ for each prompt and identity.
\State Form $\Smean\gets\{(i,u):A_{i,u}\ne0\}$ and $\Sdraw\gets\{(i,u):\exists t,\ b_{i,t,u}\ne0\}$.
\State Select $S\gets\Smean$ for mean reporting, or $S\gets\Sdraw$ for mean+cov.
\State Identify $M\gets\{(i,u)\in S:u\notin\operatorname{dom}(\mathcal K_v)\}$.
\State Acquire scores for $M$, apply the early-return rules in Algorithm~\ref{alg:acquire-details}, and update $\mathcal K_v$.
\State Recover $\bar Z_i\gets n_i^{-1}\sum_{u:A_{i,u}\ne0}A_{i,u}\mathcal K_v[u]$.
\State Combine $\widehat\theta\gets\sum_i w_i\bar Z_i$ with the original prompt weights.
\State For mean+cov, reconstruct every $Z_{i,t}$ and compute $\widehat\Sigma_i,\widehat V$ as in Section~\ref{sec:recover}.
\State \Return $\widehat\theta$, requested covariances, $|S|$ and the new paid-call count.
\end{algorithmic}
\end{algorithm}

\subsection{Implementation}\label{sec:interface}
Our library supports existing banks, local Independent Star generation, versioned caches and external scorers. It plans support before reading scores. The scorer sees only the prompt and response. Appendix~\ref{app:software} documents export--import, returned statistics and query logs, with runnable examples and tests.
\section{Exact Recovery and Required Feedback}\label{sec:theory}

The guarantees condition on a fixed bank's responses, scoring identities and report. The scorer is an otherwise unrestricted oracle.

\subsection{Exact Recovery and Required Mean Feedback}\label{sec:theory-mean}
\begin{proposition}[Pointwise recovery and necessary mean queries]\label{prop:mean}\label{prop:cancel}
For every bank and every scorer, the estimate~\eqref{eq:cancel} computed from $\Smean$ equals the fully scored estimate $\widehat\theta$. Conversely, suppose each unknown score can take any value in a nondegenerate interval, independently of the others. Then any procedure that returns $\widehat\theta$ exactly for every such scorer must query every identity in $\Smean$ whose score it does not already hold.
\end{proposition}
Sufficiency follows by collecting terms with the same scoring identity. For necessity, suppose an initially unknown score at $(i,u)\in\Smean$ remains unqueried. Two scorers that agree on every queried input and differ at $u$ produce the same transcript but reports differing by $w_i n_i^{-1}A_{i,u}\,(r(u)-r'(u))\ne0$, as detailed in Appendix~\ref{app:A-mean}.

\subsection{Joint Mean and Covariance Reporting}\label{sec:theory-cov}\label{sec:reporting-support}
\begin{proposition}[Required support for the mean with ordinary covariance]\label{prop:cov}\label{prop:report-support}
Condition on the bank, report and identities, and let $n_i\ge2$. Suppose each unknown score can vary independently over a nondegenerate interval. The scores on $\Sdraw$ suffice for exact recovery of $\bar Z_i$ and $\widehat\Sigma_i$. Every score not already held in this support must be queried to guarantee recovery for every such scorer. The same holds for $\widehat\theta$ with the stratified covariance $\widehat V$ over distinct prompt identities with positive weights.
\end{proposition}
The scores on $\Sdraw$ determine every $Z_{i,t}$, establishing sufficiency. For necessity, an unknown score with $A_{i,u}\ne0$ changes the mean. When $A_{i,u}=0$ but some draw coefficient is nonzero, fix the other scores and write $Z_{i,t,k}=b_{i,t,u,k}s+v_{i,t,k}$, where $s=r(u)$ and $v_{i,t,k}$ collects contributions from the other identities. Define $\bar b_{i,u,k}=n_i^{-1}\sum_t b_{i,t,u,k}$ and $\bar v_{i,k}=n_i^{-1}\sum_t v_{i,t,k}$. The sample-covariance diagonal is
\begin{equation}\label{eq:cov-score-dependence}
 \widehat\Sigma_{i,kk}(s)=\frac{1}{n_i-1}\sum_t
 \bigl[(b_{i,t,u,k}-\bar b_{i,u,k})s+v_{i,t,k}-\bar v_{i,k}\bigr]^2.
\end{equation}

For an output $k$ with a nonzero draw coefficient, $\bar b_{i,u,k}=0$ and the squared-term coefficient is $\sum_t b_{i,t,u,k}^2/(n_i-1)>0$. Thus this score affects the covariance even though it cancels from the mean. Positive prompt weights preserve this dependence in $\widehat V_{kk}$, as proved in Appendix~\ref{app:A-cov}. Hence $|\Sdraw\setminus\Smean|$ counts the additional scoring identities required for joint mean-and-covariance reporting. The result does not characterize the minimum number of queries needed to recover covariance alone, which may require fewer scores when an unknown contribution is constant across draws.

\subsection{Operational Consequences}\label{sec:theory-ops}\label{sec:output-support}
Fix a reporting mode (mean alone or mean with covariance), and write $S(C)$ for its required support on a fixed bank.
\begin{corollary}[Changing coordinates versus changing the report]\label{cor:report}\label{cor:report-support}
If $H$ is invertible, $S(HC)=S(C)$. Stacking reports takes the union, $S\bigl([C;D]\bigr)=S(C)\cup S(D)$. In particular, reporting all gains over the base together with its absolute mean requires every observed identity: $S=\Scache$.
\end{corollary}
An occurrence vector counts the appearances of a scoring identity under each policy. Invertible maps preserve zero vectors, and a stacked vector vanishes only if both blocks do. $[-\1,I_J]$ stacked with $e_0^\top$ is invertible, so no nonzero occurrence vector is mapped to zero. Both claims hold for $\Smean$ and separately within each draw for $\Sdraw$. Changing the outputs requires recomputing their joint support and missing-score cost, which need not equal $d$ times single-output cost because supports can overlap.

Under Independent Star with base-relative rows, a token sequence $z$ receives a negative coefficient for update $j$ only when it appears as a rejected root ($p_j(z)<p_0(z)$), and a positive coefficient only when it appears as a residual draw ($p_j(z)>p_0(z)$). Its nonzero per-draw coefficients cannot cancel across draws. With identities defined by original token sequences, $\Smean=\Sdraw$: immediate CAFA attains the batch required-input count, and the same scores recover the covariance, as established in Proposition~\ref{prop:signpure} in Appendix~\ref{app:A-immediate}. For decoded-text identities, this conclusion requires $p_j-p_0$ to have a consistent sign within each decoded group. The appendix gives a counterexample when this condition fails. Other samplers can require the covariance increment, as the experiments show.
\section{Experiments}\label{sec:experiments}

\subsection{Setup}\label{sec:exp-setup}
To assess CAFA's practical benefits and the factors that determine them, we ask three questions:
(RQ1) How much feedback does CAFA remove beyond caching? (RQ2) How does coupled generation with CAFA affect cost and precision? (RQ3) How do report choice and checkpoint position affect savings?
We compare controlled updates and online GRPO using the regimes in Table~\ref{tab:regimes} and the protocol in Appendix~\ref{app:protocol}. Controlled updates use three fixed endpoints per model, one per training seed, for Qwen2.5-1.5B-Instruct and SmolLM2-1.7B-Instruct. Online GRPO uses three Qwen2.5-1.5B-Instruct trajectories, comparing the base with steps 10, 50 and 250. Offline checkpoints saved at steps 2, 8 and 32 are evaluated in Section~\ref{sec:configs}. Evaluation prompts come from No Robots \citep{h4norobots}, WildChat \citep{zhao2024wildchat} and ELI5 \citep{fan2019eli5}. Policies sample at temperature 1 with a 768-token cap. Sequence probabilities are evaluated on the original token IDs. Judges return scalar scores from 1 to 10, reused from versioned caches: DeepSeek Flash for online GRPO evaluation, and DeepSeek Flash and Qwen Flash separately for controlled updates. Judge settings and sensitivity are in Appendices~\ref{app:B-scoring} and~\ref{app:F-judges}.

\Needspace{4\baselineskip}
The controlled strict holdout has 32 prompts and 64 joint draws per prompt. Figure~\ref{fig:controlled} uses two repetitions per sampler, with all four Independent Star repetitions in Appendix~\ref{app:controlled}. GRPO uses 128 held-out prompts, split equally for selection and testing. Independent AR, shared-token AR, Independent Star and TV tree are evaluated at 4, 8 and 16 draws per prompt, across three seeds and two generation repetitions: 72 banks and 504 cells over seven reports.

For each cell, fixed-prompt precision is summarized by $\mathrm{RMS\ MCSE}=\sqrt{d^{-1}\sum_{k=1}^{d}\widehat V_{kk}}$, with $d=3$ for the joint-gain report. This is the root mean square of the individual gains' MCSEs. Costs in USD price generation from uninterrupted runtime measurements and feedback from measured token usage at frozen tariffs, as detailed in Appendix~\ref{app:cost}. Fixed-bank comparisons hold the response bank, score values and full recorded generation cost fixed to compare Cache-only with CAFA. Workflow comparisons also change the sampler, relative to Independent AR + Cache-only.

\begin{figure}[!t]
\centering\includegraphics[trim=7.38bp 12.8bp 1.5bp 3.18bp,clip,width=0.938\linewidth]{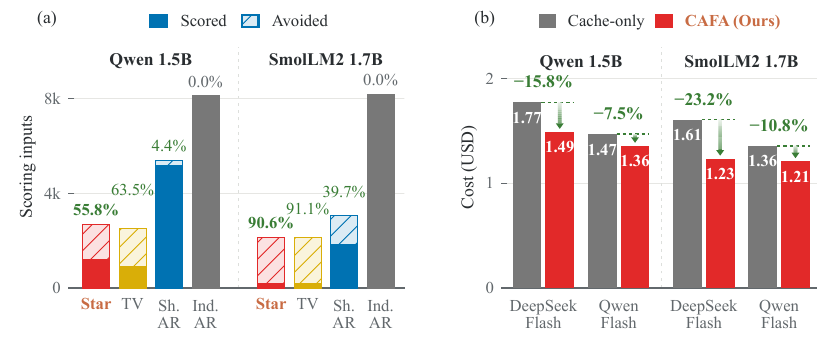}
\caption{CAFA reduces scoring inputs and same-bank cost. Controlled strict holdout: 32 prompts, three updates and 64 joint draws per prompt, means over the first two repetitions for every sampler. (a) Full bars count deduplicated inputs. Solid portions are required for means and covariance, and hatched portions are avoided. (b) Independent Star generation-plus-feedback cost in USD, comparing Cache-only and CAFA on the same banks with the full generation charge fixed. Sh. AR, Ind. AR and TV denote shared-token AR, independent AR and TV tree. All four Independent Star repetitions appear in Appendix Table~\ref{tab:controlled}.}\label{fig:controlled}
\end{figure}
\subsection{Controlled Updates: Isolating Feedback Reduction}\label{sec:controlled}
\textbf{CAFA removes substantial feedback beyond caching while preserving the sampled reports exactly.} On the first two Independent Star repetitions for each model, it eliminates 55.8\% of the scoring inputs required after deduplication for Qwen and 90.6\% for SmolLM2, as shown in Figure~\ref{fig:controlled}a. The same rule removes 63.5\% and 91.1\% on TV tree banks, but 4.4\% and 39.7\% on shared-token AR banks, respectively, and essentially none on independent AR banks. Across samplers, savings depend on the pattern of identical responses in each bank and the requested report. Mean and mean-plus-covariance supports coincide in all four Independent Star repetitions. For Qwen with independent AR, the mean-only support averages 8132.5 scoring inputs, compared with 8133 for the mean-with-covariance report, as reported in Table~\ref{tab:controlled} in Appendix~\ref{app:controlled}.

\textbf{Avoiding score acquisitions also lowers total recorded cost.} With the complete joint-star generation cost held fixed, Figure~\ref{fig:controlled}b compares Cache-only and CAFA total costs, yielding 7.5--23.2\% lower generation-plus-feedback cost across the two models and two judges on the first two repetitions. Savings increase with the feedback share.

\begin{figure}[!t]
\centering\includegraphics[trim=21.69bp 19.29bp 7.48bp 9.61bp,clip,width=0.680\linewidth]{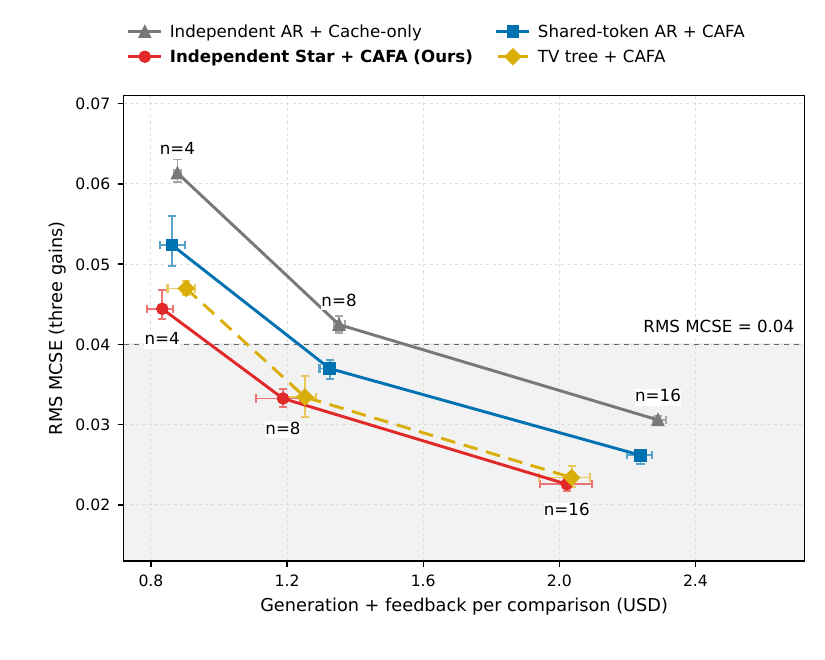}
\caption{Online GRPO cost--precision tradeoffs. Points average total cost and cell-level RMS MCSE over three training seeds $\times$ two generation repetitions. Whiskers span the six-cell minima and maxima. Budgets $n=4,8,16$ increase from left to right along each curve. Cost includes generation and feedback per 128-prompt, three-gain comparison with covariance, including the TV-tree pilot. The dashed line marks RMS MCSE $=0.04$. All 72 banks appear individually in Appendix Figure~\ref{fig:grpo-cost}.}\label{fig:grpo-cost-main}
\end{figure}

\subsection{Online GRPO: Practical Model Comparison}\label{sec:grpo-main}
\textbf{The coupled workflow reduces recorded cost while improving precision.} Independent test references, using 64 autoregressive draws per policy and test prompt, estimate final step-250 gains of 0.317--0.350 judge-score points over the base across the three training seeds, as reported in Appendix Table~\ref{tab:review3-reference}. Across all 18 equal-budget comparisons, Independent Star + CAFA has lower recorded cost and lower RMS MCSE than Independent AR + Cache-only, as shown in Figure~\ref{fig:grpo-cost-main}. At 16 draws per prompt, the workflow in Table~\ref{tab:workflow} uses 31.1--36.9\% fewer scoring inputs, costs 9.4--14.6\% less and has 0.716--0.772 times the baseline RMS MCSE. TV tree and shared-token AR lower RMS MCSE in all 18 comparisons but exceed baseline cost in five and two four-draw cells, respectively. TV-tree cost includes its pilot stage. At 16 draws per prompt, Independent Star's reduction in feedback cost exceeds its increase in recorded generation cost in all six cells. The per-cell breakdown and CAFA bookkeeping timings are in Appendix Table~\ref{tab:grpo-cells} and Appendix~\ref{app:G-bookkeeping}.

\textbf{The precision advantage lowers the required budget on the evaluated grid.} At the post hoc threshold RMS MCSE $\le 0.04$, the smallest eligible budget is eight draws per prompt for Independent Star + CAFA and sixteen for Independent AR + Cache-only in all six seed--repetition cells. Comparing these budgets gives 46.1--51.3\% lower recorded generation-plus-feedback cost, as shown in Figure~\ref{fig:highlight}. This is a descriptive comparison of completed fixed-budget runs. The empirical MCSE threshold does not certify population error or interval coverage. Appendices~\ref{app:F-proxy}--\ref{app:F-rare} examine error and coverage in finite-proxy and synthetic settings.
\par

\begin{table}[!t]\centering\small\setlength{\tabcolsep}{7.55618pt}
\caption{Online GRPO workflows at 16 draws per prompt. Each workflow reports three gains and their covariance. The ``Scored'' column reports input counts: $\Qc$ for Cache-only and $\Qd=|\Sdraw|$ for CAFA. Counts and generation-plus-feedback costs per 128-prompt comparison are means over three training seeds $\times$ two generation repetitions. Reduction and RMS MCSE-ratio ranges are the six-cell minima and maxima relative to Independent AR + Cache-only in each cell.}\label{tab:workflow}\label{tab:grpo-decision}
\begin{adjustbox}{max width=\linewidth}
\begin{tabular*}{\linewidth}{@{\hspace{\tabcolsep}\extracolsep{\fill}}llrrrrr}\toprule
& & \multicolumn{2}{c}{Recorded workload $\downarrow$} & \multicolumn{2}{c}{Reduction $\uparrow$} & Precision $\downarrow$\\
Sampler & Scoring strategy & Scored & Cost (USD) & Inputs (\%) & Cost (\%) & MCSE ratio\\\midrule
Independent AR & Cache-only & 8,170 & 2.29 & -- & -- & 1\\
Shared-token AR & CAFA (Ours) & 7,853 & 2.24 & 2.2--6.5 & 1.2--3.3 & 0.826--0.875\\
\rowcolor{PaperAccentLight} Independent Star & CAFA (Ours) & 5,424 & \textbf{2.02} & 31.1--36.9 & 9.4--14.6 & 0.716--0.772\\
TV tree & CAFA (Ours) & \textbf{4,950} & 2.04 & 36.7--42.5 & 8.5--14.7 & 0.733--0.823\\
\bottomrule\end{tabular*}
\end{adjustbox}
\end{table}

\textbf{CAFA adds a report-dependent cost reduction.} On fixed Independent Star banks with full four-policy generation cost retained, CAFA lowers recorded cost by 20.9--27.0\% for the step-10 gain over the base and 1.54--2.07\% for the joint three-gain report, as reported in Table~\ref{tab:targets}.

\begingroup\interlinepenalty=0
\textbf{CAFA preserves all 504 sampled reports.} An independent recomputation finds zero mean differences and a maximum absolute covariance difference of $4.3\times10^{-19}$ from full scoring, with one cached-score access per required identity, as detailed in Appendix~\ref{app:D-preservation}. Mean-only and mean-plus-covariance supports coincide in 485 cells, while 19 autoregressive-sampler cells need additional scores for covariance, as established in Proposition~\ref{prop:cov}. All 72 fixed checkpoint selections match full scoring of the same banks. 71 of 72 also match the independent 64-draw test-reference best checkpoint. The remaining selection has estimated regret 0.151 with MCSE 0.020, as detailed in Appendix~\ref{app:D-selection}.
\par\endgroup

\subsection{What Determines the Feedback Requirement?}\label{sec:configs}
\begin{figure}[!t]
\centering\includegraphics[trim=6.26bp 9.72bp 7.48bp 3.92bp,clip,width=0.850\linewidth]{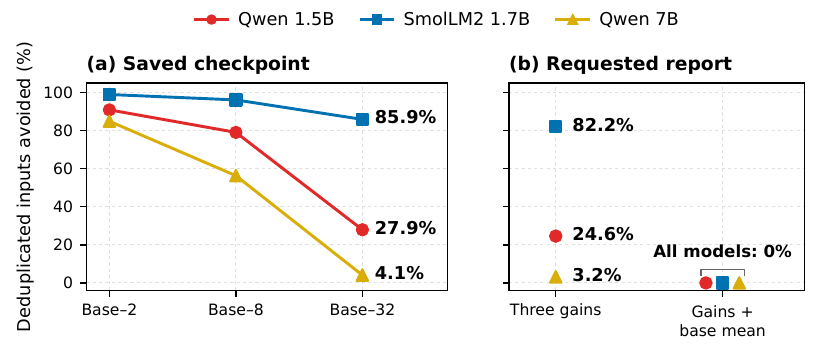}
\caption{Feedback savings vary with checkpoints and requested outputs. Independent Star + CAFA on offline checkpoints: 32 prompts, 64 draws per prompt and a 768-token cap. The vertical axis is the fraction of deduplicated inputs avoided. Left: individual base-relative gains at steps 2, 8 and 32. Right: all three gains, with and without the base mean. Points average three trajectories per small model and one for Qwen 7B.}\label{fig:configs}
\end{figure}

\paragraph{Joint Reports Combine Required Supports.}
Table~\ref{tab:targets} compares seven reports on six fixed Independent Star banks. Joint reports require the union of constituent supports, as established in Corollary~\ref{cor:report}. Scores can cancel in one output but be needed for another. Gains plus the base mean require full Cache-only support. Selection uses gains alone because subtracting a common base mean preserves the maximizer.

\begin{table}[!htbp]\centering\small\setlength{\tabcolsep}{5pt}
\caption{Same banks, different reports. Online GRPO, Independent Star, 16 draws per prompt. Counts are means over three seeds $\times$ two generation repetitions, and percentages are cellwise ranges. $\Qc$ counts deduplicated inputs. $\Qm$ and $\Qd$ count scores needed for means and means with covariance, respectively, and coincide on these banks. ``Removed'' is $1-\Qd/\Qc$. Cost savings compare CAFA with Cache-only while retaining the full four-policy generation charge.}\label{tab:targets}
\begin{adjustbox}{max width=\linewidth}
\begin{tabular*}{\linewidth}{@{\hspace{\tabcolsep}\extracolsep{\fill}}lrrrr}\toprule
Report & $\Qc$ & $\Qm{=}\Qd$ & Removed (\%) & Cost saved (\%)\\\midrule
\rowcolor{PaperHeader}\multicolumn{5}{l}{\textit{Single gains}}\\
\rowcolor{white}Base--10 & 2,413.3 & 747.0 & 59.7--79.6 & 20.86--26.96\\
\rowcolor{white}Base--50 & 3,431.8 & 2,777.5 & 14.5--23.3 & 6.96--10.68\\
\rowcolor{white}Base--250 & 3,844.0 & 3,601.7 & 5.6--6.8 & 2.49--2.92\\
\addlinespace[2pt]
\rowcolor{PaperHeader}\multicolumn{5}{l}{\textit{Adjacent gains}}\\
\rowcolor{white}10--50 & 3,449.8 & 2,809.0 & 14.2--22.6 & 6.79--10.35\\
\rowcolor{white}50--250 & 3,906.7 & 3,717.7 & 4.3--5.3 & 1.87--2.40\\
\addlinespace[2pt]
\rowcolor{PaperHeader}\multicolumn{5}{l}{\textit{Joint reports}}\\
\rowcolor{white}Three gains & 5,611.5 & 5,424.3 & 2.9--3.8 & 1.54--2.07\\
\rowcolor{white}\textbf{Gains + base mean} & 5,611.5 & 5,611.5 & \textbf{0.0} & \textbf{0.00}\\
\bottomrule\end{tabular*}
\end{adjustbox}
\end{table}

\Needspace{6\baselineskip}
\begingroup
\setlength{\parskip}{3pt plus1pt minus1pt}
\titlespacing*{\paragraph}{0pt}{3pt plus1pt minus1pt}{0.75em}
\titlespacing*{\section}{0pt}{2ex plus2pt minus2pt}{4pt}
\paragraph{Earlier Checkpoints Save More Feedback in the Offline Trajectory.}
Figure~\ref{fig:configs} compares feedback reduction across offline checkpoints with a 768-token cap. On the single Qwen2.5-7B trajectory, reductions in scoring inputs are 84.9\%, 56.3\% and 4.1\% at steps 2, 8 and 32, respectively, and 3.2\% for the joint report. Steps label checkpoints rather than measured distribution distances. Output-cap and temperature effects are in Appendices~\ref{app:E-configs}--\ref{app:E-temperature}.
\par

\paragraph{Higher Feedback Cost Shares Increase Total Savings.}
For fixed feedback reduction, total cost savings increase with the feedback cost share. At one quarter of the frozen feedback tariff, shared-token AR is the cheapest workflow at 16 draws per prompt in all six GRPO cells, as detailed in Appendix~\ref{app:G-tariff}. Appendix~\ref{app:extensions} evaluates shared acceptance, residual grouping and budgeted scoring. Residual grouping reduces joint support by 2.1--3.9\% while raising recorded cost by 1.6--5.1\%.
\par

\section{Related Work}\label{sec:related}

\paragraph{Coupling and Shared Randomness.}
Couplings construct joint laws with prescribed marginals \citep{levin2017markov,thorisson2000coupling}. Common random numbers exploit dependence to sharpen simulated comparisons \citep{glasserman1992crn,glasserman2004monte}, and related ideas underlie unbiased MCMC \citep{jacob2020unbiased}, Gumbel-based sampling \citep{maddison2014astar,kool2019stochastic} and coupled rejection samplers \citep{corenflos2022coupled}. The Independent Star is the classical maximal coupling applied to each base--update pair. CAFA determines which scores their response banks require for a report.
\par

\Needspace{4\baselineskip}
\paragraph{Coupled Evaluation and Feedback Sharing.}
Coupled autoregressive generation \citep{benz2026coupled} uses shared randomness to reach the same language-model benchmark conclusions with provably fewer samples, although pairwise-comparison rankings can change. Tree-Coupled A/B Testing (TCAB) \citep{wen2026tcab} couples policies on a tree, sharing reward queries for matched components. Beyond caching identical responses, CAFA identifies their contribution to the report before scoring. All-matched draws need no scores for zero-sum gains. Single-gain reports exclude other updates. Covariance can require scores that cancel in the mean. The rule accepts all these samplers, including independent sampling, where score reductions are typically small.

\paragraph{Evaluating with Fewer Labels.}
Active comparison of prediction models \citep{sawade2012comparison} compares predictive models directly under a labeling budget. Active testing \citep{kossen2021testing} and active surrogate estimation \citep{kossen2022active} select informative labels, while subset selection \citep{polo2024tiny} and anchor points \citep{vivek2024anchor} reduce the evaluation set. These methods balance labeling cost against estimation error. CAFA removes only scores irrelevant to the reported statistic and composes with budgeted selection, as detailed in Appendix~\ref{app:H-budget}. We assume a scalar, versioned scorer. Pairwise judges \citep{zheng2023judge,chiang2024arena} define a different estimand whose value can depend on the coupling itself, as detailed in Appendix~\ref{app:A-checks}.
\section{Conclusion}\label{sec:conclusion}
CAFA makes score acquisition depend on the requested report. For a fixed response bank and an otherwise unrestricted scorer, its mean and mean-plus-covariance supports are both sufficient and necessary for exact recovery. Controlled updates show substantial feedback reductions beyond caching, and coupled workflows in the online GRPO study lower recorded cost at a common observed precision threshold. Numerical audits recover the fully scored reports and preserve every fixed checkpoint selection. The savings depend on response overlap and report choice: all gains together with the base mean require the full cached support, whereas focused comparisons can avoid many scores. Net workflow savings also depend on generation overhead and feedback prices. The evidence is limited to the evaluated models, checkpoints and scalar judges. Exact sample-statistic recovery does not establish population accuracy or agreement with human preferences.
\label{main-text-end}
\par\endgroup
\clearpage

\subsection*{Reproducibility Statement}
The assumptions and complete proofs of Propositions~\ref{prop:mean}--\ref{prop:cov} and Corollary~\ref{cor:report} are in Appendix~\ref{app:theory}. Models, prompts, the three checkpoint regimes, decoding and stopping settings, random-stream derivation, samplers, scoring identities and judge settings are in Appendix~\ref{app:protocol}. The online GRPO protocol and bank inventory are in Table~\ref{tab:grpo-protocol}.
\par

Every cell behind the main-text results remains available: Appendices~\ref{app:controlled}--\ref{app:grpo-evidence} visualize or tabulate them, and the complete per-cell records for the parts that are summarized there are in the canonical result tables of the supplementary material, which Appendix~\ref{app:I-artifacts} maps number by number. Cost conventions are in Appendix~\ref{app:cost}. The supplementary material contains the canonical result tables, the offline scripts for all empirical figures and generated tables, editable method diagrams, complete chart CSV files, the audit of every main-text number, and a code archive with the acquisition method, its tests and its local-generation adapters, as detailed in Appendix~\ref{app:software}. Regenerating the reported support, precision and cost values requires no GPU and no judge call. Regenerating the banks themselves requires the recorded model checkpoints.
\par
\subsection*{AI Use Statement}
Generative AI tools assisted with experiment and analysis code, software implementation and debugging, figure and table generation from frozen results, numerical verification, and language polishing. Responses and scores were produced by the model and judge calls specified in the experimental protocol in Section~\ref{sec:exp-setup}. The supplementary scripts compute the reported values from those frozen records. The authors take responsibility for the reported results and released artifacts.

\begingroup
\interlinepenalty=10000
\setlength{\bibsep}{1pt plus0.5pt minus0.5pt}
\hypersetup{urlcolor=black}
\fontencoding{T1}\selectfont

\endgroup

\clearpage
\appendix
\makeatletter
\setlength{\@fptop}{0pt}
\setlength{\@fpsep}{12pt plus 2pt minus 2pt}
\setlength{\@fpbot}{0pt plus 1fil}
\makeatother
\phantomsection

\section*{Appendix}\label{app:overview}

\startcontents[appendix]
\PrintAppendixContents

\par\noindent\begin{minipage}[t]{\linewidth}
\paragraph{Notation and Source-Record Aliases.}
The following table summarizes the notation and display aliases used in the appendix. The naming convention of Section~\ref{sec:intro} applies throughout.
\par\vspace{6pt}
\begingroup\small
\AddToHookNext{env/tabular/begin}{\renewcommand{\arraystretch}{1.0}}
\setlength{\LTleft}{0pt}
\setlength{\LTright}{0pt}
\edef\PaperTableBeforeNotation{\arabic{table}}
\renewcommand{\theHtable}{notation}
\begin{tabular}{@{}>{\raggedright\arraybackslash}p{0.28\linewidth}>{\raggedright\arraybackslash}p{\dimexpr0.72\linewidth-2\tabcolsep\relax}@{}}
\toprule
Notation & Description\\\midrule
$x_i$, $i$ & Prompt and prompt index.\\
$p_0,p_j$ & Base and updated policies.\\
$x=x_i$, $n=n_i$ & Single-prompt notation suppressing the prompt index.\\
$Y_{tj}=Y_{i,t,j}$ & Response of policy $j$ in joint draw $t$, with the prompt index suppressed.\\
$B_t[y]=b_{i,t,u}$ & Single-prompt draw coefficient.\\
$A[y]$, $A_{i,u}$ & Single-prompt and indexed forms of the aggregated coefficient map.\\
$y$, $u$ & Scoring identity, written as $y$ in single-prompt derivations and $u$ in the main text.\\
$\phi$, $\kappa$ & Identity map, also called ``decoded text'' in the source records.\\
$a_y$, $\bar b_y$ & Aliases for the per-prompt coefficient $A_{i,u}/n_i$.\\
$S$, $Q$ & Support set and its cardinality. The three counts are $\Qc=|\Scache|$, $\Qm=|\Smean|$ and $\Qd=|\Sdraw|$.\\
$\Fc$, $\Fm$, $\Fd$ & Feedback fees with the same cache, mean and mean-plus-covariance subscripts.\\
$G$, $F$, $T=G+F$ & Generation cost, feedback cost and total workflow cost.\\
Independent Star, Independent AR, Shared-token AR, TV tree & Samplers. ``star'' means Independent Star unless ``shared'' is stated. ``Token'' or ``shared tokens'' means shared-token (Gumbel) autoregressive sampling. ``Tree'' means the total-variation spanning tree.\\
CAFA (Ours), Cancellation & The post-sampling score-acquisition module and its coefficient-cancellation operation. All sampler configurations using CAFA use this module.\\
Cache-only & The scoring-strategy control for CAFA. Fixed-bank CAFA comparisons and cross-sampler workflow comparisons have different interpretations.\\
Checkpoint regimes & The three regimes in Table~\ref{tab:regimes} are never pooled.\\
Source-record names & Aliases affect display only. Original CSV keys, file paths and program arguments keep their recorded names. Appendix~\ref{app:I-artifacts} holds the mapping.\\
\bottomrule\end{tabular}
\setcounter{table}{\PaperTableBeforeNotation}
\endgroup
\end{minipage}\par

\section{Exact Recovery and Required Feedback}\label{app:theory}

\subsection{Pointwise Recovery and Mean-Query Necessity}\label{app:A-mean}

\paragraph{Proof of Proposition~\ref{prop:mean}.}
Fix prompt $x_i$. By definition of $b_{i,t,u}$,
$\sum_t C\,\boldsymbol r(Y_{i,t})=\sum_t\sum_j C_{:j}\,r(\kappa(x_i,Y_{i,t,j}))=\sum_u\bigl(\sum_t b_{i,t,u}\bigr)r(u)=\sum_u A_{i,u}r(u)$, and terms with $A_{i,u}=0$ vanish for every scorer. Dividing by $n_i$ and weighting by $w_i$ gives the fully scored estimator. For necessity, let an algorithm (possibly adaptive) return the report exactly for every admissible scorer while never querying some $(i,u)\in\Smean$ whose score it does not hold. Take two scorers equal on every queried input and differing at $u$ by $\delta\ne0$. Both are admissible because unknown coordinates vary independently over a nondegenerate interval. The algorithm sees the same transcript under both, but the reports differ by $w_i n_i^{-1}A_{i,u}\delta\ne0$, a contradiction. Scores already held need not be purchased, which is why the necessary support and the new paid calls are separate quantities. The argument assumes no restriction linking the scores of different inputs. Under such a restriction fewer queries could suffice.
\par

\subsection{Mean and Ordinary Covariance}\label{app:A-cov}
This subsection gives the proof of Proposition~\ref{prop:cov} with its oracle assumptions, the single-prompt argument and the stratified extension used for the main-experiment MCSEs. The argument requires distinct prompt identities. If the same full prompt appears in two nominal strata, those identities must first be merged. The software validates this condition when loading a bank, as detailed in Appendix~\ref{app:I-integration}.
\par
\subsubsection{Reporting Support: Proof, Assumptions, and Checks}\label{app:reporting-proof}

\paragraph{Oracle and Identity Assumptions.}
The bank, target coefficients, stopping policy and scoring identity are fixed before score acquisition. Each unqueried scoring coordinate may independently take any value in a nondegenerate interval. No known functional relationships link distinct coordinates. The lower bound also applies after some coordinates have been observed, because the argument holds other scores fixed. An adaptive exact-query algorithm that omits a required coordinate has the same transcript under two admissible completions with different outputs.  The ten-point score cache used in the experiment is fixed. The same argument applies to an unrestricted score alphabet with at least three distinct values, because a nonconstant quadratic cannot take the same value at three distinct points.
\par

\paragraph{Single Prompt.}
Set $\bar b_y=n^{-1}\sum_t b_{t,y}$. If $\bar b_y\ne0$, $\bar Z$ has a nonzero linear dependence on $r(y)$. If $\bar b_y=0$ but some $b_{t,y}\ne0$, choose a coordinate $k$ with a nonzero coefficient. Holding other scores fixed, write $Z_{t,k}=b_{t,y,k}s+v_{t,k}$. Then
\begin{equation}\label{eq:appendix-covariance-coordinate}
 \widehat\Sigma_{kk}(s)=\frac1{n-1}\sum_t
 \bigl((b_{t,y,k}-\bar b_{y,k})s+v_{t,k}-\bar v_k\bigr)^2.
\end{equation}
Its squared-term coefficient is positive. Hence at least one of the required outputs depends on every coordinate in $S_{\rm draw}$. Outside that support every draw coefficient is zero, so no required output depends on the score. This proves the exact support characterization.
\par

\paragraph{The Stratified MCSE Used in the Paper.}
Let prompt $x_i$ have $n_i\ge2$ draws, positive fixed weight $w_i$, sample mean $\bar Z_i$ and ordinary covariance $\widehat\Sigma_i$. The reported mean and Monte Carlo covariance estimate are
$\widehat\theta=\sum_i w_i\bar Z_i,\quad \widehat V=\sum_i\frac{w_i^2}{n_i}\widehat\Sigma_i$.
Here $w_i=1/P$ and $n_i=n$. The scoring key contains the exact prompt, so a score identity belongs to a single distinct prompt group. Its nonzero average coefficient changes $\widehat\theta$. If that coefficient is zero, a nonzero draw coefficient gives a strictly positive squared-term coefficient in a diagonal of $\widehat V$. The positive factor $w_i^2/n_i$ cannot remove this dependence. Thus the union of the prompt-wise draw supports is necessary and sufficient. If identical prompt keys are reused across nominal strata, their identities and coefficients must instead be collected together. The distinct-prompt argument must not be applied without this check. These algebraic conclusions do not require draws to be independent, although interpreting $\widehat V$ as a Monte Carlo covariance estimator does require the sampling assumptions stated elsewhere.
\par

\subsection{Reporting More Outputs or Changing Coordinates}\label{app:A-report}

\paragraph{Proof of Corollary~\ref{cor:report}.}
For a fixed identity write $h_u$ for its occurrence vector over policies, so that its aggregated coefficient is $Ch_u$ (and per draw $Ch_{t,u}$). If $H$ is invertible, $HCh_u=0$ if and only if $Ch_u=0$, so $S(HC)=S(C)$ for $\Smean$ and, draw by draw, for $\Sdraw$. Outputs transform as $H\widehat\theta$ and covariances as $H\widehat VH^\top$. A stacked coefficient $[C;D]h_u$ vanishes exactly when both $Ch_u$ and $Dh_u$ vanish, giving the~union.
\par

For $C=[-\1,I_J]$ stacked with $e_0^\top$ the square matrix is invertible, and every observed identity has $h_u\ne0$. Hence all observed identities are required and $S=\Scache$. The zero-sum gains alone suffice for checkpoint selection because subtracting the common base mean does not change the maximizer. Appendix~\ref{app:D-selection} reports the selection audit for the fixed online-study selectors. Changing the target, for example from base-relative to adjacent-checkpoint gains, is not a change of coordinates of the same report unless the map is invertible on the requested outputs. Its support is recomputed, as detailed in Appendix~\ref{app:E-reports}.
\par

\subsection{Immediate and Windowed CAFA}\label{app:A-immediate}
This subsection gives the full statement and proof of sign purity in Proposition~\ref{prop:signpure}, its decoded-text condition, and the property that windowed reporting with a cumulative coefficient map buys a superset of the batch support and never refunds a score. The measured window and support audits are in Appendix~\ref{app:C-mechanism}.
\par
\subsubsection{Immediate Acquisition and the Requested Statistic}\label{sec:immediate-theory}
Using the coefficients from Section~\ref{sec:reporting-support}, define
\begin{equation}\label{eq:appendix-acquisition-supports}
 S_{\rm draw}=\{y:\exists t,\ b_{t,y}\ne0\},\qquad
 S_{\rm mean}=\{y:\textstyle\sum_t b_{t,y}\ne0\}.
\end{equation}
Cancelling within a draw and caching acquires $S_{\rm draw}$. Final-batch aggregation acquires $S_{\rm mean}\subseteq S_{\rm draw}$. The set difference $S_{\rm draw}\setminus S_{\rm mean}$ is the additional mean-only saving available by waiting for the full batch. The simple pairwise implementation is therefore an exact implementation of the mean rule whenever these sets agree.
\par

\begin{proposition}[Sign purity at the sampled support]\label{prop:signpure}
For each base-relative contrast row under the Independent Star construction, fix a sampled sequence $z$. Across draws its nonzero coefficients in that row all have the same sign. Thus at sequence-level identity, $S_{\mathrm{mean}}=S_{\mathrm{draw}}$. This conclusion also holds for decoded-text groups if, for each contrast row, $p_j(z)-p_0(z)$ does not change sign among sequences in any one decoded group.
\end{proposition}
\par

A positive term at $z$ requires a residual endpoint, where $p_j(z)>p_0(z)$. A negative term requires a rejected root, where $p_j(z)<p_0(z)$. These supports are disjoint. Appendix~\ref{app:online} gives the decoded-text counterexample and the support audits. This result explains when immediate reporting already attains the final-batch CAFA query count.
\par

Once $S_{\rm draw}$ is acquired, every draw vector, its sample covariance and any fixed resampling statistic are available. The strict inclusion can occur in the mirror-draw example of Figure~\ref{fig:mirror-draws}. The requested statistics therefore remain an explicit input to CAFA.
\par

\subsubsection{Windowed Acquisition}
Let $A_w$ be the running coefficient map after window $w$ and $S_{\mathrm{acq}}^{\mathrm{mean}}=\bigcup_w\{y:A_w[y]\ne0\}$ the inputs accessed during the mean stage. Since $A_{n/W}=A$ is the full-batch map, $S_{\mathrm{mean}}\subseteq S_{\mathrm{acq}}^{\mathrm{mean}}\subseteq S_{\mathrm{draw}}$, with the first inclusion an equality at $W=n$. Windowing returns the fully scored mean at every window boundary. Earlier accesses remain in $S_{\mathrm{acq}}^{\mathrm{mean}}$ when later draws cancel their coefficients, so its size depends on the bank and $W$, as detailed in Appendix~\ref{app:C-mechanism}. For the final joint mean-and-covariance report, Algorithm~\ref{alg:cafa} completes the accesses to $S_{\mathrm{acq}}=S_{\mathrm{acq}}^{\mathrm{mean}}\cup S_{\mathrm{draw}}=S_{\mathrm{draw}}$ for every $W$. Cache hits and newly paid queries are recorded separately. Interval procedures are specified in Appendix~\ref{app:F-rare}.
\par

\subsection{Finite Verification and Interface Examples}\label{app:A-checks}
This subsection supplies the boundary examples (all-copy draws, mirror draws, a repeated pair, decoded-text merging and the absolute anchor), the finite enumeration checks of the source study, and a pairwise-preference counterexample showing why a pairwise judge changes the estimand. The enumeration counts are given below. These checks were run in the source study, and finite enumeration supplements rather than replaces the proofs of Appendices~\ref{app:A-mean}--\ref{app:A-immediate}. The released software has its own independent test suite, including the twelve acceptance fixtures and a dense oracle that does not share code with the coefficient path, as detailed in Appendix~\ref{app:I-tests}.
\par

\paragraph{Boundary Examples.}
All-copy draws have zero coefficients for zero-sum targets and require no feedback. The mirror draws of Figure~\ref{fig:mirror-draws} are the named boundary case in which mean-only acquisition is insufficient. Repeating $(A,B)$ twice has zero sample variance for all scores, but its mean still requires both scores: the proposition must request mean and covariance jointly. Decoded-text merging can create the first type of cancellation from distinct sampled sequences. The score key may merge them while their probability identities remain distinct. Appending an absolute anchor requires the full cached support even on all-copy banks.
\par

\paragraph{CPU Checks and Empirical Audit.}
The accompanying test enumerates all 729 three-draw, two-policy banks over three response identities, adds 700 vector-target cases, five named boundary cases and 200 stratified coefficient cases, and checks dependence by varying one score over three admissible values. It also checks invertible row changes and appended outputs. These finite checks supplement the proof. The separate frozen-bank recomputation of the main-study report cells is an audit of real banks and is reported in Appendix~\ref{app:D-preservation}.
\par

\paragraph{Why an Unrestricted Pairwise Judge Changes the Target.}
For a concrete boundary example, take three responses $A,B,C$ and an antisymmetric preference function with $q(A,B)=q(B,C)=q(C,A)=1$. It cannot equal $r(y')-r(y)$ on every pair: summing these three scalar differences would give zero rather than three. If both policies are uniform on the three responses, a coupling that always selects $(A,B),(B,C),(C,A)$ has expected preference $1$, while the reversed coupling has expected preference $-1$, with identical policy marginals.  Scalar-score expected differences are marginal quantities. Arbitrary pairwise preferences need an explicitly specified joint sampling law. A new preference API therefore cannot be inserted as a same-estimand baseline without additional assumptions.
\par

\section{Models, Data and Implementation Protocol}\label{app:protocol}

\subsection{Prompt and Checkpoint Provenance}\label{app:B-provenance}
Three checkpoint regimes appear in this paper. They answer different questions and are never pooled. In particular, the three controlled endpoints are three training seeds, not three points of one trajectory, and generation repetitions are never counted as training seeds.
\par

\begin{table}[!htbp]\centering\small\setlength{\tabcolsep}{4pt}
\caption{The three checkpoint regimes. Sources are the frozen job~records. The online protocol is Table~\ref{tab:grpo-protocol}.}\label{tab:regimes}
\begin{adjustbox}{max width=\linewidth}
\begin{tabular*}{\linewidth}{@{\hspace{\tabcolsep}\extracolsep{\fill}}>{\raggedright\arraybackslash}p{0.2\linewidth}>{\raggedright\arraybackslash}p{0.36\linewidth}>{\raggedright\arraybackslash}p{0.36\linewidth}}\toprule
Regime & Checkpoints & Training and use\\\midrule
Controlled fixed endpoints & Qwen2.5-1.5B-Instruct and SmolLM2-1.7B-Instruct, three endpoints \texttt{blend75\_202609084/085/086}, one per training seed & 8 optimizer steps at learning rate $10^{-7}$ on fixed responses using the objective in Appendix~\ref{app:B-updates}, saved offline and never retrained. Used in Section~\ref{sec:controlled} and Appendix~\ref{app:controlled}.\\
Offline distance scan & Steps 2, 8 and 32 of one trajectory per seed: three trajectories per small model, one for Qwen2.5-7B-Instruct & 32 steps at $10^{-7}$ repeating the eight fixed batches four times, FP32, not online GRPO, output caps 256 and 768. Used in Section~\ref{sec:configs} and Appendix~\ref{app:E-configs}.\\
Online GRPO & Qwen2.5-1.5B-Instruct base and steps 10, 50 and 250 of three 250-step trajectories & LoRA GRPO, learning rate $10^{-5}$, 16 prompts $\times$ 8 fresh completions per step. Used in Section~\ref{sec:grpo-main} and Appendix~\ref{app:grpo-evidence}.\\
\bottomrule\end{tabular*}
\end{adjustbox}\end{table}

The controlled studies use an initial bank of 16 fixed prompts, a comparison bank of 32 prompts drawn from historical response pools, and a strict holdout of 32 prompts that excludes all 614 recorded historical prompt texts, as detailed in Appendix~\ref{app:C-sets}. All prompts are open-ended generation tasks from No Robots \citep{h4norobots}, WildChat \citep{zhao2024wildchat} and ELI5 \citep{fan2019eli5}. The strict holdout has two sampling strata: three No Robots and thirteen WildChat prompts, and sixteen ELI5 prompts.
\par

The online study evaluates 128 prompts excluded from training and from historical pools with 12 No Robots, 52 WildChat and 64 ELI5 prompts, as specified in Table~\ref{tab:grpo-protocol}, split into 64 selection and 64 test prompts. The test split has 32 ELI5, 26 WildChat and six No Robots prompts. Strata name data sources. They are not replications. This subsection closes with how sampling budgets, residual ceilings and incomplete prefixes were handled in the controlled campaign.
\par

\subsubsection{Evidence Boundaries}

\paragraph{Sampling Budgets.}
Completed distance evaluations retain every frozen prompt and checkpoint. Residual pools resume from their saved proposal prefix when necessary, with a final ceiling of 8,192 proposals. This changes the computation allowance while retaining the candidate stream, model checkpoints and stopping policy. Saved-prefix hashes are audited in Appendix~\ref{app:revision-cost}. No exhausted prefix is treated as a zero-gain observation or removed from the completed results.
\par

The initial four-repeat Qwen study reuses a fixed prompt set and fixed short offline updates. The separately reported strict holdout provides new-prompt validation. Generation repetitions do not constitute additional training-seed replication. Finite-support expected scores are exact for the constructed proxy distributions and fixed cached scores only. The repeated baselines on the comparison bank and both 7B generation conditions have passed reconstruction checks. Their query-count evidence is reported in Appendices~\ref{app:controlled} and~\ref{app:targets}. Judge-conditioned precision and total-cost results are included only after their scoring coverage and accounting checks pass.
\par

\subsection{Update Procedures}\label{app:B-updates}
The controlled and distance checkpoints come from a fixed-response, length-normalized, advantage-weighted log-probability objective. Its definition and schedule are given here. It is not online GRPO. The online GRPO protocol, the evaluation grid and the final 93-bank inventory follow.
\par
\subsubsection{Fixed Update Trajectories}\label{app:update-training}
The distance study uses a frozen offline, group-relative log-probability objective. Each prompt group contains eight saved responses with fixed 40-rubric score vectors $s_i$. For the frozen weight vector $w$, let $R_i=w^\top s_i$ and $A_i=(R_i-\overline R_g)/(\operatorname{sd}_g(R)+10^{-4})$, with sample standard deviation within the eight-response group. The minibatch objective is
\begingroup
\setlength{\abovedisplayskip}{6pt plus1pt minus1pt}
\setlength{\belowdisplayskip}{6pt plus1pt minus1pt}
\setlength{\abovedisplayshortskip}{3pt plus1pt minus1pt}
\setlength{\belowdisplayshortskip}{6pt plus1pt minus1pt}
\begin{equation}
 \mathcal L(\phi)=-\frac1m\sum_{i=1}^m\frac{A_i}{|y_i|}\log p_\phi(y_i\mid x_i).
\end{equation}
\par\endgroup

FP32 Adam updates use learning rate $10^{-7}$ and gradient-norm clipping at 1. The same eight frozen minibatches are repeated four times, saving parameters after steps 2, 8 and 32. Qwen 1.5B and Smol 1.7B use seeds 202609084, 202609085 and 202609086. Qwen 7B uses the first seed. The 7B training candidates are retokenized with its own tokenizer, retaining decoded text and stopping status. These are fixed offline trajectories, not online GRPO rollouts. The evaluated-policy guarantees apply to their saved distributions. The released numeric configuration includes the full reward vector, minibatch sizes, source hashes and all seven trajectory protocol hashes.
\par

\subsubsection{Final GRPO Protocol and Result Inventory}\label{app:grpo-protocol}
The final bundle distinguishes the controlled fixed-response update experiments from the three fresh-response GRPO trajectories. The former use fixed offline endpoints reached by a fixed number of low-learning-rate update steps. The latter use online sampled completions and group-normalized~rewards.
\par
\begin{table}[!htbp]\centering\small
\captionsetup{skip=4pt}
\AddToHookNext{env/tabular*/begin}{\renewcommand{\arraystretch}{1.0}}
\caption{Frozen online GRPO configuration. Policy exports merge the saved adapters for FP32 evaluation.}\label{tab:grpo-protocol}
\begin{adjustbox}{max width=\linewidth}
\begin{tabular*}{\linewidth}{@{\hspace{\tabcolsep}\extracolsep{\fill}}lp{0.63\linewidth}}\toprule
Item & Value\\\midrule
Policy & Qwen2.5-1.5B-Instruct, FP32\\
Training implementation & TRL 0.24.0 GRPOTrainer, token importance ratios, group reward scaling, one iteration per rollout\\
Seeds and steps & 202609121, 202609122, 202609123, 250 optimizer steps each\\
Optimizer & AdamW, learning rate $10^{-5}$, 10 warmup steps\\
GRPO coefficients & KL coefficient 0.04, clipping parameter 0.2\\
Batch & 16 prompts $\times$ 8 fresh answers, microbatch 2, gradient accumulation 64\\
LoRA & Rank 32, alpha 64, dropout 0, q/k/v/o and gate/up/down projections\\
Training reward & Local Qwen2.5-7B-Instruct-AWQ rubric judge, original 40-criterion natural weights\\
Evaluation judge & Separate DeepSeek Flash cache with a frozen returned model identity and fingerprint\\
Sampling & Temperature 1, top-p 1, no top-k truncation, output cap 768, EOS IDs 151645 and 151643\\
Compared policies & Base and steps 10, 50, 250\\
Evaluation prompts & 128: No Robots 12, WildChat 52, ELI5 64, 64 selection and 64 test\\
Residual safeguards & 8,192 proposals per prompt/update/repetition, probability replay tolerance 0.01\\
\bottomrule\end{tabular*}
\end{adjustbox}\end{table}

The main matrix has 72 comparison banks and three independent reference banks. Residual-group validation adds 18 banks, for 93 in total. The versioned cache contains 338,653 successfully scored unique inputs across these banks. The count is an experimental union, not a deployment query count.
\par

Full scoring permits direct auditing of CAFA. Query savings in the paper compare the inputs each fixed task would purchase, while validation procurement, failed API responses and generation restarts remain in separate raw records. Historical studies retain their original model/judge identities and are not pooled into this cache count.
\par
\newpage
\enlargethispage{\dimexpr-2\baselineskip-\parskip\relax}

The released analysis contains 504 main-study target reports and 126 residual-group reports, with all mean and covariance identities checked. Twelve earlier isolated timing replays reproduce the original budget-4 independent-AR/shared-token response identities. Seven further uninterrupted replays cover all identified main-grid cells with failure/recovery histories. These 19 timing replays add no statistical samples. The six new Tree budget-4 banks do. The final grid, including its new cells, is independently reconstructed under the original 5,000-resample stratified test protocol. Three complete training backups, all frozen banks, token probabilities, namespaces, pilot decisions and attempt ledgers remain preserved. The paper-source bundle supplies the generated tables and figures. Raw-model and scoring assets remain separate because they are substantially larger.
\par

\subsection{Joint Samplers and Probability Computation}\label{app:B-samplers}

\paragraph{Independent Star on Token Sequences.}
Every policy is sampled with temperature 1, top-$p$ 1, top-$k$ 0, repetition penalty 1 and the stated length cap. The first end-of-sequence token ends a response, otherwise it is cut at the cap and kept. For a root sequence $X$ drawn from $p_0$, the log-probability of the complete token sequence (including the stopping event) is replayed under each $p_j$ with the same transform, and $X$ is copied into policy $j$ when $U_j\le\exp(\min\{0,\log p_j(X)-\log p_0(X)\})$. Otherwise residual proposals $Z\sim p_j$ are accepted with probability $\max\{0,1-p_0(Z)/p_j(Z)\}$ in their original order. Proposals drawn in batches are all charged, and a cell that reaches the proposal ceiling is resumed from its saved prefix or reported as incomplete, never filled with another sample. Densities are recorded and replayed on the original token IDs, within the replay tolerance in Table~\ref{tab:grpo-protocol}. Random streams are named and reproducible. Giving all models the same random seed and sampling them separately does not reproduce the Independent Star law.
\par

\paragraph{Baselines.}
Independent autoregressive sampling draws each policy with its own random stream. Shared-token (Gumbel) sampling uses the same named random stream to sample tokens for every policy. Two policies emit the same token whenever the shared noise yields the same argmax. The total-variation tree first estimates pairwise total-variation distances from a score-free pilot (16 selection prompts, eight draws per policy, log-probabilities only), builds the minimum spanning tree rooted at the base, and couples each child with its parent by the same copy-or-residual construction as the star, with the parent in place of the base. The pilot is charged to every tree task. These are established constructions used as inputs to CAFA. Shared acceptance and residual grouping are extensions, as detailed in Appendix~\ref{app:extensions}.
\par

\paragraph{Configuration a Reimplementation Must Match.}
Assumptions, decoding, stopping, identity hashing, random-stream derivation, proposal caps and scoring calls of the controlled campaign are extracted from the frozen job~records.
\par
\subsubsection{Assumptions and Reproduction Configuration}\label{app:repro}

\paragraph{Assumptions the Results Actually Need.}
\textbf{(A1) Fixed policies.} $p_0,\ldots,p_J$ do not change during evaluation and do not
depend on acquired feedback. This is what allows a query to be deferred until the whole
batch is aggregated. An adaptive policy may need feedback to choose its next action.
\textbf{(A2) A report fixed in advance.} $C$, the prompt set and the budget $n$ are all
fixed before acquisition, because each of them changes the query set. The zero-query result
for an all-copy draw additionally needs a zero-sum contrast, $C\1=0$. Anchored and other
general linear reports use the same aggregation with their own support.
\textbf{(A3) A scoring function that is deterministic given its version.} $r_x(y)$ depends
on the version, the prompt and the response and on nothing else. A stochastic judge is
reached through a versioned cache, so every claim here is a claim about that cached
function --- not about repeated judge noise, and not about human preference.
\textbf{(A4) Bounded experimental scores, otherwise unrestricted.} Boundedness gives
finite population moments and supports the concentration bounds. Fixed-bank recovery
does not require it. Independent score variation over a nondegenerate interval
makes the remaining queries necessary in Proposition~\ref{prop:cancel}. If scores
were known to be, say, a linear function of features, fewer queries could suffice.
\textbf{(A5) Two identities, kept apart.} Within a fixed scorer version, scoring identity
is exact equality of the prompt and decoded response. Probability identity is the sampled
token sequence under a given prompt and policy.
\textbf{(A6) Densities available on realized tokens.} Both star constructions need
$p_j(y)/p_0(y)$ on the realized tokens. We replay it and accept the run only within a
stated tolerance. The cancellation rule itself needs none of (A6).
\par

\paragraph{Configuration.}
Table~\ref{tab:repro} is extracted from the frozen protocol, density and judge records of
the completed jobs. A reimplementation that matches the
window, the aggregation scope, the cache key and the two identities will reproduce the
query counts exactly, because the counts are set functions of the realized texts. Matching
the generated texts themselves additionally requires the random-stream rows and the same
kernels.
\par

\begin{table}[!htbp]
\centering
\caption{Small-model star configuration for the comparison and strict-holdout banks, extracted from frozen job~records.}\label{tab:repro}
\small
\begin{adjustbox}{max width=\linewidth}
\begin{tabular*}{\linewidth}{@{\hspace{\tabcolsep}\extracolsep{\fill}}p{0.23\linewidth}>{\raggedright\arraybackslash}p{0.71\linewidth}}\toprule
Item & Value\\\midrule
Base policies $p_0$ & \texttt{Qwen/Qwen2.5-1.5B-Instruct} at snapshot \texttt{989aa7980e4c}, \texttt{HuggingFaceTB/SmolLM2-1.7B-Instruct} at snapshot \texttt{31b70e2e869a}\\
Fixed updates $p_1..p_{3}$ & \texttt{blend75\_202609084,\allowbreak{} blend75\_202609085,\allowbreak{} blend75\_202609086}, learning rate 1e-07, 8 optimizer steps, saved offline and never retrained during evaluation\\
Contrast & $C=[-\1,I_{3}]$, fixed before acquisition\\
Prompt sets & Comparison bank, 32 prompts (\texttt{30815c9a2a46}), strict holdout 32 prompts (\texttt{6e2d4578862c}) after excluding 614 recorded historical prompt texts\\
Sampling budget & 64 root draws per prompt, generated as 8 batches of 8, analysed at $n\in\{8, 16, 32, 64\}$\\
Decoding & temperature 1.0, top-$p$ 1.0, top-$k$ 0, typical-$p$ 1.0, repetition penalty 1.0, no-repeat $n$-gram 0, 1 beam, renormalized logits, cap 768 new tokens\\
Precision & FP32, \texttt{sdpa} attention\\
Stopping rule & first end-of-sequence token ends the response, otherwise the response is cut at 768 tokens and recorded as \texttt{length}, capped responses stay in the analysis\\
End-of-sequence ids & Qwen \texttt{[151645,\allowbreak{} 151643]\allowbreak{}} / pad \texttt{151643}, SmolLM2 \texttt{[2]\allowbreak{}} / pad \texttt{2}\\
Scoring identity & decoded text with special tokens removed, hashed as \texttt{sha256(json(\{QUESTION,\allowbreak{} RESPONSE\})\allowbreak{})\allowbreak{}}\\
Probability identity & the sampled token ids, every density is replayed on them and accepted within 0.01 nats\\
Random streams & seed \texttt{int(sha256(namespace)\allowbreak{}[:\allowbreak{}8]\allowbreak{},\allowbreak{}16)\allowbreak{}}, uniforms \texttt{(int(sha256(namespace)\allowbreak{}[:\allowbreak{}13]\allowbreak{},\allowbreak{}16)\allowbreak{}+0.5)\allowbreak{}/2**52}\\
Acceptance namespaces & independent \texttt{PREFIX:\allowbreak{}copy:\allowbreak{}independent:\allowbreak{}\textless{}checkpoint\textgreater{}\allowbreak{}:\allowbreak{}\textless{}pid\textgreater{}\allowbreak{}:\allowbreak{}\textless{}rid\textgreater{}\allowbreak{}}, shared \texttt{PREFIX:\allowbreak{}copy:\allowbreak{}shared:\allowbreak{}\textless{}pid\textgreater{}\allowbreak{}:\allowbreak{}\textless{}rid\textgreater{}\allowbreak{}} --- the update index is absent, so one uniform serves every update\\
Residual stream & \texttt{PREFIX:\allowbreak{}residual:\allowbreak{}\textless{}checkpoint\textgreater{}\allowbreak{}:\allowbreak{}\textless{}pid\textgreater{}\allowbreak{}:\allowbreak{}\textless{}used\textgreater{}\allowbreak{}}, proposal $Z\sim p_j$ accepted with probability $(1-p_0(Z)/p_j(Z))_+$\\
Residual proposal cap & 128 batches of 8 = 1024 proposals per prompt and update, one accepted pool serves both star arms in a job\\
Scoring versions & deepseek\_flash = \texttt{deepseek-flash}, qwen\_flash = \texttt{qwen3.8-flash}, qwen\_max = \texttt{qwen3.8-max-0902}\\
Scoring call & temperature 0, at most 256 output tokens, thinking disabled, json\_object with one integer key overall\_score in 1..10\\
Cache key & the eight fields \texttt{system}, \texttt{requested\_model}, \texttt{returned\_model}, \texttt{system\_fingerprint}, \texttt{temperature}, \texttt{thinking}, \texttt{question}, \texttt{response}, the last two are the hashed input, and the system prompt hashes to \texttt{538965bce0b3}\\
Aggregation & \texttt{coefficients[uid(base)\allowbreak{}]\allowbreak{} -= e\_c ;\allowbreak{} coefficients[uid(endpoint)\allowbreak{}]\allowbreak{} += e\_c}, query set \texttt{\{uid :\allowbreak{} coefficient vector is not the zero vector\}}\\
Identity assertion & \texttt{assert np.max(abs(total-naive)\allowbreak{})\allowbreak{}\textless{}1e-10}, evaluated for every method, budget and cached judge\\
Strong baselines & \texttt{independent\_ar}, \texttt{gumbel\_ar}, \texttt{TV\_MST}, the tree is chosen from a pilot of 8 samples per policy per prompt with no score access\\
Incomplete cells & primary comparisons retain every frozen prompt and propagate unavailable scores through $[1,10]$, historical common-coverage audits state their intersection explicitly\\
\bottomrule
\end{tabular*}
\end{adjustbox}
\end{table}

\subsection{Scoring, Identities and Frozen Analyses}\label{app:B-scoring}
A scoring identity is the SHA-256 of the exact prompt and decoded response (special tokens removed). The cache key adds the scorer version (rubric, judge model or deployment, temperature 0, thinking disabled, output schema). DeepSeek Flash returns a system fingerprint, checked on every call. Qwen Flash can only be pinned by model name. Successful scores, terminal refusals and pending calls are stored separately. Accessed identities (including cache hits) and newly paid calls are counted~separately.
\par

\subsection{Acquisition Algorithms}\label{app:B-algorithm}
The detailed algorithms below expand the core procedure in Algorithm~\ref{alg:acquire}. Both use the per-draw and per-prompt coefficients $b_{i,t,u}$ and $A_{i,u}$ of Equation~\eqref{eq:coefficients} and the supports $\Smean\subseteq\Sdraw$ of Section~\ref{sec:cancel}. They differ in when scores are bought and which statistics are requested. Algorithm~\ref{alg:acquire-details} is the specification of the released software: a fixed bank and an explicit report mode, so a mean-only report buys $\Smean$ and a mean-and-covariance report buys $\Sdraw$.
\par

Algorithm~\ref{alg:cafa} describes the implementation used in the source study to produce all reported bank results. It processes draws as they arrive, can report after every window of $W$ draws, always acquires the covariance support before reporting uncertainty, and accumulates its coefficient map rather than resetting it, so a score bought early is never refunded when later draws cancel it, as detailed in Appendix~\ref{app:A-immediate}.
\par

\begin{algorithm}[htbp]
\caption{Complete CAFA procedure for a fixed bank.}\label{alg:acquire-details}
\setlength{\baselineskip}{12.4pt}
\begin{algorithmic}[1]
\Require bank $\{Y_{i,t,j}\}$; report $C$ with statistics $s\in\{\text{mean},\ \text{mean+cov}\}$; scorer $r$ at version $v$ with available-score cache $\mathcal K_v$; optional score range $[r_{\min},r_{\max}]$; $n_i\ge2$ if $s=$ mean+cov
\For{each prompt $x_i$ and draw $t=1,\ldots,n_i$}
  \State $b_{i,t,\cdot}\gets 0$
  \For{each active column $j$} \Comment{no score is read here}
    \State $u\gets\kappa(x_i,Y_{i,t,j})$;\quad $b_{i,t,u}\gets b_{i,t,u}+C_{:j}$
  \EndFor
\EndFor
\State $A_{i,u}\gets\sum_t b_{i,t,u}$ for each prompt $i$ and identity $u$
\State $\Smean\gets\{(i,u):A_{i,u}\ne0\}$;\quad $\Sdraw\gets\{(i,u):\exists t,\ b_{i,t,u}\ne0\}$
\State $S\gets\Smean$ if $s=$ mean, else $\Sdraw$
\For{$(i,u)\in S$ such that $u\notin\operatorname{dom}(\mathcal K_v)$} \Comment{acquire missing feedback}
  \State $\mathcal K_v[u]\gets r(u)$, or record the failure; an unavailable score is never set to $0$
\EndFor
\State $M\gets\{(i,u)\in S:u\notin\operatorname{dom}(\mathcal K_v)\}$
\If{$M\ne\varnothing$}
  \If{a score range is declared}
    \State \Return per-output mean bounds (Appendix~\ref{app:F-missing}), no covariance; $|S|$ and the new paid calls
  \Else
    \State \Return an incomplete report with missing identities $M$; $|S|$ and the new paid calls
  \EndIf
\EndIf
\State $\bar Z_i\gets n_i^{-1}\sum_{u:A_{i,u}\ne0}A_{i,u}\mathcal K_v[u]$;\quad $\widehat\theta\gets\sum_i w_i\bar Z_i$
\If{$s=$ mean+cov}
  \State $Z_{i,t}\gets\sum_{u:b_{i,t,u}\ne0} b_{i,t,u}\mathcal K_v[u]$ for all $t$, including all-zero draws
  \State $\widehat\Sigma_i\gets$ sample covariance of $\{Z_{i,t}\}_t$;\quad $\widehat V\gets\sum_i w_i^2\widehat\Sigma_i/n_i$
\EndIf
\State \Return $\widehat\theta$, and $(\widehat\Sigma_i,\widehat V)$ if requested; $|S|$ and the new paid calls
\end{algorithmic}
\end{algorithm}

The windowed variant is presented below, together with notes from the source study on provenance, online reporting and the distinction between caching and contrast evaluation.
\par
\begin{algorithm}[t]
\caption{Contrast-Aware Feedback Acquisition (CAFA) for one prompt $x=x_i$.}\label{alg:cafa}
\setlength{\baselineskip}{12pt}
\begin{algorithmic}[1]
\Require contrast matrix $C\in\R^{d\times(J+1)}$ with $C\1=0$; budget $n\ge2$; window $W$
 dividing $n$; joint sampler $\mathcal{S}$ with marginals $p_0,\ldots,p_J$; scoring
 interface $R_v$ at version $v$ with cache $\mathcal{K}$ and key $\kappa$
\Ensure gain estimate $\widehat\theta(x)$, the query set $S_{\mathrm{mean}}$, the inputs
 accessed $S_{\mathrm{acq}}$, and an uncertainty record $U$
\State $A\gets$ empty map, scoring input $\mapsto$ coefficient in $\R^{d}$;\quad
 $S_{\mathrm{acq}}\gets\emptyset$
\For{$t=1,\ldots,n$}
  \State $(Y_{t0},\ldots,Y_{tJ})\gets\mathcal{S}(x)$
  \Comment{keep token IDs; densities use these IDs}
  \State $B_t\gets$ empty coefficient map
  \For{$j=0,\ldots,J$}
    \State $y\gets\mathrm{text}(Y_{tj})$
    \Comment{decode, drop special tokens}
    \State $B_t[y]\gets B_t[y]+C_{:j};\quad A[y]\gets A[y]+C_{:j}$
    \Comment{aggregate before querying}
  \EndFor
  \If{$t \bmod W=0$}
    \Comment{$W=n$: one batch; $W=1$: every draw}
    \State $S\gets\{y:A[y]\ne0\}\setminus S_{\mathrm{acq}}$
    \State $r[y]\gets\mathcal{K}[\kappa(v,x,y)]$, else $R_v(x,y)$ stored there, for $y\in S$
    \State $S_{\mathrm{acq}}\gets S_{\mathrm{acq}}\cup S$
  \EndIf
\EndFor
\State $S_{\mathrm{mean}}\gets\{y:A[y]\ne0\}$;\quad
 $\widehat\theta(x)\gets n^{-1}\sum_{y\in S_{\mathrm{mean}}}A[y]\,r[y]$
\State $S_{\mathrm{draw}}\gets\bigcup_{t=1}^n\{y:B_t[y]\ne0\}$;\quad
 $S_{\mathrm{acq}}^{\mathrm{mean}}\gets S_{\mathrm{acq}}$
\State $S_+\gets S_{\mathrm{draw}}\setminus S_{\mathrm{acq}}^{\mathrm{mean}}$
\State $r[y]\gets\mathcal{K}[\kappa(v,x,y)]$, else $R_v(x,y)$ stored there, for $y\in S_+$
\State $S_{\mathrm{acq}}\gets S_{\mathrm{acq}}\cup S_+$
  \Comment{count uncertainty accesses before reporting}
\State $U\gets\textsc{Uncertainty}(\{\sum_{y:B_t[y]\ne0} B_t[y]r[y]\}_{t=1}^n)$
\State \Return $\widehat\theta(x)$, $S_{\mathrm{mean}}$, $S_{\mathrm{acq}}$, $U$
\end{algorithmic}
\end{algorithm}

\paragraph{Acquisition and Provenance.}
Coefficients are accumulated separately for each prompt, contrast and budget. With $W=n$ the mean stage accesses exactly $S_{\mathrm{mean}}$. Uncertainty can enlarge that set. Smaller windows also support earlier reports and never refund an acquired score, because the coefficient map is cumulative: an early nonzero coefficient may cancel later. $|S_{\mathrm{acq}}|$ counts accessed inputs, including cache hits. Newly paid calls are recorded separately. Scoring identities, the cache key and the replay of probabilities on the original sampled token IDs are defined in Appendices~\ref{app:B-samplers} and~\ref{app:B-scoring}, and Appendix~\ref{app:repro} specifies the configuration.
\par

\paragraph{Online Reporting and Uncertainty.}\label{sec:uncertainty-interface}
The retained maps $B_t$ determine which feedback each draw requires, and their union support suffices to reconstruct all per-draw contrast vectors and hence their sample variance and covariance. The algorithm explicitly acquires that additional support before passing the draw vectors to the uncertainty routine. $U$ records the chosen estimand, sample covariance and interval procedure, and a terminally unavailable score instead produces the bounded record of Appendix~\ref{app:finalcontract} under the missing-feedback rules of Appendix~\ref{app:F-missing}.
\par

Proposition~\ref{prop:signpure}, as detailed in Appendix~\ref{app:A-immediate}, states the sequence-level condition under which $W=1$ incurs no extra feedback and the decoded-text condition it requires. The measured window and support audits are in Appendix~\ref{app:C-mechanism}. Other samplers can need extra accesses, and acquisition equality does not turn a finite-sample variance estimate into a coverage guarantee.
\par

\begin{figure}[!htbp]\centering
\includegraphics[trim=14.13bp 12.54bp 24.27bp 13.53bp,clip,width=0.952\linewidth]{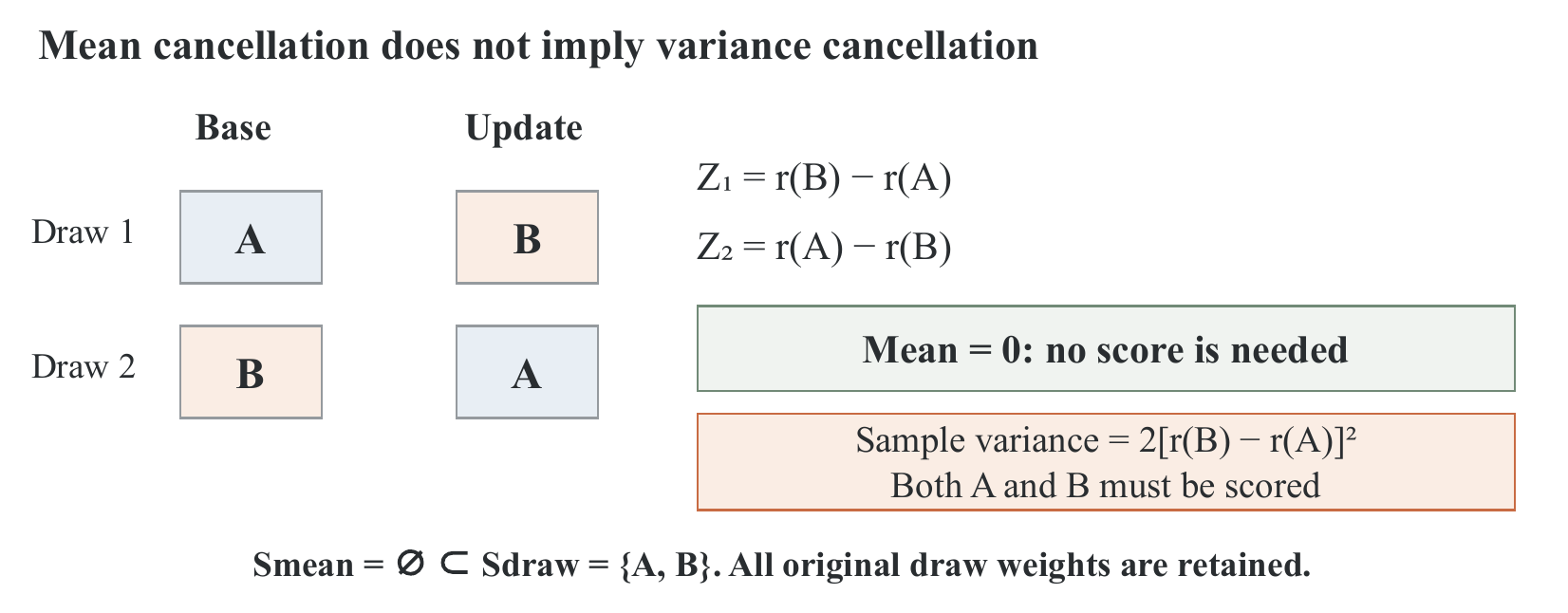}
\caption{The requested statistic determines which scores survive. The two mirror draws cancel in the mean, but their ordinary sample variance still depends on both scoring identities. Both draws and their original weights are retained.}\label{fig:mirror-draws}
\end{figure}

\section{Controlled-Update Evidence}\label{app:controlled}

\subsection{Experiment Sets and Completed Coverage}\label{app:C-sets}
The controlled experiments use the fixed-endpoint models of Table~\ref{tab:regimes} and the sampling and scoring protocol of Appendix~\ref{app:protocol}. Three prompt sets serve different purposes. An initial bank of 16 fixed prompts supports the first shared-acceptance replication in Appendix~\ref{app:H-shared}. The comparison bank contains 32 prompts drawn from historical response pools, with four generation repetitions for every sampler. The strict holdout contains 32 prompts and excludes all 614 recorded historical prompt texts. On the strict holdout each star arm has four generation repetitions and each other sampler two, and the four eight-prompt shards of a baseline repetition are scheduling units that together form one repetition, not independent replicates.
\par
\newpage

Each repetition uses 64 joint draws per prompt. Figure~\ref{fig:controlled} uses repetitions 1 and 2 for every sampler, whereas Figure~\ref{fig:highlight} and Table~\ref{tab:controlled} retain all four Independent Star repetitions, with repetitions 3 and 4 listed in Table~\ref{tab:controlled-extra-star}. Every realized pair and proposal is reconstructed from the saved tokens, probabilities and random namespaces. Appendix~\ref{app:repro} lists the configuration a reimplementation must match. Nonthinking API judges return integer scores from one to ten, so the query counts of Appendix~\ref{app:C-acquisition} do not depend on the returned values, and judge-conditioned precision results stay separated by returned model identity. Missing feedback is handled as specified in Appendix~\ref{app:F-missing}. Every bank retains all prompts and none is removed from the primary analysis.
\par
\newpage
\suppressfloats[t]
\enlargethispage{\dimexpr-3\baselineskip-\parskip\relax}

\subsection{Complete CAFA Results}\label{app:C-acquisition}
Tables~\ref{tab:fullcancel} and~\ref{tab:samplers} report the comparison bank, while Table~\ref{tab:controlled} reports the strict holdout. The former is not a holdout. In Table~\ref{tab:fullcancel}, the two Independent Star columns use identical independent-star samples under Cache-only and under CAFA, whereas the third column keeps CAFA and changes the sampler to shared acceptance: the CAFA saving and the shared-acceptance increment have different denominators and must not be added.
\par

Table~\ref{tab:controlled}(A) retains the complete strict-holdout results for all five sampler arms, with four generation repetitions for each star arm and two for every other sampler. Main-text Figure~\ref{fig:controlled} instead uses the first two repetitions for every sampler, so its Star averages differ slightly from this full-set table. Part~(A) keeps $\Qm$ and $\Qd$ in separate columns: for Qwen independent AR the mean-only support 8132.50 is the mean over two repetitions of 8132 and 8133, whereas the mean-plus-covariance support is 8133 in both repetitions, which is the value in Table~\ref{tab:revision2-unified} and in the archived full-repetition precision table. Part~(B) is in USD. The remaining tables of this appendix keep the original CNY of the frozen records.
\par

\begin{table}[t]\centering\small\setlength{\tabcolsep}{8.36118pt}
\caption{Controlled strict holdout. The complete repetition set uses 32 prompts, three fixed updates, 64 joint draws per prompt and $R$ generation repetitions, with counts averaged over repetitions. Figure~\ref{fig:controlled} restricts every sampler to repetitions 1 and 2. (A)~Scoring inputs for the three gains on the same sampled banks, for all five sampler arms: $\Qc$ deduplicates, $\Qm$ and $\Qd$ are the supports for the mean and for the mean with ordinary covariance. ``Removed'' is $1-\Qm/\Qc$, the mean-only convention. Independent Star and Shared-acceptance Star are two different samplers scored under the same CAFA rule, each with its own sampled bank. Their rows are not two readings of one bank. (B)~Independent Star, cost per repetition in USD at the frozen token tariffs, as detailed in Appendix~\ref{app:G-definitions}, with the full joint-star generation $G$ charged to both sides, so the saving is feedback avoided on the same responses.}\label{tab:controlled}
\begin{adjustbox}{max width=\linewidth}
\begin{tabular*}{\linewidth}{@{\hspace{\tabcolsep}\extracolsep{\fill}}llrrrrr}\toprule
\rowcolor{PaperHeader}\multicolumn{7}{l}{\textbf{A. Scoring support} (same responses on both sides of each row)}\\
Model & Sampler & $R$ & $\Qc$ & $\Qm$ & $\Qd$ & Removed (\%)\\\midrule
\rowcolor{PaperAccentLight}Qwen 1.5B & Independent Star & 4 & 2700.00 & 1193.25 & 1193.25 & 55.8\\
 & Shared-acceptance Star & 4 & 2708.75 & 1058.25 & 1058.25 & 60.9\\
 & TV tree & 2 & 2509.50 & 917.00 & 917.00 & 63.5\\
 & Shared-token AR & 2 & 5409.00 & 5169.50 & 5169.50 & 4.4\\
 & Independent AR & 2 & 8133.00 & 8132.50 & 8133.00 & 0.0\\
\addlinespace[2pt]
\rowcolor{PaperAccentLight}SmolLM2 1.7B & Independent Star & 4 & 2157.50 & 212.75 & 212.75 & 90.1\\
 & Shared-acceptance Star & 4 & 2166.25 & 190.50 & 190.50 & 91.2\\
 & TV tree & 2 & 2144.00 & 191.50 & 191.50 & 91.1\\
 & Shared-token AR & 2 & 3068.50 & 1850.00 & 1850.00 & 39.7\\
 & Independent AR & 2 & 8192.00 & 8192.00 & 8192.00 & 0.0\\
\midrule
\rowcolor{PaperHeader}\multicolumn{7}{l}{\textbf{B. Cost per repetition, Independent Star} (USD, same generation $G$ in both totals)}\\
Model & Judge & $R$ & $G$ & $G{+}F_{\rm cache}$ & $G{+}F_{\rm mean+cov}$ & Saved (\%)\\\midrule
Qwen 1.5B & DeepSeek Flash & 4 & 1.285 & 1.785 & 1.503 & 15.8\\
 & Qwen Flash & 4 & 1.285 & 1.481 & 1.371 & 7.5\\
SmolLM2 1.7B & DeepSeek Flash & 4 & 1.327 & 1.741 & 1.369 & 21.4\\
 & Qwen Flash & 4 & 1.327 & 1.490 & 1.344 & 9.8\\
\bottomrule\end{tabular*}
\end{adjustbox}
\end{table}
\begin{table}[!t]\centering\small
\caption{Additional Star repetitions retained outside the main comparison. Strict holdout, 64 joint draws per prompt. These are repetitions 3 and 4, excluded from Figure~\ref{fig:controlled} but included in the four-repetition averages of Table~\ref{tab:controlled} and Figure~\ref{fig:highlight}. Required inputs recover both the mean and covariance. The mean-only counts coincide.}\label{tab:controlled-extra-star}
\begin{adjustbox}{max width=\linewidth}
\begin{tabular*}{\linewidth}{@{\hspace{\tabcolsep}\extracolsep{\fill}}llrrr}\toprule
Model & Repeat & Cached & Required & Removed (\%)\\\midrule
Qwen 1.5B & 3 & 2,696 & 1,194 & 55.7\\
Qwen 1.5B & 4 & 2,715 & 1,198 & 55.9\\
SmolLM2 1.7B & 3 & 2,158 & 214 & 90.1\\
SmolLM2 1.7B & 4 & 2,168 & 231 & 89.3\\
\bottomrule\end{tabular*}
\end{adjustbox}
\end{table}

\subsubsection{How Much Feedback Is Avoidable}\label{sec:avoidable}
Table~\ref{tab:fullcancel} separates two effects on the comparison bank: eliminating zero coefficients in an already sampled batch, and changing acceptance dependence to create additional cancellations. These are averages over four repetitions, with 64 root draws per prompt and three fixed updates. Full aggregation is verified to reproduce the naive cached-score contrast in every checked cell, at every budget and for every cached judge.
\par

\begin{table}[!htbp]
\centering
\caption{Required scoring inputs per judge on the comparison bank, averaged over four repetitions. The two Independent Star columns use exactly the same independent-acceptance star samples under Cache-only and under CAFA. The third column keeps CAFA and changes the sampler to shared acceptance, so its increment has a different denominator. Counts are per judge. Monetary costs are evaluated separately.}\label{tab:fullcancel}
\begin{adjustbox}{max width=\linewidth}
\begin{tabular*}{\linewidth}{@{\hspace{\tabcolsep}\extracolsep{\fill}}lD{.}{.}{4.2}D{.}{.}{4.2}D{.}{.}{4.2}}\toprule
 & \multicolumn{2}{c}{Independent Star} & \multicolumn{1}{c}{Shared acceptance}\\
\cmidrule(lr){2-3}\cmidrule(lr){4-4}
Model & \multicolumn{1}{c}{Cache-only} & \multicolumn{1}{c}{CAFA} & \multicolumn{1}{c}{CAFA}\\\midrule
Qwen 1.5B & 2786.25 & 1322.50 & 1224.75\\
Smol 1.7B & 2170.50 & 240.00 & 194.25\\
\bottomrule\end{tabular*}
\end{adjustbox}
\end{table}

CAFA removes \AutoStageFiveQwenSaving{}\% and \AutoStageFiveSmolSaving{}\% of the initially unique inputs. Shared acceptance then removes \AutoStageFiveQwenIncrement{}\% and \AutoStageFiveSmolIncrement{}\% of what is left, a different denominator that must not be added to the first. On the strict 32-prompt holdout, whose counts are the Independent Star and shared-acceptance rows of Table~\ref{tab:controlled}(A), CAFA removes \HoldoutQwenSaving\% and \HoldoutSmolSaving\% of the deduplicated inputs, with per-repetition ranges of \AutoHoldoutQwenRange{}\% and \AutoHoldoutSmolRange{}\% that the averaged table does not show. Those prompts exclude 614 recorded historical prompt texts. The before--after comparison uses identical independent-star~responses.
\par
\subsubsection{Does the Saving Survive a Different Sampler?}\label{sec:samplers}
The same CAFA rule applies to every joint sampler. Table~\ref{tab:samplers} reports four independent generation repetitions for each strong baseline, using the same 32 prompts in the comparison bank and 64 root draws per prompt. Independent autoregressive sampling exposes essentially no removable vector-valued queries. Shared token randomness creates agreement and therefore some cancellation, while the score-free total-variation tree exposes a substantially larger reduction. On the tree samples, CAFA removes \AutoStageFiveQwenTreeSaving{}\% of the unique inputs for Qwen and \AutoStageFiveSmolTreeSaving{}\% for SmolLM2. The CAFA benefit therefore extends to a strong competing coupling construction. It is not specific to the star sampler.
\par

\begin{table}[!htbp]
\centering
\caption{Unique scoring inputs on the comparison bank at $n=64$, averaged over four independent generation repetitions per model and sampler. Cache-only and CAFA use identical sampled responses within each row. All samplers receive the same cancellation and cache rules. These counts do not require judge scores.}\label{tab:samplers}
\begin{adjustbox}{max width=\linewidth}
\begin{tabular*}{\linewidth}{@{\hspace{\tabcolsep}\extracolsep{\fill}}lD{.}{.}{4.2}D{.}{.}{4.2}D{.}{.}{4.2}D{.}{.}{4.2}}\toprule
 & \multicolumn{2}{c}{Qwen 1.5B} & \multicolumn{2}{c}{Smol 1.7B}\\
\cmidrule(lr){2-3}\cmidrule(lr){4-5}
Sampler & \multicolumn{1}{c}{Cache-only} & \multicolumn{1}{c}{CAFA} & \multicolumn{1}{c}{Cache-only} & \multicolumn{1}{c}{CAFA}\\\midrule
Independent AR & 8154.75 & 8154.75 & 8192.00 & 8192.00\\
Shared tokens & 5678.50 & 5512.50 & 3059.50 & 1845.25\\
TV tree & 2586.00 & 1059.50 & 2146.25 & 195.50\\
\bottomrule\end{tabular*}
\end{adjustbox}
\end{table}

The tree requires a mean of \AutoStageFiveQwenTree{} inputs on Qwen, with a run range of \AutoStageFiveQwenTreeRange{}, and \AutoStageFiveSmolTree{} on SmolLM2 with a run range of \AutoStageFiveSmolTreeRange{}. The shared-acceptance means it is compared with are the last column of Table~\ref{tab:fullcancel}. The cost--precision comparison of Appendix~\ref{app:C-precision} evaluates these samplers jointly on acquisition demand and Monte Carlo standard error. Baseline and star runs share prompts but use independently generated Monte Carlo streams. The strict-holdout baseline experiment repeats this comparison on the held-out bank with separately frozen sampling namespaces. Its judge-conditioned comparison is evaluated separately.
\par
\newpage

\subsubsection{Strict Holdout: The Same CAFA Scoring Strategy for Every Sampler}\label{sec:currentholdout}
The strict holdout compares samplers on the same 32 prompts under the repetition and shard convention of Appendix~\ref{app:C-sets}. All samplers receive exact-text caching and the same CAFA module, so the counts differ only through the responses each sampler produces. Table~\ref{tab:controlled}(A) reports $\Qc$, $\Qm$ and $\Qd$ for every sampler arm. Figure~\ref{fig:currentprecision} pairs the support needed for sample variances with the precision it buys.
\par

\subsection{Where the Savings Arise}\label{app:C-mechanism}
Three implementations are compared on the same responses: text caching alone, skipping zero coefficients within each draw and caching the rest, and aggregating across the batch. The theory behind the equality for the star is Proposition~\ref{prop:signpure}. The measured evidence comes from five audits with different scopes and different units of analysis:
\par
\begin{center}\small\renewcommand{\arraystretch}{1.0}
\begin{adjustbox}{max width=\linewidth}
\begin{tabular}{@{}>{\raggedright\arraybackslash}p{0.25\linewidth}@{\hspace{6pt}}>{\raggedright\arraybackslash}p{0.25\linewidth}@{\hspace{6pt}}>{\raggedright\arraybackslash}p{\dimexpr0.50\linewidth-12pt\relax}@{}}\toprule
Audit & Unit & Scope and result\\\midrule
Mechanism snapshot & 67 sampler--job cells & Historical 32-job campaign, within-draw plus cache equals batch aggregation in every cell\\
Window sweep & 469 per-job budget rows & Same campaign, windows 1 to 64, every window attains the batch count\\
Completed support audit & 56 joint-target cells & Strict-holdout and distance banks, all recorded samplers and trajectories, equal in 55\\
Online GRPO preservation & 504 report cells & Equal in 485, every exception in a non-star sampler, as detailed in Appendix~\ref{app:D-preservation}\\
Access audit & 336 base-relative star cells & 14 scored jobs, no vector-level gap\\
\bottomrule\end{tabular}
\end{adjustbox}
\end{center}
These audits share data and count different units. Their totals are not independent replications and must not be added into one verification sample size.
\par
\subsubsection{When Is the Saving Available?}\label{sec:online}
We separate text deduplication, cancellation within each root draw followed by caching, and cancellation across the entire batch. The last two have exactly equal required-input counts in the mechanism snapshot, and every audited window attains the same final acquisition count, with scopes and counts in Appendix~\ref{app:C-mechanism}. In the audited star cells these savings are therefore available online: reporting after every draw requires no additional scores. The expanded audit checks this equality again for each new sampler and budget. Generic non-star outputs can require additional inputs for variance. Appendix~\ref{app:online} gives the decomposition, the sequence-level sign-purity result and its decoded-text boundary. Exact sample variance is available at the same acquired support in the audited star cells. Confidence-interval coverage remains a separate question.
\par
\newpage
\suppressfloats[t]
\enlargethispage{\dimexpr-3\baselineskip-\parskip\relax}

\subsubsection{Online CAFA and Uncertainty Access}\label{app:online}
Section~\ref{sec:cancel} and Appendix~\ref{app:A-immediate} define the per-draw coefficient $b_{t,y}=\sum_{j:\phi(Y_{tj})=y}C_{:j}$, where $\phi$ decodes the sampled sequence, and the two supports $S_{\mathrm{mean}}\subseteq S_{\mathrm{draw}}$. Cancelling within each draw and caching accesses $S_{\mathrm{draw}}$. Whole-batch cancellation accesses $S_{\mathrm{mean}}$. Their equality is a directly checkable condition under which immediate reporting attains the batch CAFA query count, and Proposition~\ref{prop:signpure} gives the sequence-level case together with its decoded-group condition. The examples and audits below establish how it applies to the recorded scoring identities.
\par

\paragraph{Why Decoding Matters.}
Sequence-level splitting does not automatically give maximal coupling of decoded texts. A concrete boundary uses four atoms with $p_0=(.4,.1,.4,.1)$ and $p_1=(.1,.4,.1,.4)$, decoded as $(X,X,Y,Z)$. Two noncopy draws, atom pairs $(1,4)$ and $(3,2)$, have positive joint probability $0.0225$. Their decoded contrasts are $(X,Z)$ and $(Y,X)$: the score of $X$ cancels from the mean but remains in the draw contributions. With $r(Y)=0$, $r(Z)=2$, setting $r(X)$ to zero or one preserves the mean at one and changes the sample variance from two to zero. Therefore our token-space proposition is not an unconditional decoded-cache theorem. The actual text-support equality is audited directly.
\par

\begin{figure}[t]
\captionsetup{skip=3pt}
\centering\includegraphics[trim=15.66bp 5.42bp 5.5bp 5.64bp,clip,width=0.840\linewidth]{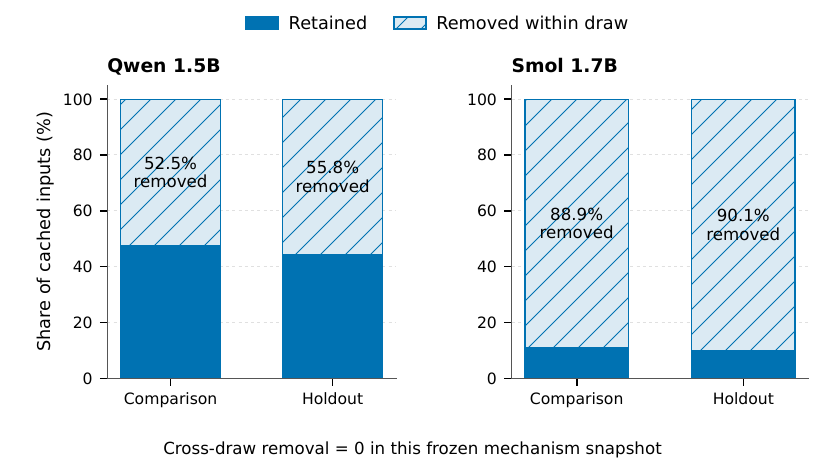}
\caption{Decomposition on fixed generated samples. Cancelling within a draw and then caching attains the complete observed reduction. Cross-draw aggregation adds no measured vector-query saving in this snapshot.}\label{fig:online-decomp}
\end{figure}

\paragraph{Measured Mechanism.}
The snapshot and window scopes are listed in Appendix~\ref{app:C-mechanism}. Within that historical snapshot, agreement holds in every cell, including the available tree and autoregressive baselines. What that list does not show is why cross-draw aggregation adds nothing here: across the audited records 115 scoring inputs occur in more than one root draw, and none of them eliminates an additional vector-valued query by cancelling across draws. The saving therefore does not require delayed feedback on these samples. Figure~\ref{fig:online-decomp} separates the mechanisms.
\par

\paragraph{Which Uncertainty Quantities Are Available?}
Given $S_{\mathrm{draw}}$, the vectors $D_t=\sum_y b_{t,y}r_x(y)$ are known exactly. Their sample variance, covariance and resampling statistics can then be computed without additional score acquisitions. This is exact recovery of sample statistics, not exact knowledge of population variance or guaranteed confidence-interval coverage. For a general sampler, the mean may have a smaller support. Appendix~\ref{app:A-cov} and Figure~\ref{fig:mirror-draws} give the boundary case. If all mean-support scores are known but a draw-support score is missing, the mean remains exactly recoverable while the covariance is unavailable.
\par

The access audit listed in Appendix~\ref{app:C-mechanism} covers 43,781 input incidences and finds no vector-level gap. For individual contrast rows of other samplers, gaps do occur: at $n=64$, the tree has 2 hidden groups among 2,556, and independent autoregressive sampling has 7 among 36,829. A row-level gap is not a joint-target gap. These row-level gaps need not produce extra purchases when another row of the jointly reported vector already acquires the same input. This explains their coexistence with the vector-level zero gap above.
\par

\subsection{Precision, Gains and Cost}\label{app:C-precision}

\paragraph{Two MCSE Summaries, Kept Apart.}
Precision on the strict holdout is summarized in two ways that use different repetition sets and aggregation orders. The resulting summaries are not interchangeable. This appendix shows the one that puts every sampler on a common repetition count: Table~\ref{tab:revision2-unified} restricts each sampler to its first two generation repetitions and averages, over those repetitions, the root-mean-square MCSE across the three gains, reporting 1190.50 mean inputs for the Qwen independent~star.
\par

\begin{figure}[!htbp]
\captionsetup{skip=3pt}
\centering\includegraphics[trim=9bp 0bp 2.3bp 3.64bp,clip,width=0.860\linewidth]{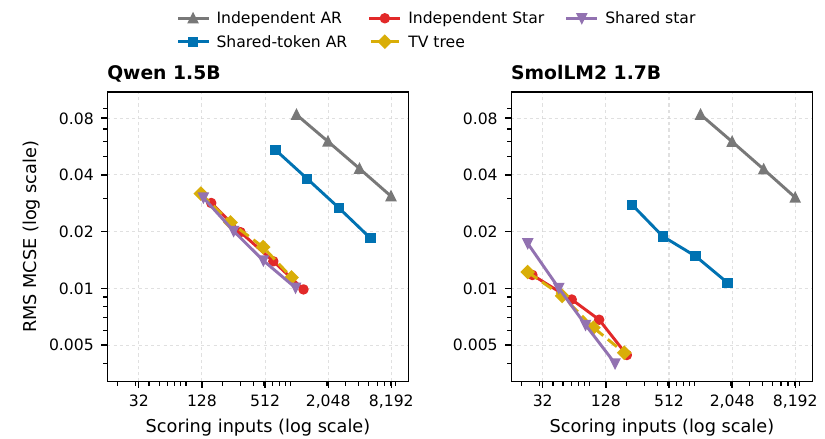}
\caption{Feedback demand and Monte Carlo precision across samplers. Strict holdout, DeepSeek Flash, fixed budgets $n\in\{8,16,32,64\}$, first two generation repetitions for every sampler. The horizontal axis includes any additional variance-support inputs. The vertical axis averages the single-repetition RMS MCSE across the three gains. Both axes are logarithmic. Known-truth error is evaluated in the finite-support experiment.}\label{fig:currentprecision}
\end{figure}

The other summary uses the complete repetition set for each sampler: four repetitions for the star arms and two for every other sampler. For each update, MCSE is averaged across repetitions before the root mean square is computed across updates. Under this convention, the same arm has 1193.25 variance-support inputs. Complete values, the aggregation definition and score status for each row are retained in the supplementary numeric archive. Both summaries were reproduced from the unrounded records for each repetition. Neither corrects the other, and their rows should not be combined in a single table. Table~\ref{tab:controlled} continues to carry the full-repetition scoring requirement, and Figure~\ref{fig:currentprecision} uses the first two repetitions with the single-repetition RMS convention.
\par

\paragraph{Cost under a Common Generation Charge.}
Table~\ref{tab:controlled}(B) charges the complete joint-star generation task to both sides of each comparison, so its percentage isolates feedback reduction on identical generated responses. It is not a comparison of two independently deployed workflows, and it is the four-repetition result, whereas main-text Figure~\ref{fig:controlled}(b) uses the first two repetitions. The same four-repetition amounts in their recorded CNY, per repetition and judge, are retained in the supplementary numeric archive. Cross-sampler workflow costs with their own overhead conventions are in Table~\ref{tab:revision2-unified}, break-even prices in Appendix~\ref{app:G-tariff}, the general pricing definitions in Appendix~\ref{app:G-definitions}, and the recorded task intervals behind them in Appendix~\ref{app:revision-cost}.
\par

At the fixed root budget, the tree requires fewer scoring inputs on Qwen, while the tree and shared star are nearly tied on SmolLM2. Figure~\ref{fig:currentprecision} places these counts beside their observed precision. A count advantage alone does not establish a precision-matched winner.
\par

\paragraph{Snapshot and Incomplete Feedback.}
The primary score snapshot is \texttt{966b4222f3b1}. The supplementary cost and application analysis is identified by \texttt{\RevisionEvidence}. Valid scores, terminal provider refusals and pending calls are tracked separately. An unavailable score contributes its full interval $[1,10]$ through its signed coefficient. No prompt is dropped and no missing value is replaced by zero. Unless a figure specifies fixed-prompt Monte Carlo intervals, gain intervals resample prompts within the two frozen sampling strata. Fixed-prompt Monte Carlo standard errors condition on these prompts and the versioned score cache.
Complete campaign usage and the distinction between task expenditure and per-evaluation costs are reported in Appendix~\ref{app:revision-cost}.
\par
\newpage
\enlargethispage{\dimexpr-2\baselineskip-\parskip\relax}

\begin{figure}[!htbp]
\captionsetup{skip=3pt}
\centering\includegraphics[trim=0bp 8.35bp 5.5bp 1.1bp,clip,width=0.870\linewidth]{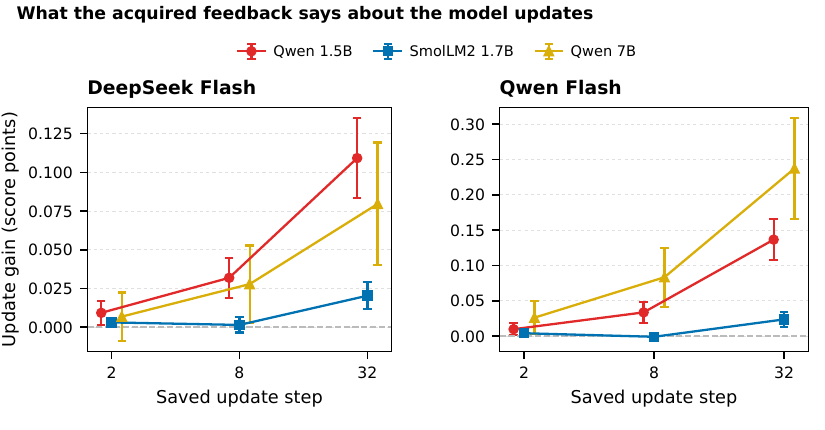}
\caption{Quality improvement and comparison resolution in the offline checkpoint scan. These are the measured quality changes at checkpoints 2, 8 and 32 of the offline distance regime. Two fixed Flash judges score the same generated responses. Error bars are the fixed-prompt Monte Carlo intervals (mean $\pm1.96$ fixed-prompt MCSE). The small-model configurations have three trajectories and the 7B extension has one.}\label{fig:update-gains}\label{tab:update-gains}
\end{figure}

\paragraph{The Gains Recovered by Those Scores.}
Appendix~\ref{app:update-training} specifies the frozen offline objective and training schedule that produce the compared checkpoints.
Figure~\ref{fig:update-gains} reports the corresponding quality changes at the 768-token cap. These gains are measured on the ten-point scale of the fixed, versioned judge. The small-model entries average three saved update trajectories. The 7B entries follow one trajectory. Each uses 32 prompts and 64 root draws per prompt per trajectory. Thus the figure shows both the magnitude of the update and the Monte Carlo resolution at which it is compared. Per-trajectory and prompt-resampling results accompany the machine-readable tables.
\par

\paragraph{Complete Cost for the Same Generated Comparison.}
We reconstruct complete task intervals from the start and exit events recorded by the persistent controller. These intervals include model loading, response generation, cross-policy probability computation, residual proposals, saving and in-process waiting. Reserved GPU time is priced at the experimental rental rate of 3 CNY per card-hour. The independent and shared star arms were generated together using a shared residual pool. We charge that entire recorded generation cost to both sides of each before--after comparison. Consequently, no allocation of shared generation overhead is needed to establish the CAFA saving.
\par

To price the scores that cancellation had made unnecessary, we query the frozen unscored responses with the same two version-checked judges and retain their actual input and output token counts. The comparison uses a common cold-cache tariff for both arms: 2/8 CNY per million input/output tokens for DeepSeek Flash and 0.8/2.7 for Qwen Flash. This separates the algorithmic difference from changing platform discounts. The generation bank, stopping rule and realized gain are identical before and after CAFA.
\par


The resulting total-cost reduction is \TotalSavingRange\%, while every realized gain is unchanged. The DeepSeek column uses its published peak tariff. Applying its half-price nonpeak tariff to both scoring policies gives \OffpeakSavingRange\% total savings for that judge. Coefficient processing is measured separately on the local CPU. Appendix~\ref{app:revision-cost} separates this per-evaluation comparison from historical campaign spending and from the observed phase durations of the independently generated strong baselines.
\par
\newpage
\enlargethispage{\dimexpr-2\baselineskip-\parskip\relax}

\paragraph{An Integrated Comparison across Recorded Workflows.}
Table~\ref{tab:revision2-unified} joins CAFA scoring requirements, precision and expenditure on the same strict holdout at $n=64$. All methods use their first two generation repetitions. The four-repeat star results remain in Table~\ref{tab:controlled}. The MCSE column is the mean, across these two repetitions, of the RMS standard error across the three gains. The score column reports the cost of acquiring the union of the mean and variance supports, computed from observed input/output token usage at the same historical normalized tariff.
\par

\begin{table}[!htbp]
\centering\footnotesize\setlength{\tabcolsep}{4pt}
\caption{Strict holdout, first two repetitions, DeepSeek Flash. ``Extra'' counts variance inputs beyond mean support. Costs are CNY per repetition. Star generation charges the complete joint-star task. The other methods charge their recorded inclusive phase plus all common startup/finalization overhead. These are reconstructed workflow costs with the stated overhead conventions.}\label{tab:revision2-unified}
\begin{adjustbox}{max width=\linewidth}
\begin{tabular*}{\linewidth}{@{\hspace{\tabcolsep}\extracolsep{\fill}}llrrrrrr}\toprule
Model & Sampler & Mean inputs & Extra & MCSE & Gen. & Score & Total\\\midrule
Qwen 1.5B & Ind. AR & 8132.5 & 0.5 & 0.0308 & 6.85 & 10.11 & 16.97\\
Qwen 1.5B & Shared tokens & 5169.5 & 0.0 & 0.0185 & 6.91 & 6.86 & 13.76\\
Qwen 1.5B & TV tree & 917.0 & 0.0 & 0.0114 & 7.93 & 1.08 & 9.01\\
Qwen 1.5B & Ind. star & 1190.5 & 0.0 & 0.0099 & 8.58 & 1.48 & 10.06\\
Qwen 1.5B & Shared star & 1000.5 & 0.0 & 0.0101 & 8.58 & 1.22 & 9.80\\
Smol 1.7B & Ind. AR & 8192.0 & 0.0 & 0.0303 & 5.74 & 10.57 & 16.32\\
Smol 1.7B & Shared tokens & 1850.0 & 0.0 & 0.0107 & 5.77 & 2.58 & 8.35\\
Smol 1.7B & TV tree & 191.5 & 0.0 & 0.0046 & 7.33 & 0.25 & 7.58\\
Smol 1.7B & Ind. star & 203.0 & 0.0 & 0.0044 & 8.07 & 0.27 & 8.33\\
Smol 1.7B & Shared star & 156.5 & 0.0 & 0.0040 & 8.07 & 0.22 & 8.28\\
\bottomrule\end{tabular*}
\end{adjustbox}
\end{table}

\paragraph{Historical Precision Diagnostics.}
An earlier snapshot of the same campaign is superseded for every current conclusion by Tables~\ref{tab:controlled} and~\ref{tab:revision2-unified}. Two of its diagnostics motivated checks in the current design: comparisons at that stage were underpowered, and some prompt cells had zero empirical variance. Both are properties of that historical snapshot, not rates for the completed experiments. The finite-support and rare-event checks that address them are in Appendices~\ref{app:F-proxy} and~\ref{app:F-rare}.
\par
\newpage
\enlargethispage{\dimexpr-3\baselineskip-\parskip\relax}

\subsubsection{Historical Precision Diagnostics and Their Support}\label{app:frontier}
In that snapshot the tree coverage was one Qwen repetition and two SmolLM2 repetitions against four star repetitions, on separate generation streams that share prompts rather than matched root draws. Its per-sampler counts and stage ordering are superseded and are retained only in the supplementary~archive.
\par

We retain two diagnostics from that historical snapshot. Across 184 pairwise standard-error comparisons, no reported interval excluded zero and 88 rows were flagged as underpowered, which does not demonstrate equivalent precision. In some SmolLM2 configurations, 30--49\% of prompt cells had zero empirical variance because all sampled pairs agreed. This rate applies to the historical snapshot rather than the completed experiments. Zero observed variance does not imply zero population variance. A precision threshold selected from such estimates can therefore look artificially favourable. The completed evidence uses all planned repetitions: strict-holdout strong baselines were absent here, and their completed generations enter Appendix~\ref{app:C-acquisition} and main-text Figure~\ref{fig:controlled}.
\par

\section{Online GRPO Evidence}\label{app:grpo-evidence}
The online study in Table~\ref{tab:regimes}, with its protocol in Table~\ref{tab:grpo-protocol}, evaluates base and steps 10, 50 and 250 of three GRPO trajectories on 128 held-out prompts. Its 72 comparison banks are four samplers, three budgets, three training seeds and two generation repetitions. Seven report targets give 504 report cells. They are fixed workflows, not 504 or 72 independent training runs, and generation repetitions are not training seeds.
\par

\subsection{Online GRPO Trajectories and Checkpoint Selection}\label{sec:grpo}
This appendix uses the three independent online LoRA GRPO \citep{shao2024deepseekmath} trajectories of Table~\ref{tab:regimes}, whose objective, optimizer, LoRA, batch, judge and sampling settings are in Table~\ref{tab:grpo-protocol}. They are not the controlled fixed-response updates of Appendix~\ref{app:controlled}, and the two regimes are never pooled.
\par

Six TV-tree budget-4 banks complete the original grid using the same exports, prompts, sampling protocol and frozen judge. They are genuinely new generation, but no additional training or reference generation was performed for them. Every TV-tree task pays for its own score-free pilot on 16 stratified selection prompts with eight answers per policy and prompt, charged to that task. Original token IDs determine probabilities. Exact prompt and decoded-response identity determines scoring reuse.
\par

\subsection{Complete Bank, Target and Budget Results}\label{app:D-cells}
Figures~\ref{tab:review3-single-1}--\ref{tab:review3-joint} show the same-bank CAFA increment for all four samplers, all three budgets and six of the seven reports, covering all 432 cells. The seventh report, gains plus the base mean, yields zero savings in all 72 banks. This result is stated in the text, with per-bank records in the supplementary tables. Table~\ref{tab:grpo-cells} reports every $n=16$ workflow cell, including RMS MCSE, generation expenditure derived from accepted runtimes and columns absent from the three-gain rows of Figure~\ref{tab:review3-joint}. The six tree banks at 4 draws per prompt that complete the grid are included in the 72 banks, as detailed in Appendix~\ref{app:completion-timing}. Figure~\ref{fig:grpo-cost} is the six-panel version of main-text Figure~\ref{fig:grpo-cost-main}, drawn from the same cells. The interpretation block retains the tariff sensitivity and the precision-threshold readout. Both are descriptive readings of the completed fixed-budget grid, not stopping rules.
\par

\subsubsection{Increment Attributable to CAFA}\label{sec:review3-increment}
We first keep each generated bank and scoring function fixed. The three implementations are text caching alone, within-draw zero-coefficient skipping followed by caching, and whole-batch aggregation followed by the additional queries needed for ordinary covariance. Text caching requires $\Qc$ inputs, and within-draw skipping requires $\Qd$. Batch aggregation first requires $\Qm$ inputs for the mean and then $\Qd-\Qm$ additional inputs for covariance. Thus within-draw skipping and batch aggregation acquire the same support $\Sdraw$, of size $\Qd$, for the joint mean-and-covariance report, as established in Proposition~\ref{prop:report-support}.
\par

Figure~\ref{tab:review3-joint} prices the CAFA layer independently of coupling. Each target uses only its active policy columns for the caching baseline. The generation charge $G$ is the same full recorded four-policy, 128-prompt task in both conditions, including for a single-target row. Thus the absolute saving is identified on the frozen bank, while the percentage uses an explicit joint-generation allocation, not a fresh pair-only runtime. $F$ prices observed input/output token usage at the frozen judge tariff. CPU execution time is reported separately in Table~\ref{tab:review3-cpu}, as detailed in Appendix~\ref{app:G-bookkeeping}.
\par

All 504 target cells reproduce the dense, fully scored means and stratified covariance to numerical precision. The independent audit reports a maximum absolute covariance difference of $4.3\times10^{-19}$, as detailed in Appendix~\ref{app:D-preservation}. All 72 fixed-rule choices are preserved. Mean and draw support agree in 485 cells. The other 19 require extra feedback for covariance, rather than a free reuse of the mean-only result. For example, independent AR, seed 1/repeat 2 at $n=16$, needs 4065 inputs for the step-10 mean and 4071 for the joint mean/covariance report.
\par
\subsubsection{Workflow Comparisons and Checkpoint Selection}\label{sec:grpo-workflow}
\begin{figure}[!t]\centering
\captionsetup{skip=3pt}
\includegraphics[trim=1.26bp 5.25bp 3.52bp 4.04bp,clip,width=0.800\linewidth]{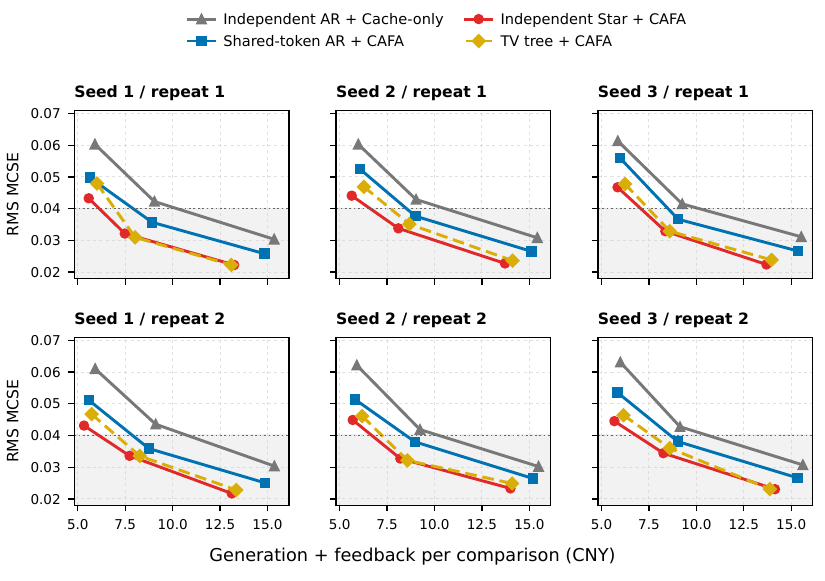}
\caption{Per-seed and per-repetition cost--precision curves underlying main-text Figure~\ref{fig:grpo-cost-main}. Panels separate three training seeds and two generation repetitions on the 128-prompt GRPO bank. Along each curve, budgets increase from 4 to 8 to 16 draws per prompt from left to right. RMS MCSE summarizes the three base-relative gains. The shaded region marks RMS MCSE $\leq0.04$. Independent AR uses Cache-only. The other samplers use CAFA (Ours). Costs include generation and feedback for means and covariance, with the Tree pilot and accepted uninterrupted runtimes. Costs are in CNY. The main-text figure uses USD.}\label{fig:grpo-cost}
\end{figure}

All 72 comparison banks reproduce the fully scored means and fixed checkpoint-selection decisions after CAFA. Variance uses the per-draw support whenever it exceeds the mean support. Figure~\ref{fig:grpo-cost} reports each training seed and generation repetition separately, with complete recorded generation costs and measured feedback token usage at the frozen tariff. Twelve earlier isolated replays replace contended independent-AR/shared-token budget-4 collection times. Seven additional uninterrupted replays replace cells with failure/recovery histories. Replays match the original token and text identities and add no statistical samples. All GPU times are normalized at 3 CNY per card-hour. This is a common accounting tariff, not the new server invoice. Historical charges are preserved separately, as detailed in Appendix~\ref{app:completion-timing}.
\par

\subsubsection{What the GRPO Results Imply for CAFA}\label{sec:grpo-interpretation}

\paragraph{Useful Training Progress and Lower Comparison Cost Coexist.}
Training progress and comparison cost are separate measurements, and the independent test references of Appendix~\ref{app:D-selection} keep them apart: each reference uses independent autoregressive samples, so improvement is not agreement induced by a sampler. For all three trajectories, the final checkpoint has a higher judge score than the base, as reported in Tables~\ref{tab:grpo-quality} and~\ref{tab:review3-reference}. The $n=16$ input, cost and MCSE ratios behind this are reported once, in main-text Table~\ref{tab:workflow}, with the per-cell breakdown in Table~\ref{tab:grpo-cells}. These comparisons hold the prompt bank and draw budget fixed. They do not select a stopping time using the observed~MCSE.
\par

\paragraph{Separating Sampler-Induced Agreement from CAFA Savings.}
Two different quantities are easy to conflate. On a fixed star bank, CAFA removes a further 2.9--3.8\% of the inputs already deduplicated within that workflow at $n=16$. The larger cross-workflow saving of Appendix~\ref{sec:grpo-workflow} also includes the agreement that the coupling itself creates, and the two must not be added. The additional reduction on a fixed bank depends on the requested report: Table~\ref{tab:targets} and the complete grid of Figures~\ref{tab:review3-single-1}--\ref{tab:review3-joint} give it per target, and it is far larger for a nearby contrast than for one that includes a distant update. Within-draw and batch supports agree in 72 of 72 joint cells. The coefficient interface identifies when the simpler within-draw implementation already attains the batch support.
\par

\paragraph{Fewer Labels Do Not Identify the Cheapest Sampler.}
At $n=16$, TV-tree reduces scoring demand by 36.7--42.5\% relative to independent AR plus caching, but its pilot and generation overhead also enter the comparison. Figure~\ref{fig:grpo-cost} and Table~\ref{tab:workflow} therefore compare total expenditure with precision, rather than selecting a winner from label counts.
\par

The timings they use are not all same-machine benchmarks: some cells were remeasured in uninterrupted replays that reproduce the original tokens and decoded scoring identities and add no statistical observations, and the accepted historical timings are retained for the rest. Appendix~\ref{app:revision-cost} records which cells were replaced, why, and what the historical-minus-replay difference contains.
\par


The independent test gains and their intervals are reported once, in Tables~\ref{tab:grpo-quality} and~\ref{tab:review3-reference} in Appendix~\ref{app:D-selection}, under the current 5,000-resample source-stratified procedure. An exploratory 2,000-resample unstratified analysis of the 64 test-prompt means gave 95\% percentile intervals $[0.220,0.417]$, $[0.256,0.455]$ and $[0.239,0.409]$ for the step-250 gains of seeds 1--3. That is a different resampling procedure on a different unit, not an earlier printing of Table~\ref{tab:review3-reference}, and the two sets of intervals are never shown as one series. Both resample test prompts conditional on the collected generations. Fixed-prompt generation MCSE is reported separately, and the training seeds remain separate throughout. The step-10 prompt-bootstrap intervals span zero for all three seeds under either procedure.
\par

Holding the measured generation costs and response lengths fixed, scaling the feedback tariff changes which $n=16$ workflow is cheapest. Appendix~\ref{app:G-tariff} reports the complete result at one quarter, one and twice the frozen tariff. These are observed expenditure comparisons at a fixed draw budget, and the complete cost--MCSE curves retain that second dimension.
\par
\newpage
\enlargethispage{\dimexpr-2\baselineskip-\parskip\relax}

A precision-constrained readout of the same fixed grid $\{4,8,16\}$ gives a larger operational difference, reported in Section~\ref{sec:grpo-main}: at RMS MCSE at most $0.04$, independent star with eight draws is the cheapest eligible workflow in all six cells while independent AR needs 16 draws. The 0.04 threshold is a post hoc readout of the completed fixed-budget grid. At four draws per prompt, the TV tree has RMS MCSE 0.0461--0.0479, so none of its six cells meets the 0.04 threshold.
\par

\paragraph{Checkpoint Selection Needs Differences, Not an Absolute Anchor.}
Adding the base mean to all three gains restores the full cached scoring support in every GRPO cell, because those outputs jointly identify all four absolute policy means. Checkpoint selection has a different requirement: subtracting the common base mean preserves the maximizer, so the zero-sum gains suffice, which is the algebra of Appendix~\ref{app:A-report}. The selection results themselves, the single disagreeing selector and its estimated regret are reported in Appendix~\ref{app:D-selection}. That agreement counts fixed workflows, not independent training trials.
\par

\paragraph{Practical Choice.}
Two decisions are separable. Given a bank and a report, use CAFA: it preserves the estimate and never increases the required score set. Choosing the generation workflow is a different decision, made from the measured cost--precision curve at the feedback tariff that applies, because how much CAFA saves depends on the report and how much the workflow costs depends on the sampler. Coverage evidence is reported separately in Appendices~\ref{app:F-proxy} and~\ref{app:F-rare}.
\par

\subsubsection{Completed-Grid Frozen-Bank Audit}\label{app:review3-audit}
All cells below are recomputed from the 75 frozen main/reference banks. The source manifest hashes every bank, its completion marker, the versioned score export, protocol, questions, cost ledger and accepted isolated timing replays. The original sources are checked again after the audit. Each CSV/JSON row retains the bank hash, complete target, seed, generation repeat, score identity and cost convention. No response generation, judge request or new training run is performed.
\par

\paragraph{Reading the CAFA Tables.}
AR, Token, Star and Tree denote independent autoregressive, shared-token, independent-star and TV-tree generation. $\Qc$, $\Qm$, $\Qd$ count the inputs required by cache-only scoring, mean-only reporting and joint mean-and-covariance reporting. Immediate skipping and batch aggregation with covariance both acquire $\Qd$ inputs. The additional covariance fee is $\Fd-\Fm$. The heatmaps and accompanying CSV files retain all budgets and all seven targets. $\Fc,\Fm,\Fd$ are scoring CNY. $\Tc=G+\Fc$ and $\Td=G+\Fd$ charge the full 128-prompt four-policy task. They do not identify fresh subset or single-target deployment costs. The machine-readable cells additionally expose the original attempt charge and whether an accepted isolated replay replaced it.
\par
\begingroup\scriptsize\setlength{\tabcolsep}{3pt}
\begin{figure}[!htbp]\centering\normalsize
\includegraphics[trim=23.47bp 20.28bp 7.48bp 4.44bp,clip,width=0.737\linewidth]{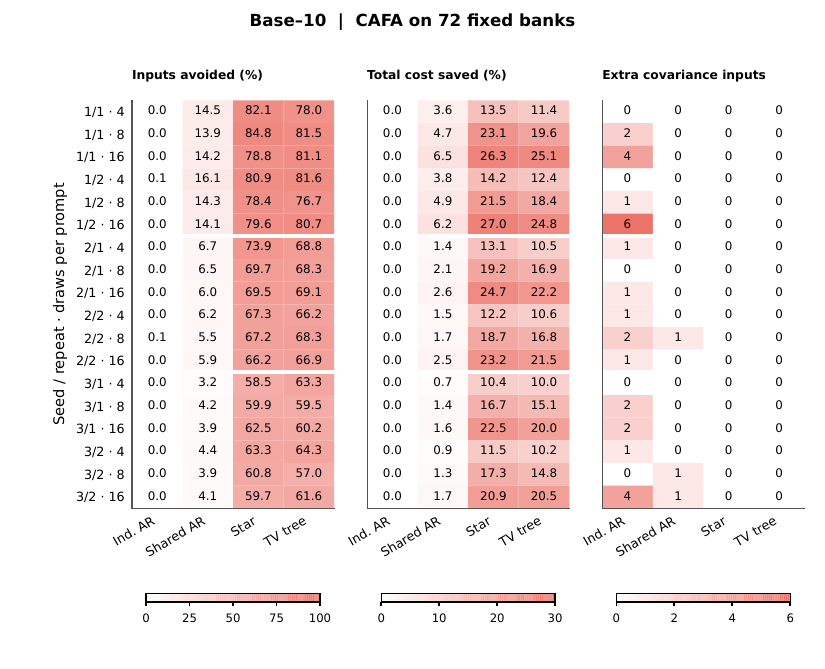}
\caption{Complete acquisition grid: Base--10. Every sampler, budget, seed and generation repeat is shown (72 cells). Left: $100(1-\Qd/\Qc)$, middle: same-bank total-cost saving $100(1-\Td/\Tc)$, right: $\Qd-\Qm$, the additional identities needed for covariance. Monetary ratios use the recorded CNY totals. The complete counts, fees and unrounded values accompany the figure in the CSV bundle. Color scales are shared across report targets for the two percentage panels.}\label{tab:review3-single-1}
\end{figure}
\endgroup
\begingroup\scriptsize\setlength{\tabcolsep}{3pt}
\begin{figure}[t]\centering\normalsize
\captionsetup{skip=3pt}
\includegraphics[trim=23.47bp 20.05bp 7.48bp 4.39bp,clip,width=0.630\linewidth]{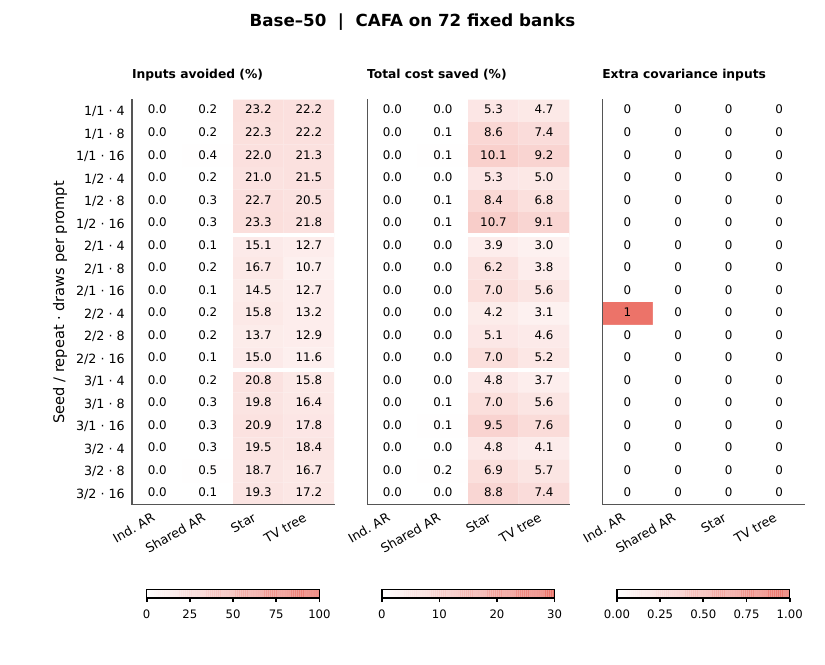}
\caption{Complete acquisition grid: Base--50. Continuation of Figure~\ref{tab:review3-single-1} for this report: all 72 cells (every sampler, budget, seed and generation repeat), the same three panels ($100(1-\Qd/\Qc)$, the same-bank total-cost saving $100(1-\Td/\Tc)$, and the covariance increment $\Qd-\Qm$), the same shared colour scales, and the same CNY basis. The complete counts, fees and unrounded values accompany the figure in the CSV bundle.}\label{tab:review3-single-2}
\end{figure}
\endgroup
\begingroup\scriptsize\setlength{\tabcolsep}{3pt}
\begin{figure}[t]\centering\normalsize
\includegraphics[trim=23.47bp 20.05bp 7.48bp 4.39bp,clip,width=0.737\linewidth]{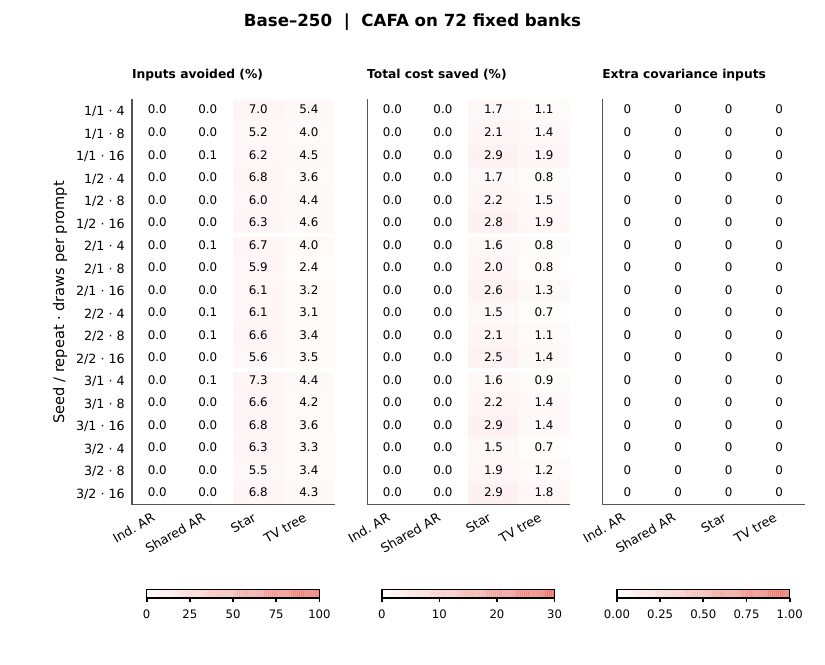}
\caption{Complete acquisition grid: Base--250. Continuation of Figure~\ref{tab:review3-single-1} for this report: all 72 cells (every sampler, budget, seed and generation repeat), the same three panels ($100(1-\Qd/\Qc)$, the same-bank total-cost saving $100(1-\Td/\Tc)$, and the covariance increment $\Qd-\Qm$), the same shared colour scales, and the same CNY basis. The complete counts, fees and unrounded values accompany the figure in the CSV bundle.}\label{tab:review3-single-3}
\end{figure}
\endgroup
\begingroup\scriptsize\setlength{\tabcolsep}{3pt}
\begin{figure}[t]\centering\normalsize
\captionsetup{skip=3pt}
\includegraphics[trim=23.47bp 20.05bp 7.48bp 4.39bp,clip,width=0.660\linewidth]{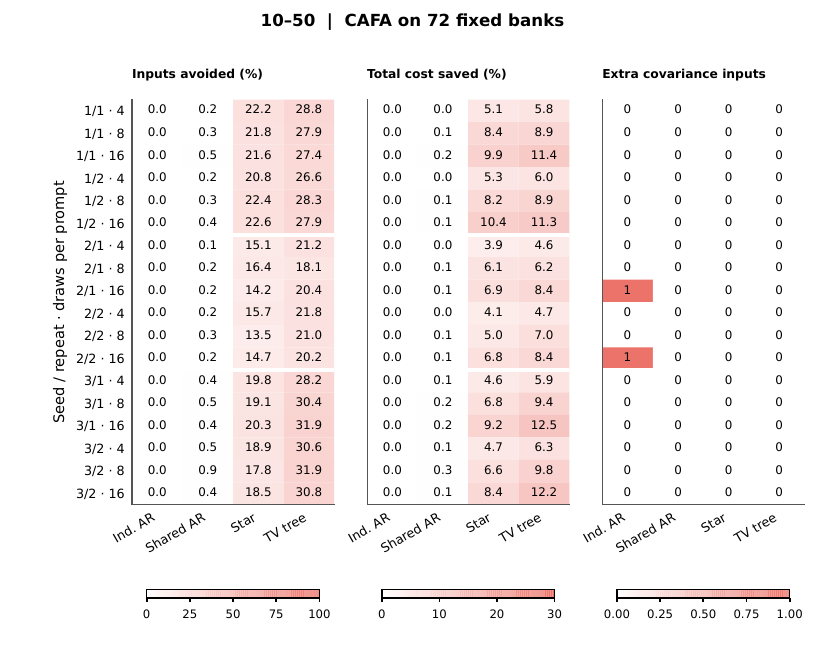}
\caption{Complete acquisition grid: 10--50. Continuation of Figure~\ref{tab:review3-single-1} for this report: all 72 cells (every sampler, budget, seed and generation repeat), the same three panels ($100(1-\Qd/\Qc)$, the same-bank total-cost saving $100(1-\Td/\Tc)$, and the covariance increment $\Qd-\Qm$), the same shared colour scales, and the same CNY basis. The complete counts, fees and unrounded values accompany the figure in the CSV bundle.}\label{tab:review3-adjacent-2}
\end{figure}
\endgroup
\begingroup\scriptsize\setlength{\tabcolsep}{3pt}
\begin{figure}[t]\centering\normalsize
\includegraphics[trim=23.47bp 20.05bp 7.48bp 4.39bp,clip,width=0.737\linewidth]{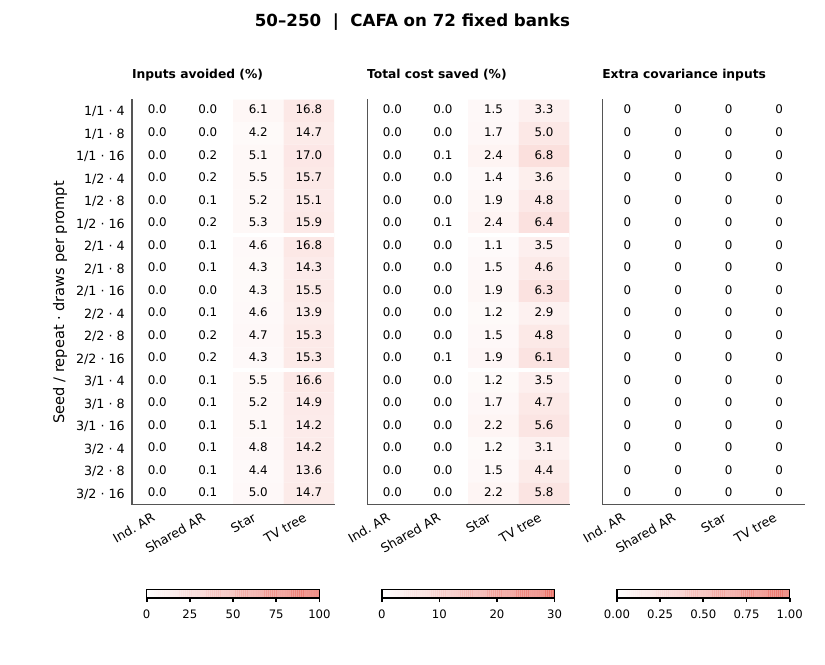}
\caption{Complete acquisition grid: 50--250. Continuation of Figure~\ref{tab:review3-single-1} for this report: all 72 cells (every sampler, budget, seed and generation repeat), the same three panels ($100(1-\Qd/\Qc)$, the same-bank total-cost saving $100(1-\Td/\Tc)$, and the covariance increment $\Qd-\Qm$), the same shared colour scales, and the same CNY basis. The complete counts, fees and unrounded values accompany the figure in the CSV bundle.}\label{tab:review3-adjacent-3}
\end{figure}
\endgroup
\begingroup\scriptsize\setlength{\tabcolsep}{3pt}
\begin{figure}[!htbp]\centering\normalsize
\includegraphics[trim=23.47bp 20.05bp 7.48bp 4.39bp,clip,width=0.737\linewidth]{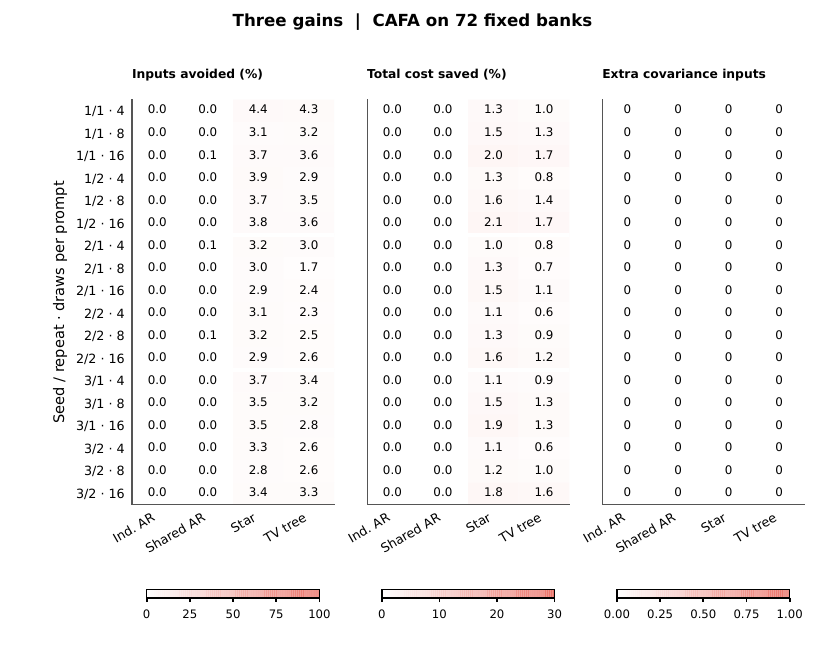}
\caption{Complete acquisition grid: three gains. Continuation of Figure~\ref{tab:review3-single-1} for this report: all 72 cells (every sampler, budget, seed and generation repeat), the same three panels ($100(1-\Qd/\Qc)$, the same-bank total-cost saving $100(1-\Td/\Tc)$, and the covariance increment $\Qd-\Qm$), the same shared colour scales, and the same CNY basis. The complete counts, fees and unrounded values accompany the figure in the CSV bundle.}\label{tab:review3-joint}\label{tab:review3-increment}
\end{figure}
\endgroup
The anchored report is omitted from this series because its scoring and cost reductions, and its covariance-support increment, are zero in every cell. Across all 72 banks, requesting all three gains together with the base mean restores the full cached support. The complete per-bank records are retained in the supplementary results. By Corollary~\ref{cor:report}, this report requires full cached support, so CAFA provides no scoring reduction.
\par
\subsubsection{Complete \texorpdfstring{$n=16$}{n=16} Workflow Cells}\label{app:grpo-cells}
Table~\ref{tab:grpo-cells} lists the 24 sampler--seed--repeat cells at 16 draws per prompt, with their generation and feedback costs and RMS MCSE.
\par
\begin{table}[!t]\centering\small\setlength{\tabcolsep}{5pt}
\caption{All $n=16$ GRPO comparisons, retaining each seed and generation repeat. $\Qc$, $\Qm$ and $\Qd$ count the inputs required by text caching, exact means, and means with covariance, respectively. $G$ is accepted full-process generation expenditure, using uninterrupted replays for failure-affected cells. $G+\Fd$ adds measured-token feedback cost for variance support. Monetary units are CNY at the frozen tariff.}\label{tab:grpo-cells}
\begin{adjustbox}{max width=\linewidth}
\begin{tabular*}{\linewidth}{@{\hspace{\tabcolsep}\extracolsep{\fill}}lllrrrrr}\toprule
Seed/repeat & Sampler & $\Qc$ & $\Qm$ & $\Qd$ & RMS MCSE & $G$ & $G+\Fd$\\\midrule
1/1 & Independent AR & 8165 & 8165 & 8165 & 0.0303 & 4.13 & 15.35\\
1/1 & Shared token & 7638 & 7634 & 7634 & 0.0258 & 4.21 & 14.84\\
1/1 & Independent star & 5394 & 5197 & 5197 & 0.0222 & 6.04 & 13.24\\
1/1 & TV-tree & 4872 & 4695 & 4695 & 0.0222 & 6.56 & 13.10\\
1/2 & Independent AR & 8166 & 8166 & 8166 & 0.0303 & 4.15 & 15.37\\
1/2 & Shared token & 7655 & 7654 & 7654 & 0.0251 & 4.21 & 14.87\\
1/2 & Independent star & 5354 & 5150 & 5150 & 0.0217 & 5.92 & 13.12\\
1/2 & TV-tree & 4901 & 4723 & 4723 & 0.0228 & 6.71 & 13.35\\
2/1 & Independent AR & 8172 & 8172 & 8172 & 0.0308 & 4.16 & 15.42\\
2/1 & Shared token & 7930 & 7929 & 7929 & 0.0265 & 4.11 & 15.11\\
2/1 & Independent star & 5751 & 5586 & 5586 & 0.0227 & 5.85 & 13.71\\
2/1 & TV-tree & 5267 & 5142 & 5142 & 0.0236 & 6.90 & 14.12\\
2/2 & Independent AR & 8175 & 8175 & 8175 & 0.0302 & 4.18 & 15.49\\
2/2 & Shared token & 7921 & 7921 & 7921 & 0.0264 & 4.18 & 15.20\\
2/2 & Independent star & 5801 & 5633 & 5633 & 0.0233 & 6.10 & 14.00\\
2/2 & TV-tree & 5310 & 5173 & 5173 & 0.0249 & 6.84 & 14.09\\
3/1 & Independent AR & 8172 & 8172 & 8172 & 0.0311 & 4.19 & 15.53\\
3/1 & Shared token & 7992 & 7992 & 7992 & 0.0266 & 4.20 & 15.35\\
3/1 & Independent star & 5638 & 5441 & 5441 & 0.0224 & 6.02 & 13.69\\
3/1 & TV-tree & 5144 & 5002 & 5002 & 0.0238 & 6.93 & 13.97\\
3/2 & Independent AR & 8171 & 8171 & 8171 & 0.0307 & 4.28 & 15.61\\
3/2 & Shared token & 7987 & 7986 & 7986 & 0.0266 & 4.19 & 15.33\\
3/2 & Independent star & 5731 & 5539 & 5539 & 0.0231 & 6.37 & 14.14\\
3/2 & TV-tree & 5135 & 4965 & 4965 & 0.0231 & 6.88 & 13.87\\
\bottomrule\end{tabular*}
\end{adjustbox}\end{table}

\subsubsection{Grid Completion and Uninterrupted Timing}\label{app:completion-timing}
The completed comparison grid is four methods $\times$ three budgets ($n=4,8,16$) $\times$ three training seeds $\times$ two generation repetitions, for 72 banks. The six Tree $n=4$ cells are inside that grid, not additional to it: they use the original exports and protocol, the same score-free pilot rule and the frozen evaluation judge, and their pilot, tree edges, copy decisions, residual acceptance and probability replay were independently reconstructed. No new training, reference generation or replay scoring was required for them.
\par

Their per-cell support counts, RMS MCSE and costs are part of the complete grid, retained in the supplementary canonical acquisition table with every other cell.
\par

Which cells use replayed rather than original timings, the replay hardware and scope, the identity checks, and why the historical-minus-replay difference is not a failure-only estimate are all recorded once, in Appendix~\ref{app:G-history}. When reading the cost curves of this appendix, the reported generation costs use uninterrupted measurements for the affected cells.
\par

\subsection{Nearby and Far-End Monitoring}\label{app:D-monitoring}
All 144 test-monitoring cells (72 banks $\times$ the base--10 and base--250 targets) are retained, positive, negative and inconclusive alike. One resampling protocol applies to all of them: the original 64 test prompts with their source counts (32 ELI5, 26 WildChat, six No Robots) held fixed, 5,000 within-source prompt resamples with seed 20260914, the same prompt resamples reused across banks, and independently generated samplers never treated as paired root draws. Training seeds and generation repetitions stay separate and are never pooled into an interval.
\par

For each fixed-budget report, an interval above zero is labelled ``$+$'' (evidence of improvement), one below zero ``$-$'' (evidence of decline) and one crossing zero ``?'' (inconclusive). These labels are exploratory: they condition on the collected generations, they are not simultaneous across the budget grid, they are not calibrated real-model coverage, and neither counting them nor choosing a favourable budget estimates a training success rate. An earlier 2,000-resample unstratified analysis of the test-prompt means is a separate, historical procedure reported in the interpretation block of Appendix~\ref{app:D-cells}.
\par

\subsubsection{Nearby Monitoring: Savings and Resolution Together}\label{sec:nearby-monitoring}
This reanalysis fixes the base--step-10 contrast and follows the monitoring protocol of Appendix~\ref{app:D-monitoring}. The six star cells are shown with the independent reference at 64 draws per policy in the base--10, $n=16$ panel of Figure~\ref{tab:review3-monitor-all-panel-3}. That reference is itself estimated from a finite sample and is not a known~truth.
\par

Four of the six star intervals lie above zero and two cross zero. All three independent-reference intervals cross zero. These six Star cells represent two generation repetitions for each of three training seeds. The independent references are also finite-sample estimates. CAFA gives the same estimates and intervals as full scoring. Its contribution is the feedback avoided in reaching that identical report.
\par

\paragraph{Test-Set Monitoring at Every Recorded Budget.}
Figures~\ref{tab:review3-monitor-all}--\ref{tab:review3-monitor-all-panel-6} retain both the nearby and the far-end contrast at every recorded budget, under the protocol and interval labels of Appendix~\ref{app:D-monitoring}. The first three figures show base--10 and the last three show base--250 at 4, 8 and 16 draws per prompt. Read each gain against the zero line and its prompt-bootstrap interval, keeping training seeds and generation repetitions separate. The acquisition heatmaps show the corresponding 128-prompt reductions. Exact costs are retained in their CSV files. The separate CSV also reports test-only support and score fees, reference uncertainty, and paired-prompt bootstrap differences from the reference. It does not infer half of the generation charge from the test split.
\par
\newpage
\enlargethispage{\dimexpr-4\baselineskip-\parskip\relax}
\begingroup\scriptsize\setlength{\tabcolsep}{3pt}
\begin{figure}[!htb]\centering\normalsize
\includegraphics[trim=24.19bp 3.8bp 3.52bp 1.7bp,clip,width=0.930\linewidth]{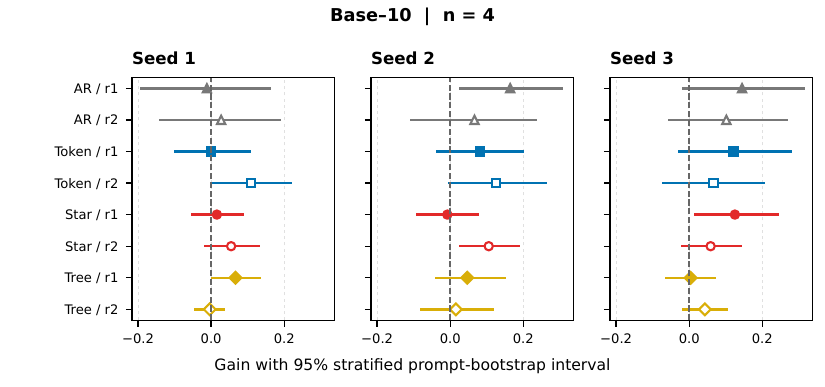}
\caption{Monitoring across all recorded budgets. Each panel set contains all 24 sampler--seed--repeat cells for the stated target and budget. Points are gains on the 64 test prompts. Intervals are the original 95\% source-stratified prompt-bootstrap intervals (5,000 resamples). Open markers indicate generation repeat 2. The zero line separates positive and negative gains. Crossing it is inconclusive. Fixed-prompt MCSE and all support/fee fields remain separately available in the CSV. These exploratory intervals are not simultaneous or anytime-valid guarantees.}\label{tab:review3-monitor-all}
\end{figure}
\begin{figure}[!htb]\centering\normalsize
\includegraphics[trim=24.19bp 3.8bp 3.52bp 1.7bp,clip,width=0.930\linewidth]{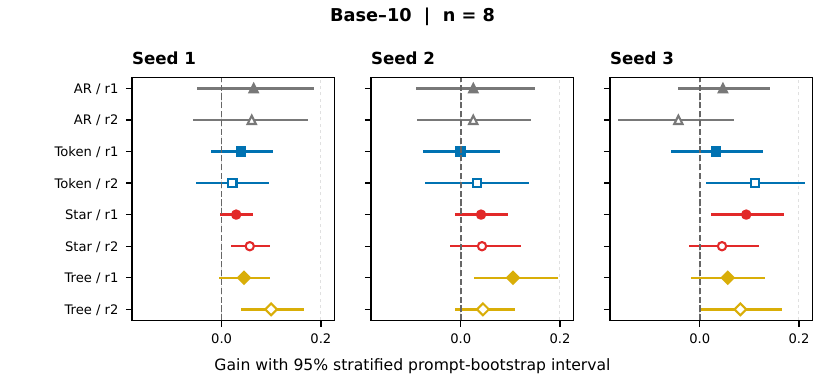}
\caption{Monitoring: base--10, $n=8$. Continuation of Figure~\ref{tab:review3-monitor-all}, with the same definitions and preserved per-cell values.}\label{tab:review3-monitor-all-panel-2}
\end{figure}
\begin{figure}[!htb]\centering\normalsize
\includegraphics[trim=12.65bp 3.72bp 3.52bp 1.66bp,clip,width=0.959\linewidth]{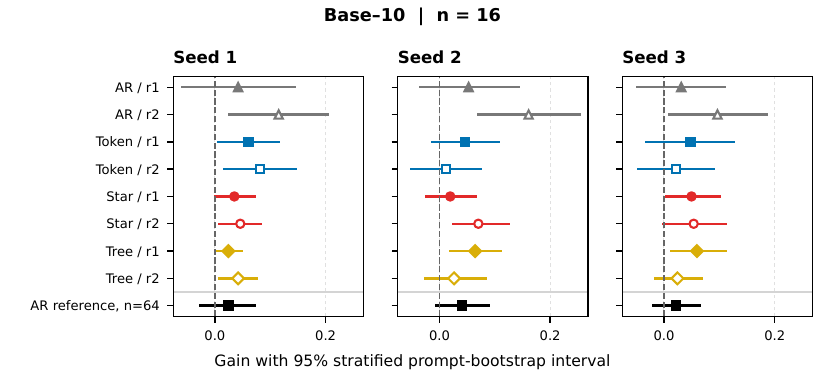}
\caption{Monitoring: base--10, $n=16$, with the independent reference. Continuation of Figure~\ref{tab:review3-monitor-all}, with the same definitions and preserved per-cell values. Below the horizontal rule, each panel includes the independent autoregressive reference for that seed, using 64 draws per policy on the same 64 test prompts with the same frozen judge. The reference is estimated from a finite sample, is separately uncertain, and is not a known~truth. Its own interval is in Table~\ref{tab:review3-reference}.}\label{tab:review3-monitor-all-panel-3}
\end{figure}
\begin{figure}[!htb]\centering\normalsize
\includegraphics[trim=24.19bp 3.8bp 3.52bp 1.7bp,clip,width=0.930\linewidth]{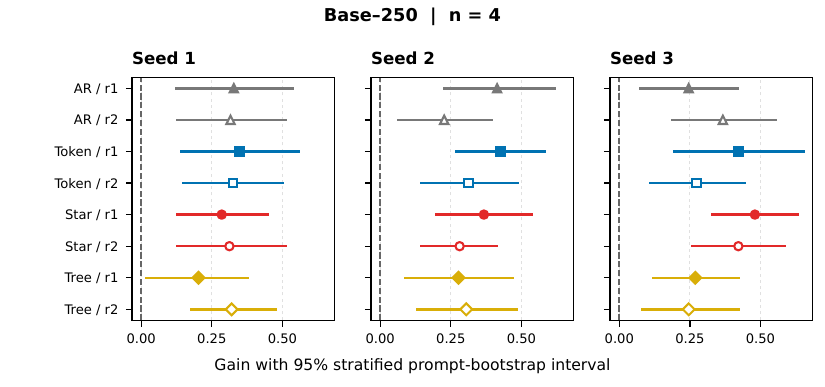}
\caption{Monitoring: base--250, $n=4$. First of the three far-end panels, continuation of Figure~\ref{tab:review3-monitor-all}, with the same definitions and preserved per-cell values.}\label{tab:review3-monitor-all-panel-4}
\end{figure}
\begin{figure}[!htb]\centering\normalsize
\includegraphics[trim=24.19bp 3.8bp 3.52bp 1.7bp,clip,width=0.930\linewidth]{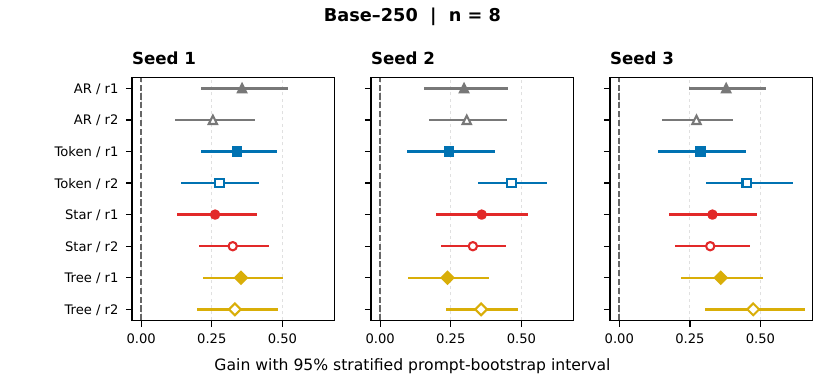}
\caption{Monitoring: base--250, $n=8$. Continuation of Figure~\ref{tab:review3-monitor-all}, with the same definitions and preserved per-cell values.}\label{tab:review3-monitor-all-panel-5}
\end{figure}
\begin{figure}[!htb]\centering\normalsize
\includegraphics[trim=24.19bp 3.8bp 3.52bp 1.7bp,clip,width=0.930\linewidth]{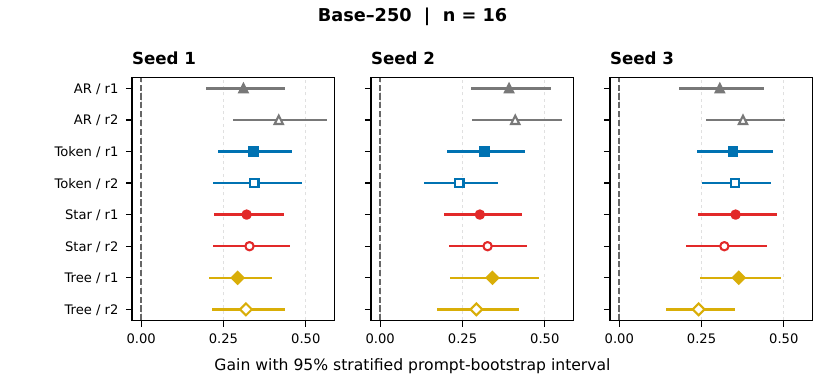}
\caption{Monitoring: base--250, $n=16$. Continuation of Figure~\ref{tab:review3-monitor-all}, with the same definitions and preserved per-cell values.}\label{tab:review3-monitor-all-panel-6}
\end{figure}
\endgroup

\subsection{Independent References and Checkpoint Selection}\label{app:D-selection}
Checkpoints are selected using 64 selection prompts and evaluated on 64 separate test prompts. Three independent-AR reference banks, one per training seed with 64 draws per test prompt and policy (49,152 answers in total), estimate test quality, as reported in Tables~\ref{tab:grpo-quality} and~\ref{tab:review3-reference}. Of the 72 fixed selectors, 71 agree with the estimated reference maximizer. The exception occurs for independent AR with seed 2, eight draws and generation repetition 2. It selects step 50, with estimated regret 0.151 and fixed-pair MCSE 0.020. CAFA preserves every selection made from the original bank. It does not eliminate sampling error relative to an independently estimated reference. The reference maximizer is not a known population optimum, and its uncertainty remains part of the interpretation. Deployment accounting charges the complete 128-prompt workflow rather than inferring half its generation cost from the selection subset.
Pairwise reference MCSE conditions on the estimated~maximizer.
\par

\begin{table}[!htbp]\centering\small
\caption{Independent test-reference quality (mean $\pm$ Monte Carlo standard error), by training seed. Training seeds and generation repetitions are not pooled. The base samples use independent reference namespaces across seeds.}\label{tab:grpo-quality}
\begin{adjustbox}{max width=\linewidth}
\begin{tabular*}{\linewidth}{@{\hspace{\tabcolsep}\extracolsep{\fill}}lrrrr}\toprule
Training seed & Base & Step 10 & Step 50 & Step 250\\\midrule
1 & $3.034\pm0.016$ & $3.058\pm0.015$ & $3.200\pm0.014$ & $3.350\pm0.014$\\
2 & $3.021\pm0.016$ & $3.061\pm0.015$ & $3.219\pm0.014$ & $3.371\pm0.014$\\
3 & $3.046\pm0.015$ & $3.068\pm0.015$ & $3.193\pm0.014$ & $3.367\pm0.014$\\
\bottomrule\end{tabular*}
\end{adjustbox}\end{table}

Complete per-seed and per-repetition counts are retained in Table~\ref{tab:grpo-cells}.
\par
\begin{table}[!htb]\centering\small\setlength{\tabcolsep}{3pt}
\caption{Independent references under the same 5,000-resample stratified test protocol, 64 draws per policy and test prompt. This reanalysis is separately labeled from earlier unstratified exploratory intervals.}\label{tab:review3-reference}
\begin{adjustbox}{max width=\linewidth}
\begin{tabular*}{\linewidth}{@{\hspace{\tabcolsep}\extracolsep{\fill}}llrrl}\toprule
Seed & Target & Gain & MCSE & 95\% interval\\
\midrule
1 & Base--10 & 0.0247 & 0.0216 & [-0.0269, 0.0725]\\
1 & Base--250 & 0.3167 & 0.0210 & [0.2285, 0.4160]\\
2 & Base--10 & 0.0408 & 0.0216 & [-0.0049, 0.0884]\\
2 & Base--250 & 0.3501 & 0.0210 & [0.2554, 0.4573]\\
3 & Base--10 & 0.0222 & 0.0214 & [-0.0200, 0.0642]\\
3 & Base--250 & 0.3206 & 0.0207 & [0.2427, 0.4102]\\
\bottomrule
\end{tabular*}
\end{adjustbox}
\end{table}

\subsection{Report Preservation}\label{app:D-preservation}
Every one of the 504 report cells reproduces the fully scored mean and Monte Carlo covariance from the acquired support, and all 72 fixed checkpoint selections are unchanged. Mean and draw supports coincide in 485 cells.
\par

The other 19 (13 independent-AR and three shared-token cells for base--10, two independent-AR cells for 10--50, one independent-AR cell for base--50) need extra scores for the~covariance.
\par

For example, independent AR with seed 1, repetition 2 and $n=16$ requires 4065 scores for the base--10 mean and 4071 to recover both the mean and ordinary covariance of that contrast. The three-gain report requires 8166 scores in either reporting mode.
\par

\paragraph{Independent Recomputation.}
For this paper the 504 cells were recomputed a second time by a separate implementation of Algorithm~\ref{alg:acquire} written without reference to the analysis code. For each of the 72 frozen banks, the implementation verified the SHA-256 hash of the bank file against the recorded hash, planned every target without reading scores and acquired cached scores through a counting scorer. It then compared the recovered statistics with the published results and an independent dense~computation.
\par

All 504 cells matched exactly in cached, mean and mean-plus-covariance support counts, and scorer-call counts equalled the required counts in every cell. Mean differences were exactly 0, and covariance differences were at most $4.3\times10^{-19}$. The full covariance matrices are retained in the supplementary tables. The fixed selections are shown below.
\par

\paragraph{Selection and Covariance Are Different Operations.}
Figure~\ref{tab:review3-selection-covariance} retains all 72 fixed selection decisions and their estimated independent-reference regret. A selected reference maximizer has zero empirical regret by construction, the one nonzero regret retains its original fixed-pair MCSE of 0.020, and no pooled coverage or selection-error guarantee is inferred from these cells. The joint-gain Monte Carlo covariance of every bank is a separate matter: CAFA preserves it exactly within a bank, a different sampler can change both it and the selected checkpoint, and the complete per-bank matrices at every budget are retained in the supplementary canonical tables.
\par
\newpage
\enlargethispage{\dimexpr-\baselineskip-\parskip\relax}

\begingroup\scriptsize\setlength{\tabcolsep}{3pt}
\begin{figure}[!htb]\centering\normalsize
\captionsetup{skip=3pt}
\includegraphics[trim=6.05bp 0bp 5.5bp 0.32bp,clip,width=0.860\linewidth]{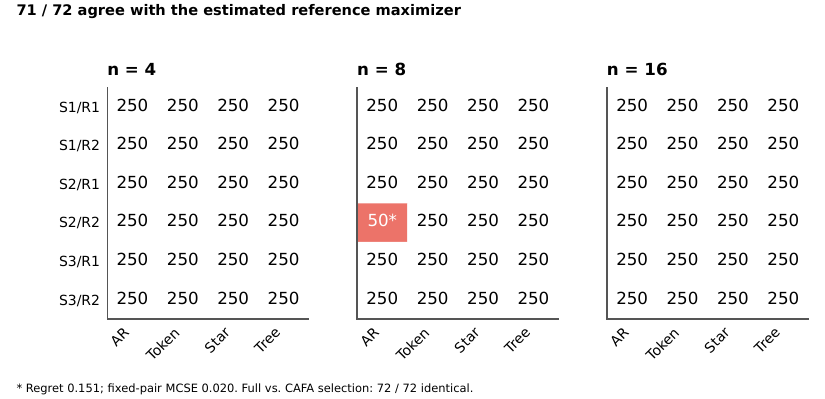}
\caption{Checkpoint selection on every frozen bank. The matrix gives the saved checkpoint step selected in each of the 72 banks and highlights the sole disagreement with the estimated independent reference, whose estimated regret is 0.151 with fixed-pair MCSE 0.020: 71 of 72 selectors agree with that reference, and full scoring and CAFA make identical selections in all 72. Regret is measured against an estimated maximizer, not a known optimum.}\label{tab:review3-selection-covariance}
\end{figure}
\endgroup

\section{Report Targets and Generation Configurations}\label{app:targets}

\subsection{Single, Joint, Adjacent and Anchored Reports}\label{app:E-reports}
The same frozen bank can support different reports, but the support must be recomputed for each report. Required counts cannot be inferred by dividing a joint count by the number of outputs or by summing counts across reports. Table~\ref{tab:targets} gives all seven online-GRPO outputs over the six independent-star $n=16$ banks. Their per-cell counts and six-cell ranges are in the supplementary canonical tables, and every GRPO target is visualized cell by cell in Appendix~\ref{app:D-cells}. Adjacent-checkpoint contrasts reuse the original joint samples. A separately generated adjacent-pair coupling would be a different experiment. Adding the absolute mean of the base model to the gains changes the report and restores the complete cached support, as established in Corollary~\ref{cor:report}. It is not a change of coordinates. Averaged over all four independent-star repetitions on the strict holdout, the required input count increases from 1193.25 to 2700.00 for Qwen and from 212.75 to 2157.50 for SmolLM2.
\par


\subsubsection{Simple Skipping, Full Aggregation, and Additional Outputs}\label{sec:target-study}
The CAFA interface exposes three stages: cache equal scoring inputs, skip zero differences within each draw and cache the remaining inputs, and aggregate coefficients across the complete batch. Whether the last two coincide on the completed strict-holdout and distance banks is measured in Appendix~\ref{app:C-mechanism}, and Proposition~\ref{prop:signpure} supplies the structural explanation for the star construction while that audit checks decoded identities.
\par
\newpage
\enlargethispage{\dimexpr-\baselineskip-\parskip\relax}

All-gains plus an anchor identifies every policy mean, so the return to the cached support noted above follows from the requested target. Source-specific gain reporting continues to use zero-sum contrasts within each source, so requesting a per-source breakdown does not itself reintroduce an anchor.
\par

\begin{table}[!t]
\centering\small
\caption{Reusing the frozen distance bank for different requested outputs and output caps. Qwen 1.5B is seed 202609084. Qwen 7B is its single trajectory. All rows use 32 prompts, 64 root draws per prompt and independent-acceptance star sampling, and every row recomputes its own support from the same cached responses for its cap. The individual rows request one contrast each, so their counts must not be added to reach a joint count. Gains use DeepSeek Flash. ``--'' means no single scalar gain is defined for that report (vector and absolute-anchor rows). ``n/r'' means the source record for those four 256-token rows reports support only, and no gain is imputed for them. Adjacent differences use the original joint samples, including their covariance.}\label{tab:revision2-targets}
\begin{adjustbox}{max width=\linewidth}
\begin{tabular*}{\linewidth}{@{\hspace{\tabcolsep}\extracolsep{\fill}}llrrrrl}\toprule
Model & Cap & Requested output & Cached & Required & Removed (\%) & Gain\\\midrule
Qwen 1.5B & 768 & Base to 2 & 2223 & 358 & 83.9 & -0.002\\
Qwen 1.5B & 768 & Base to 8 & 2477 & 865 & 65.1 & 0.025\\
Qwen 1.5B & 768 & Base to 32 & 3363 & 2636 & 21.6 & 0.121\\
Qwen 1.5B & 768 & 2 to 8 & 2556 & 1023 & 60.0 & 0.027\\
Qwen 1.5B & 768 & 8 to 32 & 3389 & 2684 & 20.8 & 0.096\\
Qwen 1.5B & 768 & All gains & 3975 & 3274 & 17.6 & --\\
Qwen 1.5B & 768 & Gains + base mean & 3975 & 3975 & 0.0 & --\\
Qwen 7B & 768 & Base to 2 & 2215 & 334 & 84.9 & 0.007\\
Qwen 7B & 768 & Base to 8 & 2620 & 1144 & 56.3 & 0.028\\
Qwen 7B & 768 & Base to 32 & 3935 & 3774 & 4.1 & 0.080\\
Qwen 7B & 768 & 2 to 8 & 2701 & 1306 & 51.6 & 0.021\\
Qwen 7B & 768 & 8 to 32 & 3939 & 3782 & 4.0 & 0.052\\
Qwen 7B & 768 & All gains & 4674 & 4525 & 3.2 & --\\
Qwen 7B & 768 & Gains + base mean & 4674 & 4674 & 0.0 & --\\
\addlinespace
Qwen 7B & 256 & Base to 2 & 2173 & 250 & 88.5 & n/r\\
Qwen 7B & 256 & Base to 8 & 2475 & 854 & 65.5 & n/r\\
Qwen 7B & 256 & Base to 32 & 3777 & 3458 & 8.4 & n/r\\
Qwen 7B & 256 & All gains & 4329 & 4027 & 7.0 & --\\
\bottomrule\end{tabular*}
\end{adjustbox}
\end{table}

Table~\ref{tab:revision2-targets} compares base-relative and adjacent-checkpoint targets using the same generated bank. An adjacent contrast can drop inputs needed only by the base or another endpoint. The result measures the value of requesting a different output from existing samples. A newly generated adjacent-pair coupling would be a separate experiment. The machine-readable results include both output caps, all available trajectories, separate mean and variance supports, and both cached judges.
\par

\subsection{Checkpoints, Model Configurations and Stopping Caps}\label{app:E-configs}
These results use the offline distance regime of Table~\ref{tab:regimes}: three saved trajectories per small model, one for Qwen2.5-7B, checkpoints after 2, 8 and 32 steps, and two output caps. Step counts are checkpoint labels, not measured distances, and the 1.5B and 7B configurations are not a controlled model-size comparison. The 7B bank contains an independent-star and a shared-acceptance arm on one trajectory. The figures quoted for 7B in the main text are the independent-star arm. A length cap defines a stopped-response policy: at a 256-token cap, 99.0\% of the 7B independent-star endpoint slots (copies included) stop at the cap, versus 9.7\% at 768 tokens, so 768 is the primary configuration and 256 a supplementary one.
\par

\subsubsection{Checkpoint and Output-Cap Sensitivity}\label{sec:cost}
The prompt sets and their exclusions are in Appendix~\ref{app:B-provenance}, the samplers and strong baselines in Appendix~\ref{app:B-samplers}, and the cost conventions in Appendix~\ref{app:cost}. We examine how scoring reductions vary with the checkpoint, output cap and model: a distance study over checkpoints at 2, 8 and 32 update steps under 256- and 768-token output limits, and a 7B extension at a larger generation cost. Generation and probability-replay verification are complete for both 7B output limits and for all eight strong-baseline jobs on the comparison bank, and unavailable provider scores propagate through the reported~bounds.
\par
\newpage
\enlargethispage{\dimexpr-2\baselineskip-\parskip\relax}

\paragraph{Model and Output Configurations Change Accessible Savings.}
Table~\ref{tab:revision2-targets} evaluates Qwen2.5-7B with checkpoints after 2, 8 and 32 update steps at both output caps, alongside the 1.5B rows. For the joint target under independent star sampling, CAFA removes \AutoSevenShortRemoved{} of \AutoSevenShortBefore{} unique inputs at a 256-token cap and \AutoSevenLongRemoved{} of \AutoSevenLongBefore{} at a 768-token cap. Thus large savings for nearby updates do not imply equally large savings for a collection that also includes a more distant update. The contrast target itself, alongside policy overlap and response length, determines the accessible saving. Figure~\ref{fig:currentscale} compares individual checkpoint gains with the joint report to show how this saving changes across the saved checkpoints. Each output cap defines a stopped-response policy. Figure~\ref{tab:truncation} reports how often endpoints reach that cap, Table~\ref{tab:revision2-targets} carries both caps, and Figure~\ref{fig:update-gains} gives the quality change these supports are bought to measure.
\par

For the joint target, shared acceptance requires \AutoSevenShortShared{} and \AutoSevenLongShared{} inputs at the two caps, versus \AutoSevenShortAfter{} and \AutoSevenLongAfter{} for independent acceptance. The sign of this realized increment changes across the two conditions, whereas coefficient cancellation preserves the estimate in either case. This distinguishes a general CAFA rule from a coupling choice whose practical advantage depends on the generated support. The 7B experiment evaluates one saved update trajectory at both response caps, connecting the removable support to quality changes on that trajectory. Training steps index the saved checkpoints rather than measured distribution distances, and the 1.5B and 7B runs use different training configurations.
\par
\newpage
\begingroup
\setlength{\intextsep}{6pt}
\begin{figure}[H]
\centering\captionsetup{skip=3pt}
\includegraphics[trim=6.26bp 10.15bp 3.52bp 4.75bp,clip,width=0.760\linewidth]{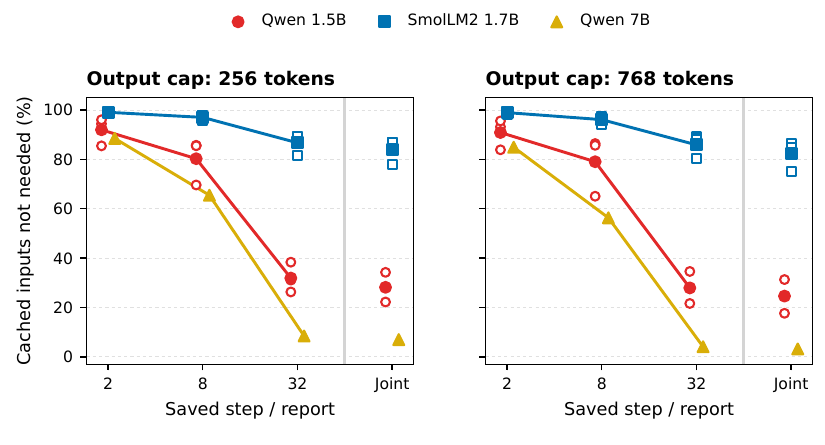}
\caption{The requested comparison determines the saving. Independent Star + CAFA at 64 roots per prompt and 32 prompts. Steps 2, 8 and 32 are checkpoint labels, not measured distribution distances, and request individual base--update gains. ``Joint'' requests all three gains from the same draws. Hollow points retain individual trajectories and filled points show their means. The 7B extension has one trajectory. Output limits define different generation distributions.  Individual query counts must not be added to infer the joint cost.}\label{fig:currentscale}
\end{figure}
\begin{figure}[H]\centering\normalsize
\captionsetup{skip=3pt}
\includegraphics[trim=7.14bp 5.74bp 3.52bp 2.17bp,clip,width=0.760\linewidth]{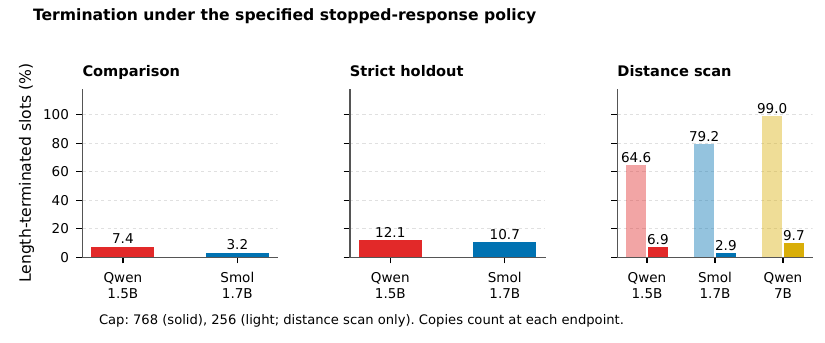}
\caption{Termination at the frozen output caps. The denominator is endpoint policy--root slots, including copied answers at every represented policy. Solid bars use cap 768. Light bars use cap 256 in the distance scan. Comparison and holdout banks have four star repetitions. The distance scan has three trajectories per small model and one for Qwen 7B. These are stopped-response policies, not uncapped completions. The original exact displayed cells and source counts are included in the CSV bundle.}\label{tab:truncation}
\end{figure}
\endgroup

\subsection{Temperature and Deterministic Generation}\label{app:E-temperature}
Lower temperature is a different generation policy, not a cheaper way to evaluate the same one: greedy decoding has no Monte Carlo repetition, and temperature 0.3 changes both the agreement between policies and the estimand. The study freezes the same 32 strict-holdout prompts and the base plus three Qwen 1.5B updates. At $T=0.3$, independent AR, shared-token randomness and independent-acceptance star each use 16 draws per prompt and two independent generation repetitions. The $T=1$ reference reuses the first 16 draws of the first two original repetitions. Greedy decoding uses one response per policy and prompt, 128 in total.
\par

Every condition keeps the original 768-token cap, EOS, chat template and FP32 execution, and temperature is applied identically in generation and in cross-policy density replay. The costs in Table~\ref{tab:temperature-cost} come from standalone generation runs. The break-even column is a threshold feedback price, not a measured one, and no judge was called for it.
\par
\newpage
\subsubsection{Does Ordinary Caching Suffice at Lower Temperature?}\label{sec:temperature}
\begin{figure}[!htb]\centering\normalsize
\includegraphics[trim=6.24bp 0bp 2.01bp 0bp,clip,width=0.979\linewidth]{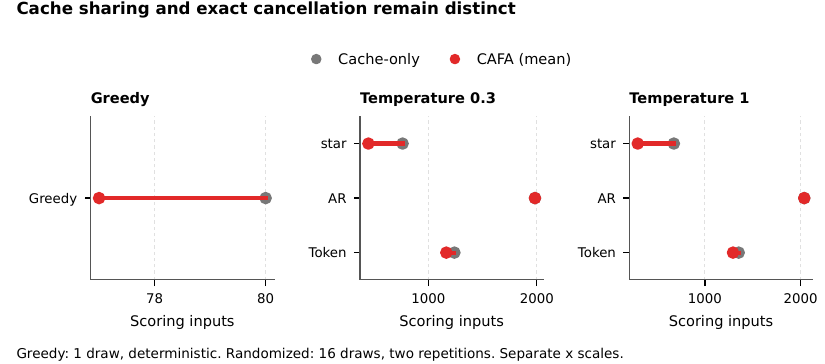}
\caption{Cache sharing and cancellation at different temperatures. Gray points show cached inputs. Red points show mean support. Randomized policies use 16 draws per prompt and two repetitions. Greedy uses one deterministic draw and has no Monte Carlo replication. Each temperature defines a different response policy. Axes are separate so that the deterministic case remains readable. At $T=1$, independent AR needs 2038.5 inputs for the mean versus 2039 for the mean with covariance. Both supports and the complete original columns remain in CSV.}\label{tab:temperature}
\end{figure}

Even with deterministic decoding, cache sharing and contrast cancellation are distinct operations: the greedy bank has 80 cached inputs and needs 77 scores for the requested gains. Greedy policies have a unique joint response vector, so changing the coupling cannot improve their overlap. Contrast cancellation can still remove scores. Lower-temperature comparisons therefore measure both the sharing achieved by ordinary generation and the extra queries removed by the target. Each temperature defines its own response policy. Figure~\ref{tab:temperature} measures CAFA requirements for that policy without needing a judge score.
\par

\begin{table}[H]
\centering\small
\caption{New $T=0.3$ standalone sampler runs, CNY per repetition at 3 CNY per card-hour. Each arm runs in a fresh process with its own residual streams, including loading, generation, probability replay and saving. Thresholds compare with independent AR plus caching. Actual judge calls are zero.}\label{tab:temperature-cost}
\begin{adjustbox}{max width=\linewidth}
\begin{tabular*}{\linewidth}{@{\hspace{\tabcolsep}\extracolsep{\fill}}lrrrr}\toprule
Sampler & Generation & $\Delta G$ & $\Delta L$ & CNY/1k labels\\\midrule
Shared tokens & 0.916 & 0.011 & 817.0 & 0.013\\
Ind. AR & 0.905 & 0.000 & 0.0 & --\\
Ind. star & 1.395 & 0.490 & 1535.0 & 0.319\\
\bottomrule\end{tabular*}
\end{adjustbox}
\end{table}

The supplementary result package reports both repetitions, each individual gain, response lengths, truncation, proposal counts and per-prompt support. The protocol and technical canary were frozen before the new outputs. Neither query savings nor judge values were used to select prompts, checkpoints or a stopping time.
\par
\newpage

\section{Scoring and Statistical Inference}\label{app:inference}

\subsection{Versioned Judge Sensitivity}\label{app:F-judges}
This paper reports four judge evaluations with different coverage, which are analyzed separately. (i) Every main-text GRPO result uses one version-pinned DeepSeek Flash judge, whose returned fingerprint was checked on every call. There is no multi-judge re-scoring of the online study. (ii) The controlled studies score the same responses with DeepSeek Flash and Qwen Flash in separate caches, covering all 32 strict-holdout prompts. (iii) New Qwen Max calls are allocated to a fixed stratified subset of four prompts per bank, which is an audit, not full-bank coverage.
\par

(iv) A historical version-matched three-judge comparison is retained under its original identities, on 31--32 matched prompts per cell. The current paired comparison and the historical broad-coverage one differ in coverage and are reported separately. A separate agreement check compares Flash scores with previously collected Max scores: of 500 previously Max-rated inputs, 499 returned valid Flash scores, with Spearman correlation 0.937 and mean absolute difference 0.339 on the ten-point scale. That is an operational agreement screen. None of these checks validates agreement with human preference, and a pinned model name cannot rule out a silent backend change where no fingerprint is returned.
\par

\subsubsection{Sensitivity to the Scoring Function}\label{sec:judges}
\begin{figure}[!htbp]
\centering\includegraphics[trim=7.56bp 3.27bp 3.52bp 2.38bp,clip,width=0.972\linewidth]{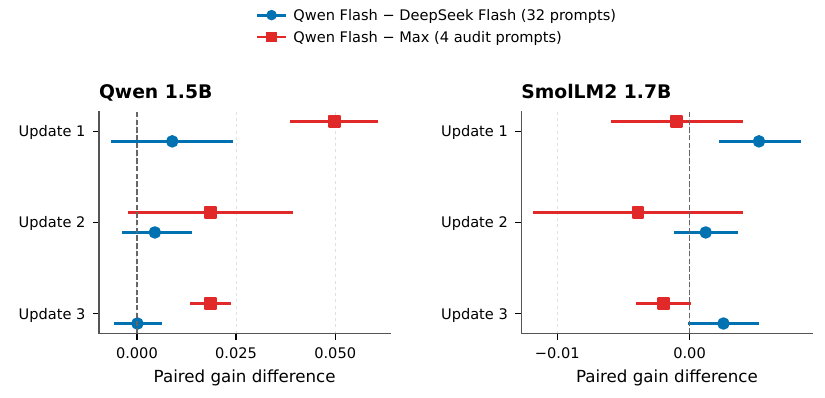}
\caption{Judge comparisons preserve input pairing. Shared-star strict holdout, $n=64$, four generation repetitions. Flash--Flash comparisons use all 32 prompts. Flash--Max uses the four fixed audit prompts. Points show differences between the gain estimates from the two judges on identical inputs. Bars are 95\% prompt-bootstrap intervals within the frozen sampling strata, conditional on collected generations. The small subset audited with Max does not provide full-bank coverage. An interval containing zero does not establish~equivalence.}\label{fig:currentjudges}
\end{figure}

\begin{figure}[!t]
\centering\includegraphics[trim=3.94bp 3.99bp 5.5bp 4.28bp,clip,width=0.879\linewidth]{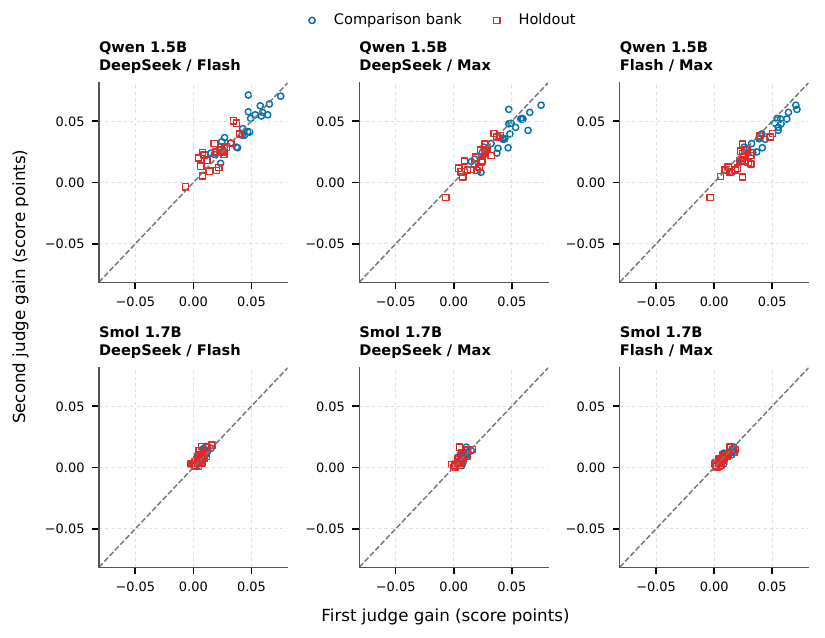}
\caption{Paired update gains under three scoring functions at $n=64$. Each marker denotes a fixed checkpoint, sampler, generation repetition, and judge pair, averaged over their common complete prompts. The dashed line is equality. Columns separate judge pairs. Rows separate model families. Blue and red distinguish the two prompt banks. The two methods and four budgets reuse generated data.}\label{fig:judges}
\end{figure}

The current strict-holdout comparison is Figure~\ref{fig:currentjudges}. The diagnostics below instead refer to the historical version-matched analysis of scope~(iv) above, whose incomplete cells are excluded from both judges with no score imputation. Reconstructing its per-sample differences reproduces all 28,704 available versioned per-prompt means exactly. For each fixed checkpoint, we compute the variance of the paired judge difference directly. Across 1,152 checkpoint--sampler--budget--repetition--judge-pair cells, the median ratio of this standard error to the value obtained under independent judge estimates is 0.451. These cells share data and are not independent replications. At $n=64$, median within-prompt correlations on non-copy samples range from 0.858 to 0.895 across judge pairs. Identical base and update responses contribute zero under every judge. Excluding them from this correlation diagnostic avoids crediting the judges for algebraically forced agreement. Figure~\ref{fig:judges} shows the paired gains. Of 288 plotted cells, 30 have at least one gain of magnitude at least 0.05 score points. None of these has opposite signs. Across all four budgets, the corresponding count is 131 of 1,152. The 0.05 band is a descriptive threshold for summarizing sign agreement. Most gains are small, and this check does not establish that near-tied updates are correctly ordered. Uncertainty depends on the target population. On the independent-acceptance star arm at $n=64$, 13 of 36 fixed-prompt repetition-bootstrap intervals exclude zero, compared with 5 of 36 source-stratified prompt-bootstrap intervals. A crossed resampling of prompts and generation repetitions gives 0 of 36, as detailed in Appendix~\ref{app:judges}. We report all three as sensitivity analyses.
\par

\paragraph{Versioned Feedback and Strong-Judge Audit.}
New full evaluations run with thinking disabled and separate caches per judge. Version identity is the returned model name together with the system fingerprint where one is returned. Historical aliases retain their original identities and are never renamed retrospectively, which is what keeps the historical analysis of Appendix~\ref{sec:judges} separable from the current one. For shared-star sampling at $n=64$ on the strict holdout, the largest absolute Flash--Flash difference across the three fixed update gains is 0.0089 score points on Qwen and 0.0052 on SmolLM2. The largest Flash--Max difference on the four fixed audit prompts is 0.0498. These numerical gaps are small on the ten-point scale, but some prompt-bootstrap intervals exclude zero and the comparison does not establish judge equivalence. Judge scores affect estimated gains and precision. The acquired-input counts depend only on the frozen responses and contrast coefficients.
\par
\newpage
\enlargethispage{\dimexpr-5\baselineskip-\parskip\relax}

\subsubsection{Judge Comparison Units and Coverage}\label{app:judges}
All judge comparisons condition on the saved policy endpoints. With $D_{it}^{(a)}$ and $D_{it}^{(b)}$ denoting same-sample base--update score differences, the within-prompt variance estimate is $s^2(D^{(b)}-D^{(a)})/n$, including covariance. Prompt averages use equal weights over the common complete cells. Version identity includes the returned model and, when available, the system fingerprint. Absent Qwen fingerprints leave model-name pinning weaker than full backend pinning.
\par

The independent-acceptance star analysis at $n=64$ contains 36 population-comparison cells: two prompt sets, two model families, three fixed checkpoints, and three judge pairs. Each uses four generation repetitions and 31--32 common prompts. Ten thousand bootstrap draws use the same selected indices for both judges. The fixed-prompt scheme resamples repetition rows. The prompt scheme averages repetitions first and resamples prompt columns within source strata. The combined scheme independently resamples the row and stratified column indices. It is a crossed row--column bootstrap, not a nested bootstrap over new training runs. These are descriptive uncertainty checks for different conditioning choices.
\par

The strong-judge cache covers substantially more than the new four-prompt audit allocation, but is not literally complete: six stage--job--budget--judge-pair combinations lose one checkpoint--sampler--prompt cell (191 of 192 retained). The missing cell is removed symmetrically from both judges. Source snapshots and extraction manifests retain those exclusions. Comparisons against tree sampling use matched prompts, but independently generated streams. Assigning the same repetition number to two runs does not create common-random-number pairing.
\par
\newpage
\enlargethispage{\dimexpr-3\baselineskip-\parskip\relax}

\subsection{Known-Truth Finite-Proxy Error}\label{app:F-proxy}
To measure error against known means, this study defines finite-support proxy policies over 32 stored responses for each of 16 prompts, with two historical fixed scoring functions. The base is uniform on the response indices, endpoint probabilities are proportional to the stored likelihood ratios, and every cached score is known, so the expected scores are exact. Five samplers---independent categorical, shared categorical Gumbel, independent star, shared acceptance and a total-variation spanning tree---receive the same aggregation and deduplication over 2,000 Monte Carlo repetitions at 8 to 128 draws per prompt, alongside a full-pool exact evaluation that removes Monte Carlo error entirely and makes finite-cache saturation visible.
\par

The proxies are not the unrestricted language models. Exactness at the full pool is exactness for a finite constructed distribution, not for any finite budget on a real model.
\par

\subsubsection{Does the Reported Answer Change?}\label{sec:answer}
\begin{figure}[!htbp]
\centering\includegraphics[trim=20.1bp 0.27bp 5.5bp 4.29bp,clip,width=0.935\linewidth]{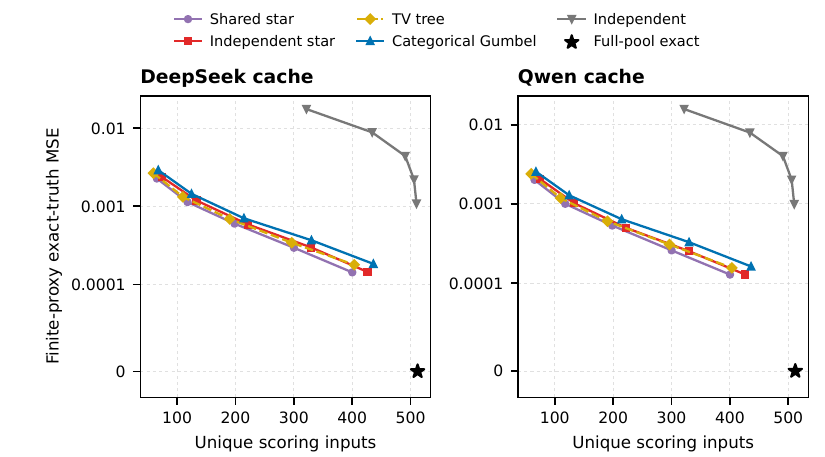}
\caption{Cost--error diagnostic for finite-support proxy policies on 16 prompts with 32 stored responses each, using two historical fixed scoring functions. Each sampled budget uses 2,000 Monte Carlo repetitions, with 8--128 draws per prompt. The vertical scale is symmetric logarithmic, linear near zero. The exact full-pool point includes all unique support inputs and has zero estimation error. Categorical Gumbel is a finite-policy baseline, not an autoregressive token implementation.}\label{fig:finite}
\end{figure}

Shared acceptance has overlapping empirical MSE ranges with independent star coupling and uses fewer required labels at the largest sampled budget, as shown in Figure~\ref{fig:finite}. The tree shows a different cost--error tradeoff. Ranking error can stay large when true means are nearly tied even where MSE is small, so we report gain estimation first and selection regret as secondary.
\par
\newpage
\enlargethispage{\dimexpr-3\baselineskip-\parskip\relax}

\subsection{Rare Events and Interval Diagnostics}\label{app:F-rare}
We conduct synthetic coverage checks with known means for three distinct estimands: a single prompt, a fixed 32-prompt bank and a prompt population with a fixed source mixture. Five fixed conditions are used---a symmetric null (C1), a small nonzero gain (C2), a one-sided gain (C3), correlated updates with unchanged individual marginals (C4), and heterogeneous source means with unequal sampling allocation (C5)---each with 2,000 independent simulation repetitions at the budgets $n=8,16,32,64$. The all-copy Clopper--Pearson envelope applies to the specified conservative procedure rather than providing a universal sampling bound.  These are not calibrated coverage claims for the language~models.
\par

\subsubsection{Rare-Event Interval Diagnostics}\label{app:coverage}
\begin{figure}[!t]
\centering\includegraphics[trim=8.73bp 15.17bp 5.5bp 3.65bp,clip,width=0.868\linewidth]{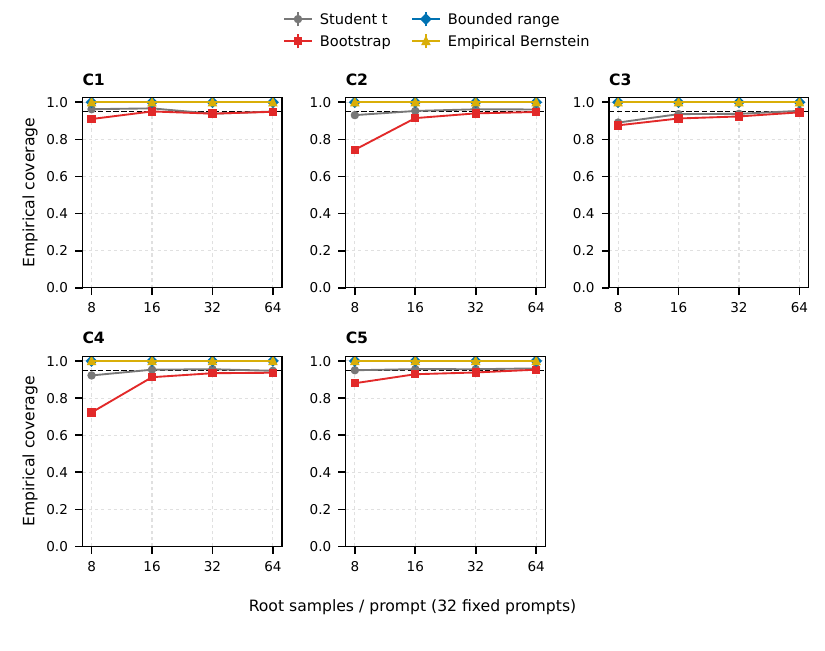}
\caption{Fixed-prompt marginal coverage in all five finite-support diagnostic conditions. The dashed line is 0.95. Error bars show $1.96$ simulation standard errors. The range and empirical Bernstein intervals retain a nonzero range term when all observed differences are zero. This figure compares fixed-budget intervals, not anytime-valid confidence sequences. C1--C5 are defined in the text.}\label{fig:coverage}
\end{figure}

For the fixed-prompt diagnostics, let $D_{i,t}$ be one update's score difference at prompt $x_i$ and draw $t$, and set $U_t=P^{-1}\sum_{i=1}^{P}D_{i,t}$ with $P=32$. Student intervals use the standard error $\sqrt{s_U^2/n}$, where $s_U^2$ is the sample variance of the $n$ draw averages. The bootstrap resamples these $U_t$ values. The bounded-range and empirical Bernstein procedures also use the $n$ draw averages. With independent prompt streams, $s_U^2/n$ and the stratified estimator in Section~\ref{sec:recover} estimate the same variance of the fixed-prompt mean. Their finite-sample values can differ.
Figure~\ref{fig:coverage} compares marginal coverage across the five fixed-prompt diagnostic conditions. The dashed line marks the nominal coverage, while the error bars show simulation uncertainty.
\par

The implementation also records interval width, all-zero sample frequency, missed-tail direction, and simultaneous coverage across the three updates. Three unadjusted marginal 95\% intervals do not promise joint 95\% coverage. The ordinary bootstrap and Student intervals can collapse to $[0,0]$ on an all-zero sample, whereas bounded-range and empirical Bernstein procedures retain uncertainty \citep{maurer2009empirical}. This is why an empirical Monte Carlo standard error of zero is reported as a sample statistic, not as proof of a zero population variance. The full diagnostic data include the single-prompt and population settings and preserve their distinct estimands. Results from confidence sequences are retained separately with their stronger time-uniform validity requirement \citep{howard2021time,waudby2024estimating}.
\par

\paragraph{What an All-Copy Cell Can Certify.}
For one fixed prompt and update, let $d$ be the probability that the two decoded responses differ. Since the score difference is in $[-9,9]$, $|\theta(x)|\le9d$. With no disagreements in $n$ independent draws, the upper endpoint of a two-sided 95\% Clopper--Pearson interval is $d_U=1-0.025^{1/n}$ \citep{clopper1934limits}. Thus this envelope is $|\theta(x)|\le9(1-0.025^{1/n})$, approximately $0.504$ at $n=64$. Reaching $0.05$ with this particular envelope requires 663 all-copy draws. These are properties of a conservative confidence procedure, not a universal sample-complexity lower bound or a requirement for the bank-level estimate. An all-zero sample makes an ordinary bootstrap or Student interval collapse to $[0,0]$: it misses a nonzero target but contains a target equal to zero. The conditional coverage results from these simulations therefore do not imply zero coverage for every underlying policy or establish coverage for real models.
\par

\subsection{Missing Feedback and Resampling Units}\label{app:F-missing}
An unavailable score is never recorded as 0. In the primary analysis, each unavailable required score contributes its admissible range through its signed, aggregated coefficient. Each unknown identity is therefore bounded once, rather than once per occurrence. A covariance point estimate is withheld whenever a required draw score is missing. Three uncertainty units are distinguished throughout: fixed-prompt Monte Carlo error (MCSE, conditional on the prompts and cache), prompt-resampling bootstrap intervals (within sampling strata), and variation across training seeds, which is described by ranges and never pooled into a confidence interval. Historical intersection analyses that drop prompts with incomplete feedback are labelled as such and do not replace the complete primary analysis.
\par

\subsubsection{Unified Evidence and Inference Contract}\label{app:finalcontract}

\paragraph{Scope of This Contract.}
The evaluated model families, prompt sources and sampling strata are specified in Appendix~\ref{app:B-provenance}, and the frozen-snapshot procedure that produces the analysis inputs in Appendix~\ref{app:I-artifacts}. These are open-ended generation tasks. No exact-answer rule is used as the primary judge. The models are Qwen2.5 \citep{qwen2024report} and SmolLM2 \citep{allal2025smol}. Generation uses PyTorch \citep{paszke2019pytorch} and Transformers \citep{wolf2020transformers}, and the statistical pipeline and vector figures use NumPy \citep{harris2020numpy} and Matplotlib \citep{hunter2007matplotlib}.
\par

\paragraph{Unavailable Scores.}
Write a requested linear statistic as $\sum_u b_u r(u)$ and let $M$ be the unavailable scoring identities. With $r(u)\in[1,10]$, the exact identified interval conditional on acquired scores is
\begin{equation}\label{eq:appendix-missing-score-interval}
 \left[\sum_{u\notin M}b_ur(u)+\sum_{u\in M}\min\{b_u,10b_u\},\quad
       \sum_{u\notin M}b_ur(u)+\sum_{u\in M}\max\{b_u,10b_u\}\right].
\end{equation}
\par

Coefficients of a shared missing identity are combined before computing this interval. A platform refusal is a terminal observation about accessibility, not a quality score. The missing input is retained in the support count. Pending calls keep a snapshot interim. Terminal refusals allow completion with explicitly bounded results.
\par

\paragraph{Mean and Variance Support.}
The two supports are defined in Section~\ref{sec:cancel} and Appendix~\ref{app:A-cov}. The pipeline reports $|\Sdraw\setminus\Smean|$ explicitly. If any required draw score is unavailable, it withholds a point variance rather than using an uncharged or imputed value. For a contrast in $[-9,9]$ and $n$ draws, $s^2/n\le81/(n-1)$ supplies a conservative upper bound. An observed zero variance is separately flagged and never treated as a population-variance certificate.
\par

\paragraph{Two Uncertainty Targets.}
For a fixed bank of $P$ prompts and $R$ independent generation repetitions, each single-update gain averages $PRn$ bounded differences. The estimated fixed-prompt Monte Carlo variance is the sum of prompt--repetition variance-of-mean estimates divided by $(PR)^2$. Assuming independent root draws under the frozen sampler and a fixed score cache, a two-sided Hoeffding interval expands the identified mean interval $[L,U]$ to
\begin{equation}\label{eq:appendix-hoeffding-interval}
 [L-\varepsilon,\ U+\varepsilon],\qquad
 \varepsilon=18\sqrt{\frac{\log(2/\alpha)}{2PRn}}.
\end{equation} This remains nondegenerate when every observed pair copies. It is a fixed-budget bound, not an optional-stopping guarantee. This is the standard bounded-difference tail bound. The terms need not be identically distributed, but they must satisfy the independence stated above. Empirical Bernstein bounds \citep{maurer2009empirical} provide a variance-sensitive alternative that retains a nonzero range~term.
\par

For generalization over prompts, the pipeline instead resamples prompts within each frozen stratum, keeping the same prompt resamples across methods and judges. It averages collected generation repetitions within each sampled prompt. These percentile intervals describe prompt heterogeneity conditional on collected generations. They do not incorporate a new distribution of training trajectories. Independently generated samplers are not paired by root or repetition index. The descriptive three-trajectory distance scan and single-trajectory 7B extension are kept separate from fixed-policy generation replication.
\par

For an independently sampled prompt population, a prompt-level mean and its across-prompt variance estimate only need the per-prompt means. Within each independently sampled stratum, total variance~gives
\begin{equation}\label{eq:appendix-prompt-total-variance}
 \operatorname{Var}(\bar\theta)=P^{-1}\{\operatorname{Var}_x[\theta(x)]+\E_x\operatorname{Var}[\widehat\theta(x)\mid x]\}.
\end{equation}
Fixed source allocations require stratum weights and within-stratum variance estimates. The pooled iid formula is not applied to the fixed mixture without adjustment. Such intervals concern the prompt population, whereas resampling independent generation repetitions concerns Monte Carlo variation on the fixed prompt set. Repetition-level intervals require the acquisition set for every repetition and incur no additional scoring cost only when those scores have already been acquired.
\par

\paragraph{Paired Judges and Repeated Looks.}
Judge differences are computed from the same per-draw contrasts before taking a variance, so their observed covariance is retained rather than assuming independent judge estimates. The resampling units and coverage scopes are in Appendices~\ref{app:judges} and~\ref{app:F-judges}. Fixed-budget confidence intervals are not interpreted as confidence sequences \citep{howard2021time,waudby2024estimating}, and failure to detect a difference is not an equivalence~claim.
\par

\paragraph{Coverage Rather than Imputation.}
The primary versioned analysis retains the complete frozen prompt set, and a required score that is still unavailable after bounded retries contributes its admissible range $[1,10]$ to the contrast and uncertainty bounds rather than being dropped or imputed. Historical common-coverage analyses instead state their matched intersection explicitly, and some therefore contain 31 prompts. They are a separate analysis object with its own coverage record and never replace the primary one. Supplementary cost-audit calls are recorded in a separate ledger and do not overwrite the primary score cache.
\par

\subsection{Extended Related Work}\label{app:F-related}
The main text keeps the most direct comparisons. The complete discussion covers the relation to tree-coupled feedback sharing, coupled rejection sampling, label-efficient evaluation, pairwise judges and uncertainty under repeated reporting. In the discussion of TCAB \citep{wen2026tcab}, Section~6.1 and Appendix~E.1 refer to the TCAB paper.
\par

\paragraph{Coupling and Controlled Randomness.}
Section~\ref{sec:related} places CAFA against coupling and common-random-number methods: they control the randomness of a comparison, whereas CAFA conditions on the resulting samples and identifies which score queries can change the requested statistic. The shared-acceptance extension in Appendix~\ref{app:H-shared} applies the classical Doeblin intersection bound \citep{makur2024doeblin} to a component-query objective.
\par

\paragraph{Coupled Evaluation and Feedback Sharing.}
Coupled autoregressive generation \citep{benz2026coupled} couples token generation for language-model evaluation. Shared token randomness is a direct baseline. Tree-coupled feedback sharing in TCAB \citep{wen2026tcab} preserves policy trajectories. Its original query rule purchases one reward per matched component, with cost $T+\sum_{t,e}D_{e,t}$. Section~6.1 of the TCAB paper includes nonadaptive language-model policies. Appendix~E.1 of that paper extends its variance bound to fixed zero-sum contrasts by expressing each contrast as a sum of pairwise differences. Our additional operation removes a score query when its realized coefficient vector is zero. For an all-matched round this changes one shared query to zero. On the comparison-bank tree samples it removes \AutoStageFiveQwenTreeSaving{}\% and \AutoStageFiveSmolTreeSaving{}\% of the already deduplicated inputs. Thus the improvement applies on top of the TCAB sampling construction.
\par

The interfaces explain the distinction. TCAB preserves reward-bearing trajectories, including histories that may determine later adaptive actions. Here the policies are frozen and feedback is a versioned scoring function: the requested contrast specifies which scores to acquire, decoded text specifies which scores to share, and the requested uncertainty statistic determines whether within-draw support must also be purchased. Proposition~\ref{prop:signpure} characterizes when immediate CAFA already reaches the final-batch count. These conclusions for CAFA are distinct from the fixed-contrast variance extension in TCAB.
\par

\paragraph{Coupled Rejection Sampling and Residual Design.}
The coupled rejection sampling framework \citep{corenflos2022coupled} constructs coupled samplers from dominating proposals and characterizes their coupling probability and execution time. Its shared acceptance and independent completion mechanisms supply prior foundations for our residual kernel. Our decision variable is a partition of residual requests, selected from an independent score-free pilot for the generation-plus-acquisition cost of a specified contrast. Proposition~\ref{prop:group-law} records the pair-law preservation needed by this evaluation interface. We evaluate pilot-selected residual partitions by their required scoring support and full recorded cost.
\par

\paragraph{Evaluation with Fewer Labels or Examples.}
The active risk-comparison method \citep{sawade2012comparison} directly compares prediction-model risks under a labeling budget, allocating labels according to model disagreement and the power of the comparison test. This shares our motivation to estimate differences directly. Its interface purchases a supervised outcome for a test instance. Our interface purchases a score for a realized generated response and first removes terms that are irrelevant for every score function. Active testing \citep{kossen2021testing} and active surrogate estimators \citep{kossen2022active} further select informative labels while controlling estimation error. Evaluation-set reduction \citep{polo2024tiny,vivek2024anchor}, uncertainty-driven allocation of repeated judge queries \citep{saha2026budget} and broad task suites \citep{srivastava2022beyond,liang2022helm} are further budget dimensions that compose with exact support pruning. Our paid-pilot experiment measures that composition using identical pilots and surrogates in its controls.
\par

\paragraph{What an Automatic Score Measures.}
An unrestricted pairwise preference judge evaluates a function $q_x(y,y')$, which need not be representable as $r_x(y')-r_x(y)$ for any scalar scoring function. Its expected preference can depend on the chosen coupling, even with fixed marginals. Pairwise rankings can differ under coupled and independent generation \citep{benz2026coupled}. A pairwise API therefore changes the estimand unless an additional representation assumption is supplied. Treating its input as an unordered pair also requires a symmetry and orientation convention, which a position-sensitive judge need not satisfy. The experiments evaluate scalar-score differences.
\par

Model-based judging is studied with pairwise arenas \citep{zheng2023judge,chiang2024arena}, structured criteria \citep{liu2023geval}, trained evaluators \citep{kim2023prometheus,kim2024prometheus2,zhu2023judgelm} and preference-model benchmarks \citep{lambert2024rewardbench}. Two boundaries carry over to this paper. A pairwise target is not the scalar-score target, and it needs a position and orientation convention that a position-sensitive judge need not satisfy. Our pointwise guarantee applies to the selected scoring function and does not establish agreement with human judgment. Known length bias \citep{dubois2024length} and self-preference \citep{wataoka2024self} motivate the paired cross-judge checks. Diverse smaller judges can offer an attractive alternative to a single large judge \citep{verga2024juries}. We retain a fixed strong-judge audit alongside full Flash evaluation.
\par

\paragraph{Uncertainty and Repeated Reporting.}
Appendices~\ref{app:F-rare} and~\ref{app:F-missing} apply the statistical background this paper relies on: resampling of the relevant experimental unit \citep{efron1979bootstrap}, evaluation design for NLP \citep{dror2018hitchhiker} and few-run reinforcement-learning comparisons \citep{agarwal2021precipice}, bounded-variable concentration \citep{maurer2009empirical} and exact binomial limits \citep{clopper1934limits} for the rare mismatches that yield zero empirical variance, and confidence sequences for optional stopping \citep{howard2021time,waudby2024estimating}. Our budget comparisons use fixed sample counts, and reporting an estimate online is a CAFA reporting property that does not by itself provide time-uniform coverage.
\par

\section{Computation and Monetary Accounting}\label{app:cost}

\subsection{Cost Definitions and Measured Scopes}\label{app:G-definitions}
Four quantities are kept apart. (i) The necessary support ($\Qc$, $\Qm$, $\Qd$) is a count of scoring identities, fixed by the bank and report before any score is seen. (ii) New paid calls are what one execution buys. With a warm cache they can be zero although the necessary support is not.
\par

(iii) Recorded workflow cost adds generation $G$ to feedback $F$: GPU time is priced at a common reference tariff of 3 CNY (USD 0.4443) per card-hour rather than a supplier invoice, and feedback at the frozen judge tariff applied to measured token usage at 2 and 8 CNY per million input and output tokens for DeepSeek Flash, as specified in Appendix~\ref{app:G-currency}. (iv) Research expenses, including full-cache audit scoring, failed calls, reference generation, training and restarts, are recorded separately from workflow costs. Same-bank comparisons hold $G$ fixed on both sides and change only $F$. Cross-sampler comparisons change both and are labelled as workflow comparisons. Unknown costs are left empty, never set to zero.
\par

\subsubsection{Separating Acquisition and Execution Cost}
For a specified comparator, write $\Delta L=L_{\rm comparator}-L_{\rm method}>0$ and $\Delta G=G_{\rm method}-G_{\rm comparator}$. With comparable average feedback price $c$, net saving is $c\Delta L-\Delta G$. For unequal response lengths, actual token-based charges replace $c\Delta L$. Cancelling a fixed batch adds only aggregation work and avoids score calls. Introducing an expensive coupling has a different computational baseline. A shared residual pool used to compare two arms experimentally does not by itself identify two separate deployment bills. Job-level timers and isolated accelerator benchmarks must therefore be distinguished from matched end-to-end cost measurements.
\par

\subsection{Currency}\label{app:G-currency}
All costs were recorded in Chinese yuan (CNY). The main text, Table~\ref{tab:controlled} and the redrawn figures display them in US dollars using one fixed rate: the CFETS central parity published by the State Administration of Foreign Exchange for 18 September 2026, 675.21~CNY per 100~USD, that is $6.7521$~CNY per USD, retrieved on 19 September 2026. Every displayed amount is $\text{cost}_{\rm USD}=\text{cost}_{\rm CNY}/6.7521$, computed from the unrounded CNY record and rounded only for display. This changes the display unit, not the prices: generation keeps the reference tariff and feedback the frozen token tariffs of the experiments. Percentages, counts, standard errors and rankings are unchanged, because a ratio of two amounts does not depend on the unit. The frozen unit prices convert as follows.
\par
\begin{center}\small
\begin{adjustbox}{max width=\linewidth}
\begin{tabular*}{\linewidth}{@{\hspace{\tabcolsep}\extracolsep{\fill}}lrr}\toprule
Frozen unit price & CNY & USD\\\midrule
Generation, per GPU-hour (reference tariff) & 3 & 0.4443\\
DeepSeek Flash, per million input / output tokens & 2 / 8 & 0.2962 / 1.1848\\
Qwen Flash, per million input / output tokens & 0.8 / 2.7 & 0.1185 / 0.3999\\
\bottomrule\end{tabular*}
\end{adjustbox}
\end{center}
With the single exception of Table~\ref{tab:controlled}(B), which is in USD, the tables and figures in Appendices~\ref{app:controlled}--\ref{app:extensions} keep their original CNY amounts so that they match the frozen records line by line. Each states its currency in its caption or column headings. The regenerated floats and a file listing every converted value (source record, CNY value, unrounded USD value, display string and where it is shown) are in the supplementary archive, as detailed in Appendix~\ref{app:I-artifacts}.
\par

\subsection{Tariff Sensitivity and Break-Even Analysis}\label{app:G-tariff}
Holding the measured generation costs and response lengths fixed, scaling the feedback tariff changes which $n=16$ GRPO workflow is cheapest: at one quarter of the frozen tariff shared tokens are cheapest in all six seed--repetition cells. At the frozen tariff the star is cheapest in four and the tree in two. At twice the tariff the tree is cheapest in all six. These are observed expenditure comparisons at a fixed draw budget and not a precision ranking, so the cost--precision curves of Appendix~\ref{app:D-cells} retain that second dimension. The separate precision-constrained readout is in Section~\ref{sec:grpo-main}. The controlled break-even prices follow. A constant per-label threshold and a token-priced threshold over the full covariance support are different quantities.
\par

For an observed-workflow comparison with independent AR plus caching, the condition is
\begin{equation}\label{eq:appendix-workflow-break-even}
 G_m-G_{\rm AR}<\lambda\bigl(F_{\rm AR,cache}-F_{m,\mathrm{mean+cov}}\bigr),
\end{equation}
where $F$ is the feedback charge from measured input and output token usage at the frozen tariff, and $\lambda$ scales that tariff. Table~\ref{tab:revision2-break} supplies this threshold together with the constant-price-per-label threshold $\max(0,\Delta G/\Delta L)$. Both are calculated from unrounded records. The complete result package contains both judges and tariff factors $0,0.25,0.5,1,2,5,10$.
\par

\begin{table}[!htbp]
\centering\small
\caption{Break-even feedback prices relative to the recorded independent-AR-plus-cache workflow. $\Delta G$ is additional generation CNY and $\Delta L$ is the number of mean labels avoided. The tariff factor instead prices the complete variance support using the actual token usage of each response under DeepSeek Flash. The different generation-overhead conventions from Table~\ref{tab:revision2-unified} remain explicit.}\label{tab:revision2-break}
\begin{adjustbox}{max width=\linewidth}
\begin{tabular*}{\linewidth}{@{\hspace{\tabcolsep}\extracolsep{\fill}}llrrrr}\toprule
Model & Sampler & $\Delta G$ & $\Delta L$ & CNY/1k labels & Tariff factor\\\midrule
Qwen 1.5B & Shared tokens & 0.051 & 2963.5 & 0.017 & 0.016\\
Qwen 1.5B & TV tree & 1.072 & 7216.0 & 0.149 & 0.119\\
Qwen 1.5B & Ind. star & 1.727 & 6942.5 & 0.249 & 0.200\\
Qwen 1.5B & Shared star & 1.727 & 7132.5 & 0.242 & 0.194\\
Smol 1.7B & Shared tokens & 0.024 & 6342.0 & 0.004 & 0.003\\
Smol 1.7B & TV tree & 1.586 & 8000.5 & 0.198 & 0.154\\
Smol 1.7B & Ind. star & 2.320 & 7989.0 & 0.290 & 0.225\\
Smol 1.7B & Shared star & 2.320 & 8035.5 & 0.289 & 0.224\\
\bottomrule\end{tabular*}
\end{adjustbox}
\end{table}

At the fixed DeepSeek tariff, the tree workflow for Qwen costs 9.01 CNY, compared with 16.97 CNY for independent AR plus caching. The corresponding RMS MCSE values are 0.0114 and 0.0308. The independent-star workflow costs 10.06 CNY under the full-joint-task charge. Scoring price can change the preferred recorded workflow: for SmolLM2 under Qwen Flash, shared tokens cost 6.79 CNY while the tree costs 7.43 CNY, with RMS MCSE 0.0119 and 0.0054 respectively. These joint measurements expose the price--precision tradeoff directly.
\par

The same-generation CAFA comparison and the reconstructed cross-sampler comparison answer complementary deployment questions. Their generation terms cannot be cancelled interchangeably. For smaller $n$ in the controlled experiments, we report measured support--MCSE curves but leave generation costs unspecified because complete timings were not collected. Full-budget task intervals do not determine execution times at lower budgets.
\par

\subsection{Bookkeeping and Residual Computation}\label{app:G-bookkeeping}
On a local CPU, score-free support aggregation for the six predecoded star $n=16$ GRPO banks took 0.081--0.150 seconds, measured as medians of five runs after a warm-up and excluding loading, hashing, scoring and generation. These timings cover only support aggregation and do not measure end-to-end latency. The signed-overhead sensitivity asks how much unrecorded extra cost a method could carry before losing its recorded advantage. Zero and negative margins are kept and give no headroom. Partial batch timers cover different scopes (tree timers include only root generation) and do not rank methods. In the idealized untruncated construction without batch overshoot, each child consumes one residual proposal per root draw in expectation when $d=\operatorname{TV}(p_0,p_j)>0$. Conditional on needing a residual, the expected count is $1/d$, so small positive residual mass produces rare but long requests. Pool-exhaustion probabilities follow the binomial bound in Appendix~\ref{app:residual}.
\par

\paragraph{Cost Identification and Unrecorded Overhead.}
The same-bank monetary increment holds recorded generation fixed. A cross-sampler comparison additionally depends on execution history. Let $M=T_{\rm AR,cache}-T_{m,report}$ be the observed cost margin. If unrecorded method overhead minus unrecorded baseline overhead is $\delta$, the observed ranking survives precisely when $\delta<M$. This signed sensitivity is not an estimate of missing execution time. The seven cells with failure/recovery histories now use uninterrupted full-process replays, as detailed in Appendix~\ref{app:completion-timing}. Their historical charges remain separate. The difference between historical and replay charges is not an identified failure-only cost. Hardware and runtime differences remain possible sources of overhead.
\par

For the six $n=16$ star cells, the margin over independent AR plus caching is 1.471--2.250 CNY. An additional net charge equal to the respective margin removes the recorded cost advantage. The complete signed margins for all workflows are retained in the audit.
\par

The signed margin of every one of the 72 frozen workflow cells against independent AR plus caching is retained in the supplementary canonical tables, in CNY: a net additional method-minus-baseline charge of that amount yields a tie. Seven margins are negative and eighteen are zero, so those cells offer no positive headroom at all. This is an accounting sensitivity on costs recorded elsewhere.
\par

\paragraph{Local CPU Aggregation Timing.}
Python 3.11.14, NumPy 2.4.6, Windows, a single local process without pinned affinity. Six star banks at 16 draws each use predecoded scoring identities, with one warm-up and five measurements per operation. These CPU seconds are not added to historical GPU bills without an identified CPU tariff.
\par
\begin{table}[!htbp]\centering\small\setlength{\tabcolsep}{3pt}
\caption{Local bookkeeping seconds: five-repeat medians. Cache-only constructs the distinct-identity set. Support aggregation additionally constructs exact integer coefficient maps and both supports. Excludes reading, hashing, scores and generation.}\label{tab:review3-cpu}
\begin{adjustbox}{max width=\linewidth}
\begin{tabular*}{\linewidth}{@{\hspace{\tabcolsep}\extracolsep{\fill}}lrrr}\toprule
Seed/repeat & Cache & Support & Difference\\
\midrule
1/1 & 0.000442 & 0.119690 & 0.119248\\
1/2 & 0.000658 & 0.135111 & 0.134453\\
2/1 & 0.001007 & 0.143030 & 0.142022\\
2/2 & 0.000641 & 0.149643 & 0.149002\\
3/1 & 0.000396 & 0.080597 & 0.080200\\
3/2 & 0.000529 & 0.113345 & 0.112816\\
\bottomrule
\end{tabular*}
\end{adjustbox}
\end{table}
\subsubsection{Residual Moments and Pool Limits}\label{app:residual}
Let $d=\mathrm{TV}(p_0,p_j)>0$, and let $N$ be the number of residual proposals consumed for one child in one root draw, without truncation or batch overshoot. A residual is needed with probability $d$. Conditional on that event, $N$ is geometric with success probability $d$, so
\begin{equation}\label{eq:appendix-residual-proposal-moments}
 \E N=1,\qquad \E N^2=\frac{2-d}{d},\qquad
 \operatorname{Var}(N)=\frac{2(1-d)}{d}.
\end{equation}
\par

At $d=0$, $N=0$ almost surely. Small positive $d$ therefore produces rare but long residual requests despite a constant unconditional mean. Both star arms have the same expected proposal count. This does not identify their incremental wall time when batches, devices, and a shared experimental pool~differ.
\par

Conditional on requiring $m$ residuals, with iid proposals independent of this requirement, a cap of $K$ proposals fails with probability
\begin{equation}\label{eq:poolrisk}
 F_{K,m}(d)=\Pr\{\operatorname{Binomial}(K,d)<m\}
 =\sum_{k=0}^{m-1}\binom{K}{k}d^k(1-d)^{K-k}.
\end{equation}
The geometric tail $(1-d)^K$ applies only to $m=1$. As conditional arithmetic, $d=0.003187$ and $K=1024$ give failure probabilities approximately 0.038 for one residual and 0.769 for five. These are illustrative probabilities at an assumed fixed $d$, not measured failure rates of a heterogeneous job~queue.
\par

Heterogeneity has no universal direction for a general pool. For $1\leq m\leq K$ and $0<d<1$,
$F'_{K,m}(d)=-K\binom{K-1}{m-1}d^{m-1}(1-d)^{K-m}$.
For $m>1$, the sign of $F''_{K,m}(d)$ is the sign of $(K-1)d-(m-1)$. The function is concave below this threshold and convex above it. For $m=1$ it is convex, and Jensen's inequality applies to a mean over the same population weights. A proposal-weighted empirical acceptance rate is not automatically that mean. Pool sizing requires the per-condition acceptance behavior and residual demand, rather than a queue-wide rate alone.
\par

\subsubsection{Computational Cost Audit}\label{app:costaudit}
The original harness recorded batch timers whose scope is narrower than the accounting used here. Those timers measure generation only, not density evaluation, model loading or bookkeeping. Independent autoregressive and shared-token runs time their generation batches, whereas tree runs time root generation and omit the residual-proposal durations, so a smaller recorded sum for the tree is not evidence of cheaper execution. A residual pool shared between two experimental arms likewise does not yield a separate measured bill for each arm, and reusing a frozen probability pilot is a different deployment scenario from paying for a new one. The 24 method--job sums themselves are retained in the supplementary audit records rather than reprinted here.
\par

The original harness also stored flat-rate API estimates. Appendix~\ref{app:revision-cost} replaces those estimates with usage-priced accounting and reconstructs the complete controller task intervals, which include the work the component timers did not cover. The main-text before--after comparison charges the same complete generation interval to both scoring policies.
\par

\subsection{Collection History, Accepted Timings and Resource Archive}\label{app:G-history}
This subsection is the one place that records which recorded timings were replaced and why. Twelve earlier isolated replays replaced contended budget-4 collection times, and seven uninterrupted full-process replays replaced cells with failure or recovery histories.
\par

All 19 reproduce the original token and decoded-text identities and add no statistical samples. The six new Tree budget-4 banks, by contrast, are real additional generation, as detailed in Appendix~\ref{app:completion-timing}.
\par

The replays were run as fresh single-process child jobs, one per GPU, on a two-RTX-5090 server (32~GB per GPU, Python 3.12.3, PyTorch 2.8.0+cu128, Transformers 4.56.2), timing model loading, the Tree pilot where applicable, generation, density calculations and saving, but not independent reconstruction, archive transfer or local analysis. Every archive was downloaded and SHA-256 checked. The remaining accepted historical timings are retained, so this is a targeted repair and not an all-cells same-machine benchmark.
\par

Historical charges, including the campaign GPU allocation, stay in the ledgers below and are never added to accepted workflow timings. The difference between historical and replay charges includes CPU, I/O and runtime differences, and several replay costs are slightly larger. The inventory of banks and of the versioned score cache is stated once, in Appendix~\ref{app:grpo-protocol}. That count is an experimental audit union, not a deployment query count or a post-saving bill.
\par

\subsubsection{Historical Cost Reconstruction}\label{app:revision-cost}
Controller events are matched by job and GPU UUID, with an explicit check that leases do not overlap. All recovered starts have exits, and resumed segments are costed under both their original and their continuation records. Reserved task time within the recorded five-card campaign came to 223.04 card-hours, or 669.12 CNY at 3 CNY per reserved card-hour, and continuous rental outside assigned tasks is a separate account-level expense. This total covers the entire research campaign, including validation, resumed stages, training, distance-scan generation, both star families and the baselines. It is not assigned to any single deployment comparison. Its per-category breakdown is retained in the supplementary audit records.
\par

The API audit reconstructs usage-priced costs from returned token counts, cache-hit counts and model identity. DeepSeek Flash is priced using its documented peak/nonpeak schedule in Beijing time. Qwen uses its model-specific input, output and cache tariffs. We checked the \href{https://api-docs.deepseek.com/zh-cn/quick_start/pricing/}{DeepSeek tariff page} and \href{https://help.aliyun.com/zh/model-studio/qwen3-8-flash}{Qwen Flash tariff page} on September 12, 2026. These reconstructed amounts represent research expenses rather than workflow costs and differ from both the earlier local flat-rate reservations and the provider account invoice. The cost breakdown by judge is retained in the supplementary audit records, including 258 calls with no returned usage data. These calls remain unresolved billing entries and are not assigned zero cost. Historical imported scores are outside that live-call accounting. Price sources, timestamps, raw responses and calculation code are retained in the result package.
\par

\paragraph{Strong-Baseline Phase Windows.}
The strict-holdout baseline process executes independent autoregressive, shared-token and TV-tree sampling sequentially. Original server save timestamps delimit their inclusive phase windows. Common loading/tree setup and finalization are listed separately in Table~\ref{tab:baseline-phase-costs}. Each complete repetition comprises four disjoint eight-prompt shards. These are observed phase costs within that workflow, including density computations in each phase. They are not independently benchmarked standalone launches.
\par
\begin{table}[!htbp]
\centering\caption{Inclusive strict-holdout baseline phase costs in CNY, reconstructed from original server timestamps. Common setup and finalization are separate.}\label{tab:baseline-phase-costs}
\begin{adjustbox}{max width=\linewidth}
\begin{tabular*}{\linewidth}{@{\hspace{\tabcolsep}\extracolsep{\fill}}llrrrr}\toprule
Model & Repetition & Independent AR & Shared tokens & TV tree & Common\\\midrule
Qwen 1.5B & 1 & 6.774 & 6.813 & 8.031 & 0.076\\
Qwen 1.5B & 2 & 6.785 & 6.849 & 7.673 & 0.074\\
Smol 1.7B & 1 & 5.681 & 5.662 & 7.857 & 0.079\\
Smol 1.7B & 2 & 5.652 & 5.719 & 6.648 & 0.077\\
\bottomrule\end{tabular*}
\end{adjustbox}
\end{table}

\paragraph{Saved-Prefix and Numerical Audit.}
Continuation preserves the saved candidate prefix, the frozen policies, the output cap and the random namespace, so the completed pools use a ceiling of 8,192 proposals exactly where the original 1,024-proposal prefix needed continuing. All 896 copied checkpoint files match their recorded SHA-256 hashes, and the detailed logs stay in the archived ledgers. No budget-exhausted cell is removed from the completed analysis, and the original numerical acceptance records and source hashes are retained with the reproduced generation audit. Table~\ref{tab:completion-replays} separates historical charges from accepted uninterrupted charges for the targeted timing replays.
\par
\begin{table}[!htbp]\centering\small\setlength{\tabcolsep}{3pt}
\caption{All seven targeted timing replays (training seed 1). Historical charges and accepted uninterrupted charges remain separate. Differences are not interpreted as isolated failure losses.}\label{tab:completion-replays}
\begin{adjustbox}{max width=\linewidth}
\begin{tabular*}{\linewidth}{@{\hspace{\tabcolsep}\extracolsep{\fill}}lrrrrr}\toprule
Sampler & $n$ & Repeat & Historical CNY & Replay hours & Replay CNY\\
\midrule
Independent star & 4 & 1 & 3.603 & 1.259 & 3.777\\
Independent star & 4 & 2 & 3.395 & 1.172 & 3.515\\
Independent star & 8 & 1 & 3.925 & 1.286 & 3.859\\
TV-tree & 8 & 1 & 4.877 & 1.564 & 4.693\\
TV-tree & 8 & 2 & 4.916 & 1.632 & 4.895\\
TV-tree & 16 & 1 & 6.574 & 2.187 & 6.560\\
TV-tree & 16 & 2 & 6.776 & 2.237 & 6.712\\
\bottomrule
\end{tabular*}
\end{adjustbox}
\end{table}

\section{Extensions of the CAFA Framework}\label{app:extensions}

\subsection{Extension Overview}\label{app:H-overview}
Three extensions change what is sampled or how many scores are bought, rather than only which scores are skipped. None is part of the exact core of the main text. Shared acceptance in Appendix~\ref{app:H-shared} uses one copy uniform for all updates to create more all-copy draws. It preserves every base--update law, but its unique-input increment on top of CAFA is small and changes sign on some banks. Residual grouping in Appendix~\ref{app:H-residual} lets updates share residual proposals. It preserves every base--update law and reduces the required support, but its recorded cost including the pilot rises, so it buys fewer scores for more money. Budgeted acquisition in Appendix~\ref{app:H-budget} buys fewer scores than the exact support with a generalized-difference estimator. Restricting the budget to the non-zero support lowers the conditional MSE substantially, while active allocation shows no consistent advantage over uniform~allocation.
\par

\subsection{Shared Acceptance}\label{app:H-shared}
Shared acceptance uses a common uniform random variable for copy decisions across updates to increase all-copy draws without changing any base--update law, as established in Proposition~\ref{prop:law}. It minimizes a restricted component-query objective, as established in Proposition~\ref{prop:cost}, although it changes the joint law across updates. On the initial fixed-prompt bank, the four-run unique-input total falls from 2728 to 2303. The conditional component ratio is 0.8925, with a 95\% prompt-cluster interval of $[0.8846,0.8994]$. This ratio has a different denominator from the unique-input reduction. On the comparison bank, shared acceptance removes a further 7.4\% of the inputs remaining after cancellation for Qwen and 19.1\% for SmolLM2. The increment changes sign on the 7B banks and in some strict-holdout repetitions, indicating a small and inconsistent additional benefit over CAFA.
\par

\subsubsection{Coupling and Budget Extensions}\label{app:coupling-extensions}
The following constructions use established sampling principles to explore how dependence changes CAFA support.
\par

\paragraph{Shared Acceptance for a Common Base.}
The Independent Star of Section~\ref{sec:coupling} and Appendix~\ref{app:B-samplers} draws $X\sim p_0$ and copies it into policy $j$ with probability $a_j(X)=\min\{1,p_j(X)/p_0(X)\}$, the classical maximal coupling \citep{levin2017markov}. Otherwise it samples the normalized residual law
\begin{equation}\label{eq:residual}
 \rho_j(y)=\frac{(p_j(y)-p_0(y))_+}{\operatorname{TV}(p_j,p_0)},
\end{equation}
which is never needed when the total variation is zero, and which is implemented by proposing $Z\sim p_j$ and accepting with probability $(1-p_0(Z)/p_j(Z))_+$ at a cost that must be counted.
\par

The single change made here is the copy uniform: the Independent Star tests $U_j<a_j(X)$ with separate $U_j\sim\mathrm{Uniform}(0,1)$, whereas we use one common $U$, independent of $X$ and of the residual streams. Conditional on $X$, the probability of all copies changes from $\prod_j a_j(X)$ to $\min_j a_j(X)$. Each child still copies with its original probability. Residual streams remain independent across children within an arm. Reproducible sharing between experimental arms is recorded separately.
\par

For the primary base-relative target, an all-copy draw requires no score. Other draws use coefficient aggregation in Eq.~\eqref{eq:cancel}. Sharing acceptance therefore targets an event with direct feedback consequences, without using quality scores to select favorable responses.
\par

\begin{proposition}[Preserved base--update laws]\label{prop:law}
Under the independent and shared acceptance constructions, every pair $(X,Y_j)$ has the same joint distribution. Consequently, for independent repetitions, each individual base--update contrast estimator has the same finite-sample distribution.
\end{proposition}
\par
For any measurable response sets $A$ and $B$, $\Pr(X\in A,\,Y_j\in B)$ is the integral over $x'\in A$ of $p_0(x')[a_j(x')\1\{x'\in B\}+(1-a_j(x'))\rho_j(B)]$. That expression involves only $X$, the conditional copy probability and the residual kernel for the fixed $j$, never the dependence of the copy uniform across children, so it is identical under independent and shared acceptance. In particular $Y_j$ has marginal law $p_j$: the copied mass is $\min(p_0(y),p_j(y))$ and the residual mass is $(p_j(y)-p_0(y))_+$. The argument requires the residual streams to remain independent of the root and of the copy uniforms. Independent generating namespaces across repetitions then preserve the complete sample law of each fixed-$j$ average. The equality does not extend to the full vector $(Y_0,\ldots,Y_J)$: cross-contrast covariances and checkpoint-selection error can~change. Reusing random streams across experimental arms changes the dependence between arms but preserves the marginal sampling law of each arm.
\par

\begin{proposition}[Restricted component-query objective]\label{prop:cost}
Fix $X=x'$ and the copy probabilities $a_1,\ldots,a_J$. Let $B_j$ indicate copying and define $N_{\mathrm{comp}}=(1+\sum_j(1-B_j))\1\{\text{not all }B_j=1\}$. Among joint copy indicators with these Bernoulli marginals,
\begin{equation}
 \E[N_{\mathrm{comp}}\mid X]\ge 1+\sum_j(1-a_j)-\min_j a_j,
\end{equation}
with equality for shared acceptance. Its reduction relative to independent acceptance is $\min_j a_j-\prod_j a_j$.
\end{proposition}
\par
An all-copy round has zero mismatches, so $N_{\mathrm{comp}}=1+\sum_j(1-B_j)-\1\{B_1=\cdots=B_J=1\}$. Taking conditional expectations gives $\E[N_{\mathrm{comp}}\mid X]=1+\sum_j(1-a_j)-\Pr(B_1=\cdots=B_J=1\mid X)$, and an intersection probability cannot exceed its smallest marginal, a bound that a common uniform attains. The multi-way overlap principle is classical \citep{makur2024doeblin}. The specific objective here removes all-copy rounds because their contrast is identically zero.
\par

Two boundaries follow from the same identity. For a single update $\min_j a_j=\prod_j a_j$, so shared acceptance has no incremental component advantage at all, and when every $a_j=1$ both arms require zero feedback. Extra text collisions and batch-level coefficient cancellation can lower realized cost further without changing this component identity.
\par

\paragraph{Residual Cost and Break-Even Feedback Price.}
The residual moments, the pool-exhaustion analysis and the break-even price $\Delta G/\Delta L$ are in Appendices~\ref{app:residual} and~\ref{app:G-tariff}. Small overlap distance does not imply cheap execution. One consequence is specific to this extension: costs common to both star constructions cancel in their incremental comparison, but they do not cancel when the full star sampler is compared against independent generation, so a within-star increment and a star-versus-AR saving must never be offset against each other.
\par

\paragraph{A Bridge from Component Counts to Text-Level Acquisition.}
Let $K=\sum_t N_{{\rm comp},t}$ for the base-relative star construction. After removing all-copy draws, its component occurrences cover $S_{\rm draw}$. Write $D_{\rm reuse}=K-|S_{\rm draw}|\ge0$ for reductions from text grouping and reuse, and $D_{\rm batch}=|S_{\rm draw}|-|S_{\rm mean}|\ge0$. Then, pathwise,
\begin{equation}\label{eq:component-text-bridge}
 |S_{\rm mean}|=K-D_{\rm reuse}-D_{\rm batch}.
\end{equation}
Shared acceptance minimizes the conditional expectation of the component term in Proposition~\ref{prop:cost}. The two correction terms also depend on the sampler. Consequently its realized unique-input increment can change sign, consistently with the 7B results. Cancellation itself always weakly reduces acquisition on a fixed generated bank.
\par
\subsubsection{Additional Coupling and Computation Results}

\paragraph{Simultaneous Agreement and Decoded Text.}
The classical multi-way agreement bound is
\begin{equation}\label{eq:appendix-multiway-agreement}
 \Pr(Y_0=\cdots=Y_J)\leq\sum_y\min_{0\leq j\leq J}p_j(y).
\end{equation}
Shared acceptance attains it: integrating $p_0(y)\min_{j\geq1}a_j(y)$ gives the right side. On a residual branch, the base and that child cannot coincide on the same token sequence. This is a specialization of simultaneous maximal coupling \citep{makur2024doeblin}, not a guarantee about the number of scored texts after batch aggregation.
\par

For example, let four token sequences decode to $A,A,B,B$, and take $p_0=(1/2,0,1/2,0)$ and $p_1=(0,1/2,0,1/2)$. All couplings have the maximum token-sequence agreement of zero. Pairing the first sequence with the second and the third with the fourth gives text agreement one. Reversing these pairings gives zero. Independent residual sampling gives $1/2$. Thus token maximality alone does not determine text-level feedback cost.
\par

\subsubsection{Does Changing the Acceptance Dependence Add More?}\label{sec:increment}
Shared acceptance is an increment over an already coupled baseline, with cancellation available in both arms. The expected component-cost ratio, averaged over repetitions within prompt and then over prompts, is \QwenExpectedRatio{} with a 95\% prompt-cluster bootstrap interval of [\QwenExpectedLower{},\QwenExpectedUpper{}]. The four-run required-input totals are \QwenIndependentInputs{} against \QwenSharedInputs{}, a \QwenQueryReduction\% reduction, with per-repetition counts in Table~\ref{tab:replication}. Proposition~\ref{prop:cost} fixes the sign of the analytical component reduction, not its magnitude, and the inequality implies no particular threshold. The interval pairs the same prompt resamples in numerator and denominator and is conditional on the fixed updates and collected repetitions. It is a component-cost statement, distinct from realized label savings and their uncertainty.
\par

The increment is also smaller and less stable than cancellation itself. On SmolLM2 strict holdout at $n=16$, shared acceptance costs more in two repetitions, ties in one, and costs less in one: the realized increment changes sign within a single bank family. Its comparison-bank means, and the tree means they sit against, are the last columns of Tables~\ref{tab:fullcancel} and~\ref{tab:samplers}. Equation~\eqref{eq:component-text-bridge} separates the component guarantee from these realized text-level increments. Running multiple arms for verification incurs their combined acquisition costs, subject to caching.
\par
\begin{table}[!htbp]
\centering\caption{Required scoring inputs per judge on the fixed Qwen prompt set.
Counts use per-run text deduplication after removing identical-pair contributions. Their sum does not deduplicate across repetitions and does not represent the total API expenditure of the experiment.}\label{tab:replication}
\begin{adjustbox}{max width=\linewidth}
\begin{tabular*}{\linewidth}{@{\hspace{\tabcolsep}\extracolsep{\fill}}lrr}\toprule
Generation repetition & Independent acceptance & Shared acceptance\\\midrule
1 & 669 & 566 \\
2 & 650 & 547 \\
3 & 713 & 627 \\
4 & 696 & 563 \\
\midrule Total & 2728 & 2303 \\
\bottomrule
\end{tabular*}
\end{adjustbox}
\end{table}

\subsection{Residual Grouping}\label{app:H-residual}
Residual grouping shares proposals across updates to increase agreement among their non-copied responses. A mixture kernel with a fixed active block and an independent fallback preserves each base--update law, as established in Proposition~\ref{prop:group-law}. A score-free pilot selects the partition. We evaluate 18 banks from three seeds, three arms and two repetitions, using 32 selection prompts and $n=8$, as detailed in Appendix~\ref{sec:residual-groups}. All three arms use CAFA and differ in their residual dependence. The control uses shared copy acceptance with independent residuals, unlike the Independent Star used in the main study. Pilot-selected grouping consistently requires fewer scores than independent residuals, but costs more after charging its pilot and historical attempts. Pooling increases costs further.
\par

\subsubsection{Using Acquisition Cost to Choose Residual Dependence}\label{sec:group-method}
The coefficient interface can also guide the sampler. For a fixed comparison target, two couplings with the same base--update laws can expose different scoring supports because their updates agree with one another at different rates. We examine this degree of freedom by choosing which residual requests share proposals. This extends the CAFA framework to a concrete design decision without making its exactness depend on that decision.
\par

Retain the shared root $X\sim p_0$ and copy uniform from the star construction. Define $f_j(y)=(p_j(y)-p_0(y))_+$, $\delta_j=\int f_j(y)\,dy$, and $\rho_j=f_j/\delta_j$ when $\delta_j>0$. For a partition $\mathcal P$ of the updates, intersect each block with the updates requesting a residual in this draw. A singleton uses independent residual sampling. For an active block $B$ of size at least two, draw an independent sequence of proposals and shared uniforms according to
\begin{equation}\label{eq:group-kernel}
 Z\sim m_B=\frac{1}{|B|}\sum_{k\in B}p_k,
 \qquad U\sim\operatorname{Unif}(0,1),
 \qquad j\text{ accepts if }U<\frac{f_j(Z)}{\sum_{k\in B}p_k(Z)}.
\end{equation}
\par

Each member keeps its first accepted proposal. The mixture retains the original active block throughout this request, including after some members finish. After $K=32$ proposals, unfinished members use independent draws from $\rho_j$. The kernel requires policy densities but not the unknown residual normalizing constants. Shared proposal acceptance and independent completion are established rejection-sampling tools \citep{corenflos2022coupled}. The design variable here is the partition selected for contrast acquisition cost.
\par

\begin{proposition}[Partition choice preserves the pair laws]\label{prop:group-law}
Fix a partition chosen from pilot randomness independent of the formal draws. Under Eq.~\eqref{eq:group-kernel} and the fixed-$K$ fallback, each residual output has law $\rho_j$, conditional on any active block containing $j$. Consequently every $(X,Y_j)$ has the same law as shared-copy star sampling with independent residuals. For independent draws, every individual base-relative gain retains its distribution and variance. Cross-gain covariance and multi-checkpoint selection error can~change.
\end{proposition}
\par

For one proposal, the joint density of $Z=y$ and acceptance by $j$ is $f_j(y)/|B|$, hence its conditional accepted law is $\rho_j$. A fixed number of failures followed by independent $\rho_j$ completion has the same output law. Appendix~\ref{app:group-proof} proves the statement, including the conditioning on the active set and the implementation cap.
\par

\paragraph{A Score-Free Partition Decision.}
With three updates there are five partitions, including independent residuals and one pooled block. On an independent pilot we choose
\begin{equation}\label{eq:partition-choice}
 \widehat{\mathcal P}=\arg\min_{\mathcal P}
 \left\{\widehat G(\mathcal P)+\lambda\widehat Q_{C,\mathrm{mean}}(\mathcal P)\right\},
\end{equation}
where $\widehat G$ estimates generation and density expenditure, $\widehat Q_{C,\mathrm{mean}}$ estimates the post-cancellation mean support, and $\lambda$ is a frozen mean scoring-input price estimated from historical token usage. The implemented surrogate uses a finite, self-normalized approximation to the policy distributions. It is a decision rule, not a population cost bound. Formal cost reporting uses actual token usage and the support for both means and variances. The pilot is charged separately whenever the selected workflow is deployed. Section~\ref{sec:residual-groups} tests the resulting partition against both fixed alternatives and relates the outcome to the broader sampler comparison.
\par

\subsubsection{Residual Grouping: What the Target-Driven Design Changes}\label{sec:residual-groups}
We evaluate the partition rule in Section~\ref{sec:group-method} on 32 prespecified selection-set prompts, with eight draws, three GRPO seeds and two generation repetitions per seed. Independent residuals, one pooled group and the pilot-selected partition give 18 complete banks. Roots and copy uniforms are paired across the three arms. This experiment isolates residual dependence. The independent-residual control uses shared copy acceptance, whereas the independent star in the main study uses independent copy acceptance.  The independent pilot uses 16 prespecified selection prompts per seed, and all three pilots select $\{\{1\},\{2,3\}\}$. This experiment is separate from the 128-prompt main study and its independent selection/test application.
\par

\paragraph{A Consistent but Small Residual-Grouping Increment under CAFA.}
All three arms here apply the same CAFA rule. The comparison varies only the residual dependence upstream of it. Pilot-selected grouping reduces joint mean/variance support by 2.1--3.9\% relative to independent residuals across all six seed/repetition cells. Pooling reduces it by 2.8--5.1\%. The single base-relative supports are unchanged in every corresponding cell: the extra saving comes from sharing across updates. For the step-50 to step-250 adjacent contrast, selected grouping reduces scoring support by 5.9--11.9\%. This secondary target was recomputed from the fixed banks. The RMS MCSE ratio for the joint gains is 0.937--1.034 for selected grouping, with no uniform precision improvement. Table~\ref{tab:group-summary} summarizes the cells. Appendix~\ref{app:group-evidence} retains target-specific results and uncertainty.
\par

\begin{table}[!htbp]\centering\small
\caption{Residual-group validation, descriptive means of six seed/repetition cells. Each bank has 32 prompts and eight draws. $\Qd$ is the number of inputs required for joint mean-and-covariance reporting, $G$ includes all historical generation attempts, $P$ is the pilot charged per deployment, and $T=G+P+\Fd$ includes feedback at the frozen tariff. All costs are CNY. Recovery overhead is not separately identifiable.}\label{tab:group-summary}
\begin{adjustbox}{max width=\linewidth}
\begin{tabular*}{\linewidth}{@{\hspace{\tabcolsep}\extracolsep{\fill}}lrrrrr}\toprule
Residual grouping & $\Qd$ & $G$ & $P$ & $T$ & RMS MCSE\\\midrule
Independent & 714.83 & 5.532 & 0.000 & 6.493 & 0.06535\\
Pooled & 686.83 & 6.773 & 0.000 & 7.701 & 0.06522\\
Pilot selected & 694.50 & 5.076 & 0.707 & 6.724 & 0.06504\\
\bottomrule\end{tabular*}
\end{adjustbox}\end{table}

\paragraph{Efficient Residual Proposals Do Not by Themselves Pay for the Workflow.}
Three quantities move in different directions here and are measured over different scopes. Selected grouping generates 55.7\% fewer proposal tokens on average than independent residuals, and its recorded generation-batch time is 13.9\% lower, but those timers exclude parts of loading, probability replay and I/O, so neither figure is an end-to-end speed-up and neither can be combined with the total cost below. Pooling instead increases recorded generation time and sends 3.34\% of shared requests to independent fallback on average, versus no shared-request fallbacks for the selected partition. Thus both which members share and how their proposals execute affect cost.
\par

\begin{figure}[!htbp]\centering
\includegraphics[trim=15.78bp 8.76bp 5.5bp 5.22bp,clip,width=0.946\linewidth]{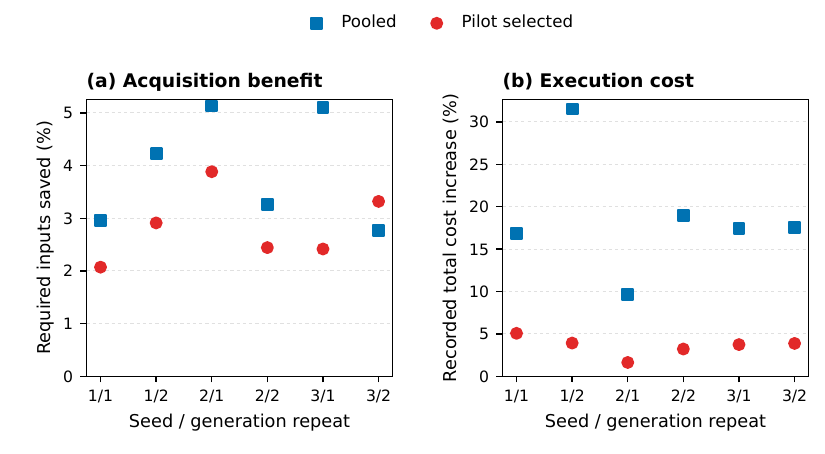}
\caption{Residual grouping separates acquisition and execution outcomes. Each point is one training seed/generation repetition, with independent residuals as the matched control. Left: reduction in joint scoring support. Right: increase in recorded generation-plus-feedback expenditure, charging the pilot for selected grouping and retaining historical recovery overhead.}\label{fig:group-decision}
\end{figure}

Charging the complete recorded attempts and one pilot per deployment, selected grouping costs 1.6--5.1\% more than independent residuals. Pooling costs 9.7--31.5\% more, as shown in Figure~\ref{fig:group-decision}. Interrupted attempts contain both retained useful work and recovery overhead. We preserve these charges and do not interpret their difference as a measured uninterrupted-deployment difference. A conditional pilot-amortization calculation in Appendix~\ref{app:group-proof} suggests modest savings if the same partition serves multiple comparable banks. It is an extrapolation under a stated stability condition.
\par

\begin{figure}[!htbp]\centering\normalsize
\includegraphics[trim=0.86bp 4.95bp 5.5bp 0.69bp,clip,width=0.935\linewidth]{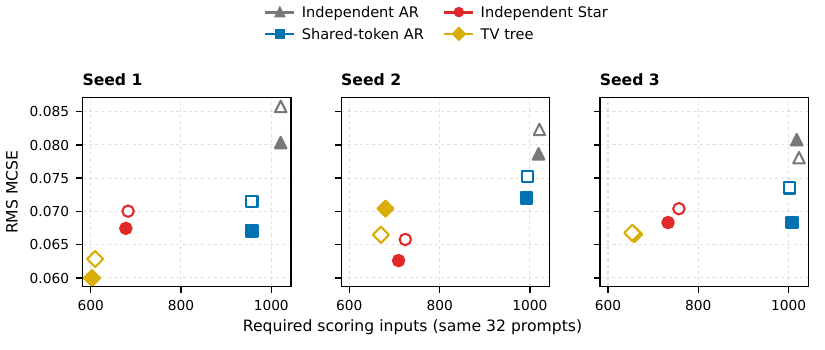}
\caption{Existing samplers on the same 32-prompt subset. Each point is an existing $n=8$ seed/repetition cell. Filled and open markers denote the two repeats. Required support includes mean and covariance. These baselines have independent random namespaces from residual grouping. Standalone subset generation cost is unavailable and is not inferred from the support axis.}\label{tab:group-strong}
\end{figure}

\paragraph{Relation to the Strong Baselines.}
Recomputing the four main samplers on the same 32 prompts and $n=8$ gives a further 24 comparison cells. TV-tree requires fewer scoring inputs than selected grouping in all six seed/repetition cells. Selected grouping requires 1.9--10.5\% more. These baselines use different random streams. Their 32-prompt subsets have no separate full deployment timers, so we compare support and MCSE without dividing the 128-prompt cost by four. Figure~\ref{tab:group-strong} shows the corresponding support and MCSE for the existing samplers.
\par
\subsubsection{Residual Grouping: Laws, Implementation, and Finite Checks}\label{app:group-proof}

\paragraph{Correctness under a Fixed Proposal Cap.}
Condition on the independent pilot, a root and its copy decisions, and a nonempty active block $B$. For $j\in B$, a single proposal satisfies
\begin{equation}\label{eq:appendix-group-acceptance}
 m_B(y)\frac{f_j(y)}{\sum_{k\in B}p_k(y)}=\frac{f_j(y)}{|B|},
 \qquad a_j=\Pr(j\text{ accepts})=\frac{\delta_j}{|B|}.
\end{equation}
\par

Zero proposal-density points have probability zero and require no acceptance ratio. Successive proposals and uniforms are independent, so the first proposal accepted by $j$ has law $\rho_j$. With $K$ shared proposals followed by an independent residual draw, the output probability of any measurable set $A$ is
\begin{equation}\label{eq:appendix-capped-residual-law}
 \sum_{t=1}^{K}(1-a_j)^{t-1}a_j\rho_j(A)
 +(1-a_j)^K\rho_j(A)=\rho_j(A).
\end{equation}
\par

Other members may finish earlier, but neither the proposal law of the fixed block nor the acceptance rule for $j$ changes. Shared uniforms affect dependence between members, not this marginal calculation. A singleton instead proposes from $p_j$ and accepts with probability $1-\min(1,p_0/p_j)$, which gives the same residual law. If $\delta_j=0$, that update copies almost surely and requests no residual.
\par

The conditional residual law is the same for every possible active block, root, and copy event. Integrating these conditions gives copied mass $\min(p_0(y),p_j(y))$ and noncopy joint density
$(p_0(x)-p_j(x))_+\rho_j(y)$.
Thus both the policy marginal and the entire base--update pair law are preserved. Independent namespaces across root draws preserve the distribution of each single-gain sample mean. The covariance between two gains depends on their residual coupling, so neither adjacent-contrast precision nor a nonlinear selector is fixed by this result. The same statements hold after applying any deterministic decoding and fixed scoring function to the token sequences.
\par

The implementation generates shared proposals in batches of eight and independent residual proposals in batches of sixteen. Accepted proposals are processed in their frozen order. Unused proposals still incur generation cost. Each prompt, update and generation repetition has an 8,192-proposal ceiling. Exhaustion returns an incomplete artifact rather than a substituted answer. All 18 reported banks completed with all 32 prompts. The fixed-$K$ fallback proof concerns the sampler.
\par

\paragraph{Pilot and Deterministic Implementation.}
For each training seed, the pilot uses 16 prespecified stratified selection prompts and eight independent responses per policy, with no score queries. At each prompt it evaluates all four policy densities on the 32 sampled answers, forms self-normalized mixture-importance weights and merges duplicate token sequences. Decoded-text collisions are also merged when computing CAFA support. Each of the five partitions is evaluated on 512 simulated banks of eight draws, using observed per-proposal generation, cross-density and weight-switch times. The surrogate includes batch overshoot and residual fallback costs. The frozen historical price estimate is approximately 0.001282 CNY per scoring input. Formal feedback charges use measured input and output token counts at the fixed 2/8 CNY per million-token tariff.
\par

The candidate partitions are
\begin{equation}\label{eq:appendix-candidate-partitions}
 \{1\}|\{2\}|\{3\},\quad
 \{1,2\}|\{3\},\quad
 \{1,3\}|\{2\},\quad
 \{1\}|\{2,3\},\quad
 \{1,2,3\}.
\end{equation}
All three pilots choose $\{1\}|\{2,3\}$ before formal validation. Pilot and validation prompts may overlap within the selection set, but their generated responses and random namespaces are independent. The validation matrix was fixed before its outcomes were available. No validation scores, sorting outcomes, or observed savings alter a partition.
\par

For completeness, a finite selection rule has a simple conditional stability property. If $|\widehat L(\mathcal P)-L(\mathcal P)|\le\epsilon$ simultaneously for all candidates, then
$L(\widehat{\mathcal P})\le\min_{\mathcal P}L(\mathcal P)+2\epsilon$.
This follows by adding the two approximation errors around the empirical minimizer. The pilot used here does not measure $\epsilon$.
\par

Root and copy streams are common across the three formal arms, while residual streams are arm-specific. Each policy export, pilot decision and protocol is identified by a hash. FP32 sequence log probabilities use the original token IDs, temperature one and the original EOS rules. Decoding only supplies the feedback identity. Independent reconstruction recomputes copy decisions, mixture indices, shared acceptance, fallback paths and the final response bank. The maximum observed probability-replay discrepancy is 0.001302 against a tolerance of 0.01.
\par

\paragraph{Exact Finite-Distribution Checks and an Attainable Improvement.}
The finite-law audit enumerates the joint distribution for all partitions with $K\in\{0,1,3,32\}$. It covers identical policies, disjoint supports, residual masses of order $10^{-8}$, unequal residual masses and singleton blocks. Marginals and base--update pair distributions are checked independently. Separate cases merge distinct token states into one decoded scoring input, and verify that a zero mean support can coexist with a nonempty variance support.
\par

An explicit three-state example shows why residual dependence can affect CAFA requirements even with all pair laws fixed. Let
$p_0=(0.5,0.25,0.25),\quad p_1=p_2=(0.3,0.35,0.35)$.
The independent-residual construction has expected joint query count $0.5$ per root draw. Pooling with the fixed-$32$ fallback gives
$0.4+0.1(0.9)^{32}=0.4034336838\ldots$.
With unlimited shared proposals the expectation is $0.4$. Linear programs over all couplings, and over couplings constrained to preserve the base--update pair laws, both give lower bound $0.4$ in this example. Merging the two residual states into one decoded identity also yields $0.4$ with finite fallback. These exact finite results establish available CAFA~improvement.
\par

\paragraph{What the Cost Records Identify.}
Formal generation charges sum all recorded attempt intervals. Interrupted attempts retain useful completed batches as well as unseparated loading, recovery and repeated work. Removing every failed interval would remove valid generation. Retaining only the last successful attempt would omit the saved prefix. The ledger therefore identifies the full observed execution cost, but does not identify a fresh uninterrupted deployment cost. The saved batch timers isolate generation and cross-density operations as a lower-bound diagnostic. They exclude parts of replay, loading and I/O.
\par

Pilot reuse is a separate deployment assumption. If one frozen partition remains applicable to $k$ comparable banks under the same policies, a descriptive amortization model is
\begin{equation}\label{eq:appendix-pilot-amortization}
 T_{\rm grouped}(k)=P+k(G_{\rm grouped}+F_{\rm grouped}),\qquad
 T_{\rm independent}(k)=k(G_{\rm independent}+F_{\rm independent}).
\end{equation}
\par

Using the recorded cell means gives about 1.9\% lower expenditure at $k=2$ and 4.6\% at $k=4$, versus 3.5\% higher expenditure for one bank. This assumes unchanged marginal throughput and prompt distribution. Reusing the pilot for a new checkpoint collection requires the same conditions to hold. The measured 32-prompt evidence supports a bounded mechanism improvement, not a lowest-total-cost deployment claim.
\par

The supplementary tables retain the per-bank decomposition behind those means. For each of the 18 banks they list the recorded generation charge including all attempts, the deployed pilot and the measured-token feedback charge, each kept separate, together with support, MCSE and proposal counts. Historical recovery overhead is not separately identifiable within the generation charge.
\par

\subsubsection{Residual-Group Acquisition, Precision, and Matched Baselines}\label{app:group-evidence}
All figures and companion numerical files retain the three training seeds and two generation repetitions separately. The 32 formal prompts belong to the selection partition. All 126 mean/variance support audits are exact against the fully scored frozen bank. Full per-cell covariance matrices and all seven target reports are retained in the machine-readable result bundle.
\par

\begin{figure}[!htb]\centering\normalsize
\captionsetup{skip=3pt}
\includegraphics[trim=44.73bp 11.97bp 5.5bp 18.25bp,clip,width=0.670\linewidth]{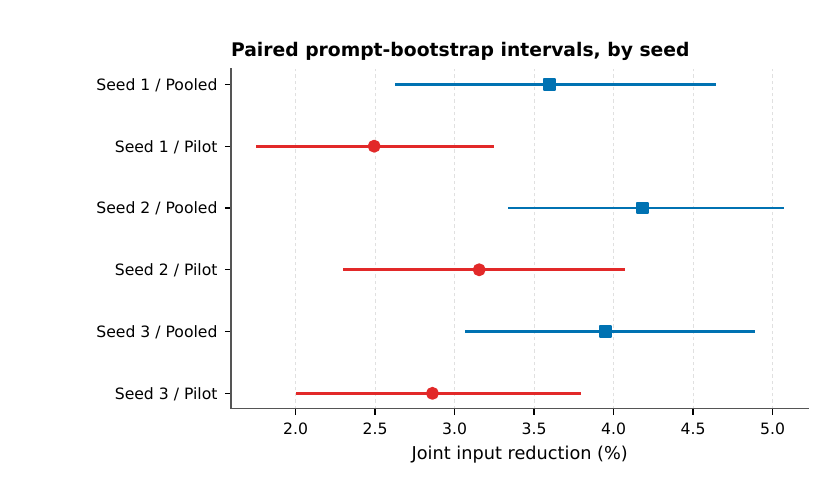}
\caption{Joint input reduction by training seed. Points and 95\% intervals use the original paired prompt bootstrap (10,000 resamples), averaging the two paired generation repetitions within prompt. Seeds are not pooled. These intervals describe CAFA increments on the frozen selection prompts.}\label{tab:group-bootstrap}
\end{figure}

\paragraph{Prompt Variability in the CAFA Increment.}
For each training seed, we average the two paired generation repetitions within prompt and bootstrap the 32 prompt clusters 10,000 times. The exploratory percentile intervals in Figure~\ref{tab:group-bootstrap} condition on the frozen pilot and model. Training seeds are not pooled. They quantify a small support improvement.
\par

\begin{figure}[!htb]\centering\normalsize
\captionsetup{skip=3pt}
\includegraphics[trim=6.78bp 6.41bp 5.5bp 4.75bp,clip,width=0.780\linewidth]{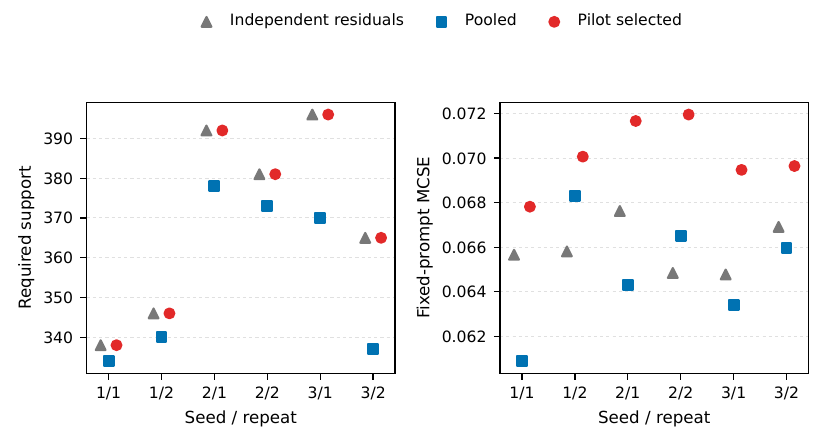}
\caption{Adjacent step 10--50: support and precision. All three residual-grouping arms are recomputed on the same 18 banks. Panels show the support needed for mean and covariance, and the fixed-prompt MCSE, for the step-10 to step-50 adjacent contrast. Side-by-side markers compare the three arms within each seed/repeat cell. Cells are not connected. Exact values are retained in CSV.}\label{tab:group-adjacent}
\end{figure}

\begin{figure}[!htbp]\centering\normalsize
\includegraphics[trim=6.78bp 6.42bp 5.5bp 4.75bp,clip,width=0.866\linewidth]{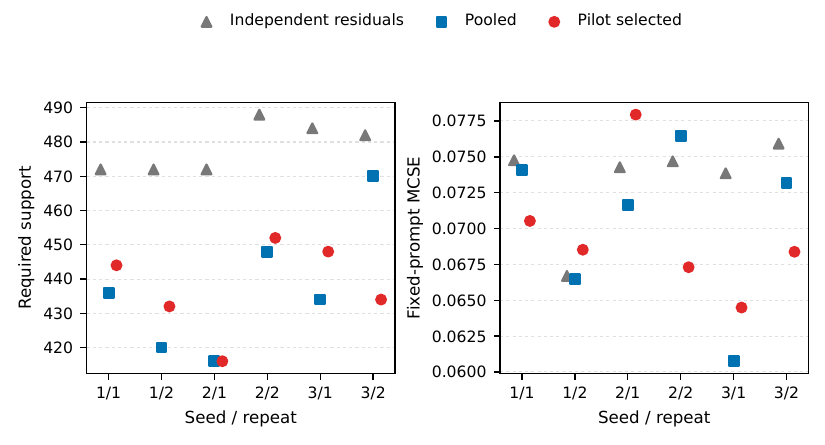}
\caption{Adjacent step 50--250: support and precision. Continuation of Figure~\ref{tab:group-adjacent}, with the same definitions and preserved per-cell values.}\label{tab:group-adjacent-panel-2}
\end{figure}

\paragraph{Single and Adjacent Targets.}
The scoring counts for each single base-relative target agree across all three arms in every matched cell. Across the six cells their means are 108.33, 369.50 and 473.50 for steps 10, 50 and 250, respectively. The adjacent targets in Figures~\ref{tab:group-adjacent} and~\ref{tab:group-adjacent-panel-2} can benefit from cross-update sharing even though the individual base-relative pair laws are unchanged. Their left panels compare required support and their right panels compare fixed-prompt MCSE within each seed and repetition. Adding the base absolute mean restores the full cached support within each arm. Different arms can still have different cached supports.
\par

\paragraph{Realized Gains across the Three Arms.}
The base-relative gains themselves, for every seed, generation repeat, arm and target with their fixed-prompt MCSE, are retained in the supplementary tables. They make one point: a different coupling can change the gain a bank realizes, while CAFA recovers each bank\textquotesingle s own statistics exactly. Changing the residual dependence is therefore a change of experiment, not a change of estimator.
\par
\newpage
\enlargethispage{\dimexpr-\baselineskip-\parskip\relax}

\subsection{Budgeted Acquisition}\label{app:H-budget}
When the budget is insufficient to score the full required support, we use a generalized-difference estimator with a paid pilot and a fitted score surrogate. Conditional on the bank and pilot, the estimator is unbiased for the fully scored bank contrast and has an exact query-induced MSE. This approximate extension is optional. We compare five acquisition policies on the strict-holdout Independent Star banks over a common budget grid, as detailed in Appendix~\ref{sec:budget-experiment}. Restricting acquisition to the non-zero support reduces error, while active allocation offers no consistent advantage over uniform allocation at the same budget. These errors are conditional on the frozen bank and do not measure language-model population error. Exact recovery requires acquiring the full necessary support.
\par
\newpage
\enlargethispage{\dimexpr-2\baselineskip-\parskip\relax}

\subsubsection{Acquisition under a Smaller Scoring Budget}\label{sec:budget-method}
Exact acquisition supplies the endpoint of a budget--accuracy curve: once all nonzero groups are scored, the realized comparison is known. Smaller budgets can use the same coefficient map. This connects contrast cancellation to active testing \citep{kossen2021testing} and surrogate-based evaluation \citep{kossen2022active}.
\par

Condition on a generated bank and a paid pilot $P$. Let $a_i\in\mathbb R^d$ be the normalized report-coefficient vector of unique input $i$, including its prompt weight, and let $\widetilde r_i$ be a score prediction fitted using only the pilot. Include each remaining nonzero input independently with probability $\pi_i>0$. With $I_i\sim\mathrm{Bernoulli}(\pi_i)$, return
\begin{equation}\label{eq:budgeted}
 \widehat\theta_B=\sum_i a_i\widetilde r_i+
 \sum_{i\in P}a_i(r_i-\widetilde r_i)+
 \sum_{i\notin P:a_i\ne0}\frac{I_i}{\pi_i}a_i(r_i-\widetilde r_i).
\end{equation}
\par

This is a generalized-difference estimator with classical inverse-probability correction. Its useful interface with cancellation is exact: zero groups receive no probability budget, and the full-support limit recovers the original comparison.
\par

Conditional on the bank and pilot,
\begin{align}
 \E[\widehat\theta_B]&=\theta_{\mathrm{bank}},\label{eq:budget-unbiased}\\
 \operatorname{Cov}(\widehat\theta_B)&=
 \sum_{i\notin P:a_i\ne0}(\pi_i^{-1}-1)
 a_i a_i^\top(r_i-\widetilde r_i)^2.\label{eq:budget-cov}
\end{align}
Taking expectations over an unbiased generation scheme preserves its target. The covariance in Eq.~\eqref{eq:budget-cov} is the additional uncertainty from purchasing only part of the frozen bank. Generation uncertainty is reported separately.
\par

\paragraph{Report Choice and Coordinate Changes.}
Corollary~\ref{cor:report} shows that an invertible reparameterization leaves the required support unchanged. Reducing that support requires changing the requested comparisons: one nearby update instead of a vector that also includes a distant one can change the support substantially.
\par

\subsubsection{Conditional Correctness and Acquisition Design}\label{app:budget-design}
Fix the generated bank, pilot observations, fitted surrogate and inclusion probabilities. In Eq.~\eqref{eq:budgeted}, $\E[I_i/\pi_i]=1$ recovers each remaining residual exactly in expectation. Independent inclusion makes cross-input covariance zero, and $\operatorname{Var}(I_i/\pi_i)=\pi_i^{-1}-1$, proving Eq.~\eqref{eq:budget-cov}. At $\pi_i=1$ for all required inputs, no query uncertainty remains. This reasoning applies to any pilot-measurable surrogate and any strictly positive probabilities on nonzero coefficients, including a poor predictor.
\par

Let $K$ be the required-support size. For $K\ge2$, each pilot samples $\min\{K-1,32,\max\{8,\lfloor0.05K\rfloor\}\}$ nonzero inputs uniformly without replacement, using twenty fixed seeds. All experimental supports exceed eight inputs. For $K=0$ the bank contrast is zero. For $K=1$ acquiring the sole required score gives the exact bank contrast. Its scores are charged to all policies. The surrogate uses an intercept, log response length, log question length, and newline, digit and punctuation fractions. Features are centered and scaled using the pilot, clipped to $[-8,8]$, and fitted by ridge regression with penalty 10 (intercept penalty $10^{-6}$). Predictions are clipped to $[1,10]$. A second ridge fit predicts the squared leave-one-out pilot residual, clipped to $[1,81]$. Active weights are $\|a_i\|_2$ times the square root of this predicted value. Inclusion probabilities are $\min(1,\lambda w_i)$, with $\lambda$ chosen for the stated expected budget. Uniform policies set $w_i=1$. The nonsurrogate policies use the fixed midpoint 5.5 in the generalized-difference estimator.
\par

The second-stage expected budget is 10, 25, 50, 75 or 100\% of the remaining required support. The cached-uniform comparator allocates that same expected count over its larger remaining cache. All support-aware comparators use exactly the same cancellation rule. The full-cache comparator buys every distinct input. Reported conditional MSE integrates the Bernoulli query randomness analytically. It is averaged over pilot splits and original frozen generation repetitions. Pilot splits are not additional model-generation or training seeds. Exact moment checks against Monte Carlo simulation accompany the implementation.
\par

\subsubsection{What Can a Smaller Scoring Budget Recover?}\label{sec:budget-experiment}
We hold every generated response fixed and compare five score-acquisition policies using Eq.~\eqref{eq:budgeted}: uniform acquisition over the cached bank with a midpoint or fitted surrogate, uniform acquisition over nonzero support with those same two predictions, and active acquisition over nonzero support with the fitted surrogate. All policies pay for the identical pilot, including its inputs in the displayed budget. Every surrogate policy uses the same features and fitted model, so the paired comparisons isolate pruning and allocation. The active policy uses only pilot labels and known coefficients to set its probabilities.
\par

The target here is the fully scored finite-bank report. Equation~\eqref{eq:budget-cov} gives its query-induced covariance. For squared Euclidean error, the conditional MSE is the trace of that covariance. Averaging squared error over the $d$ outputs divides this trace by $d$. These acquisition errors are distinct from error relative to the language-model population mean.
\par

Twenty pilot splits per frozen repetition probe allocation stability. They are splits of the paid pilot, not additional training or generation repetitions, and the four original generation repetitions remain the sampling replication unit. Every method is evaluated on the same fixed budget grid, including the exact full-cache endpoint. The budget is an expected count of unique queries under independent inclusion, rather than a fixed query count per run. The resulting error arises from randomized acquisition on the frozen bank, not from model generation.
\par

\begin{figure}[!t]
\captionsetup{skip=3pt}
\centering\includegraphics[trim=1.92bp 10.89bp 3.44bp 4.87bp,clip,width=0.775\linewidth]{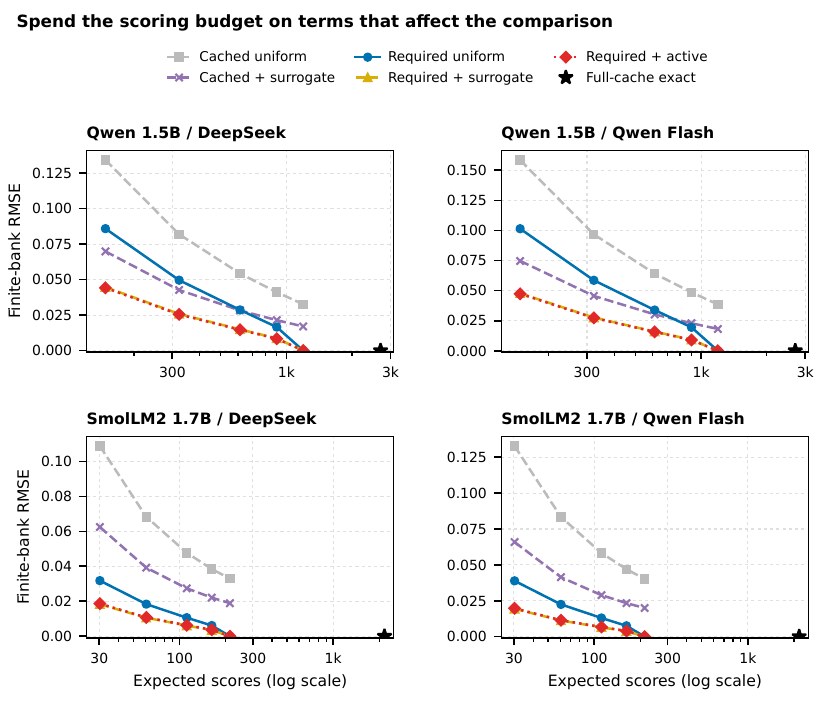}
\caption{Budget the scores that affect the comparison. Strict holdout. The same independent-star bank, 32 prompts, three updates and 64 root draws per prompt. Lines show the square root of mean conditional MSE over four generated banks and twenty paid-pilot splits per bank. Budgets are expected unique-query counts under independent inclusion. At full nonzero support, the three support-aware policies recover the bank contrast exactly. The star marks full-cache exact evaluation.}\label{fig:budget-allocation}
\end{figure}

At a second-stage expected budget equal to half the remaining required support, exact support pruning reduces finite-bank MSE by 72.2--95.0\% relative to uniform acquisition over the cache, with the same paid pilot and fitted score surrogate in both policies. The active-to-uniform-surrogate MSE ratios range from 0.96 to 1.07. Figure~\ref{fig:budget-allocation} reports both policies across the complete budget grid. These are conditional errors for the frozen response bank, integrated exactly over query randomization and averaged over four generation repetitions and twenty pilot splits per repetition.
Across the complete budget grid, the active-to-uniform-surrogate MSE ratio is 0.94--1.12.
\par

\section{Software and Reproducibility}\label{app:software}

\subsection{Installation and Minimal Workflows}\label{app:I-install}

The code archive contains an installable Python package (\texttt{contrast-eval} 0.3.0, Python 3.10 or newer) implementing Algorithm~\ref{alg:acquire}, a command-line interface, JSON schemas for banks, reports and score records, its test suite and its validation records. The exact core uses only the Python standard library and needs no GPU, no model weights and no network access. The package includes two CPU examples: a synthetic four-model bank evaluated with a single command (\texttt{contrast-eval demo}), and offline scoring of the mirror draws in Figure~\ref{fig:mirror-draws}, for which the mean requires no scores and the covariance requires both. A third example runs local generation, scoring and recovery end to end.
\par

The workflow validates the bank and plans the required support without reading scores or calling a scorer. It then acquires missing scores through a user-supplied scorer or an export--import workflow and returns the recovered report as JSON or CSV. A dry run returns the plan and makes no call. The following example illustrates the workflow using a user-supplied scorer callback, \texttt{my\_score(prompt,~response)}.
\par

\begin{center}\small

\begin{minipage}{0.94\linewidth}

\begin{verbatim}
bank   = load_bank("manifest.json", "draws.jsonl")
report = ReportSpec.gains(bank.model_ids, reference="base",
                          statistics="mean_covariance")
plan   = Evaluator().plan(bank, report)   # no score read
cache  = SQLiteScoreCache("s.db", scorer_version="rubric-v1")
result = Evaluator().compare(bank, report, my_score, cache)
\end{verbatim}

\end{minipage}

\end{center}

\subsection{Integrating Models, Banks and Scorers}\label{app:I-integration}

\paragraph{Banks.}

A bank is one set of policies, one generation repetition and one sampling configuration, with a fixed model order. Each record is a complete joint draw: one prompt, one draw index, and one response per model. Optional token IDs carry the probability identity. Incomplete joint draws, duplicated draws, prompts without draws and declared draw counts that disagree with the file are rejected rather than repaired, because each would change $n_i$ and the estimand. A prompt text that appears under two nominal prompt identities is rejected until the nominal groups are merged, so it is never silently treated as two independent coordinates, as detailed in Appendix~\ref{app:A-cov}. A bank also declares whether its draws and its prompts are independent. The Monte Carlo covariance of the mean and its standard error are reported only when both are declared. The mean and the per-prompt ordinary sample covariance are recovered in either case.
\par
\newpage

\paragraph{Reports.}

The mathematics allows any real report matrix $C$, as specified in Section~\ref{sec:setup}. The implementation represents $C$ exactly using integers or rational strings such as \texttt{"1/3"}, with columns ordered by the models in the bank. It supports contrast reports with zero-sum rows and general linear reports, including anchored reports. This is an interface choice that makes the zero test of a coefficient vector reliable: floating-point coefficients are refused rather than compared with a tolerance, because treating a small non-zero coefficient as zero would drop a score the report needs.
\par

\paragraph{Scorers and Caches.}

A scorer receives a prompt and a response and nothing else. Cached scores can be reused only with the same scorer version, defined by the rubric, judge deployment and settings. A changed fingerprint or a conflicting score for the same identity raises an error. Conflicting values are not averaged. Refusals, timeouts and parse errors are stored as failures without a value. After a timeout or an unexplained scorer error, the identity remains claimed and marked as pending. A rerun does not issue another paid request until the provider outcome has been reconciled. A required score that remains unavailable either blocks the exact report or, if a score range is declared, yields per-output bounds with no covariance point estimate.
\par

\paragraph{Generation.}

Imported banks from any sampler are supported. For local causal language models the archive implements two generators, the Independent Star and independent autoregressive sampling. Both keep the original token IDs, and both evaluate log-probabilities of given token sequences under each policy with the same temperature, top-$k$/top-$p$, repetition penalty, end-of-sequence rule and length cap used for sampling. They draw from named random streams. Giving all models the same seed is not a substitute. The packaged Independent Star implements the copy and residual acceptance rules of the sampler used in the source study. When a residual reaches its proposal ceiling, the draw is returned as an incomplete prefix that can be resumed, never replaced by an independent sample. The port does not reproduce the historical random streams from the source study bit for bit and therefore generates new banks rather than replicas of the reported banks.
\par

The adapters were checked on one GPU with the online-GRPO base model and its step-10 and step-250 LoRA checkpoints, using two prompts and two joint draws per prompt. Repeated generation reproduced the same tokens. Teacher-forced log-probabilities of the original sequences agreed with those computed by the evaluator used in the source study within an absolute tolerance of 0.01, including when temperature, top-$k$/top-$p$ and repetition penalty were changed. An interrupted residual draw resumed to the same result as an uninterrupted run, and statistics recovered from the generated banks equalled full scoring. This is a functional check. The TV tree, shared-token AR and residual grouping have no packaged generator. Banks they produced can be imported. The reported experimental results do not depend on these adapters, because every reported bank was generated by the recorded implementation used in the source study, as detailed in Appendix~\ref{app:B-samplers}.
\par

\subsection{Correctness Checks}\label{app:I-tests}

Six checks of different objects are kept separate.
\par

\begin{table}[!htb]\centering
\begingroup\footnotesize\setlength{\tabcolsep}{3pt}
\AddToHookNext{env/tabular*/begin}{\renewcommand{\arraystretch}{1.08}}

\begin{tabular*}{\linewidth}{@{\hspace{\tabcolsep}\extracolsep{\fill}}>{\raggedright\arraybackslash}m{0.12\linewidth}>{\raggedright\arraybackslash}m{0.33\linewidth}>{\raggedright\arraybackslash}m{0.36\linewidth}>{\raggedright\arraybackslash}m{0.129\linewidth}}\toprule

Check & Coverage & Result & Evidence\\\midrule

Independent recomputation of the published cells & All 504 online-GRPO report cells from the 72 frozen banks and the frozen cache, by a separate implementation of Algorithm~\ref{alg:acquire} & Bank hashes, support counts, scorer-call counts, means and covariances all match & Appendix~\ref{app:D-preservation}\\\midrule

Released-package regression & The same 504 cells, plus the 126 residual-grouping reports (18 banks $\times$ 7 targets) and their 18 joint reports, evaluated through the plan--acquire--recover path implemented in the package & Bank hash, cached, mean and mean-plus-covariance counts and one score access per required identity match in every cell, means and covariances agree within $10^{-12}$ with both the published cell and full scoring, largest observed error $4.3\times10^{-19}$ & Archive validation records\\\midrule

Package test suite & 81 unit and integration tests & All pass with the optional JSON-schema validator installed, without it three schema-shape tests are \emph{skipped}, not passed & Archive validation records\\\midrule

Acceptance fixtures & Twelve synthetic arithmetic cases: all-copy banks, a unique update, an inactive column, mirror draws, a repeated pair, an absolute anchor, a zero-contribution draw, exact rational cancellation, a tiny non-zero coefficient, an all-zero report, a single draw and a three-draw case & Each checked by an independent arithmetic routine against its expected support, mean and ordinary covariance & Archive validation records\\\midrule

Main-text numbers & The 32 audited claims and every prose number & 26 reproduce exactly as stated, four under a scope the text now states explicitly, and two simulation-design statements were matched to frozen records without re-running the simulations, none contradicted & Supplementary numeric audit\\\midrule

Source-study checks & The finite enumerations of Appendix~\ref{app:A-checks} and the dense audits of the original analysis & Reported as run in the source study, \emph{not} re-run here & Appendix~\ref{app:A-checks}\\

\bottomrule\end{tabular*}

\endgroup
\end{table}

The first two rows describe checks of the same cells using separate implementations: one verifies the reported numerical results, and the other validates the released package. Both checks were rerun on a CPU using the archived source for this paper. The validation records in the archive separately list passed, failed and skipped cases and document the environment for each run.
\par
\newpage

\subsection{Artifact Map and Regeneration}\label{app:I-artifacts}

\paragraph{Frozen Analysis Snapshot.}

Analysis uses a frozen database snapshot created through SQLite's backup API in a read transaction. It records ledger, generation and protocol hashes, accepted generation audits, judge identities and the sampling grid. Validation rejects changed hashes, incomplete generations or four-shard repetitions, duplicate draws and missing root indices. New and imported scores must match their recorded judge identities, including fingerprints where available. Historical versions remain distinct.
\par

\paragraph{Artifact Map.}

The supplementary archive links each reported number to its source record. Canonical tables preserve frozen outputs, verified natural keys and source-record indices. Keys distinguish model family, bank, output cap, sampler, budget, repetition, training seed, target and judge. Allocation analyses also include acquisition method and budget fraction. Separate provenance fields record checkpoint regime, bank hash where available, prompt set, score version, statistic and cost convention. Explicit sampler-name mappings and key checks prevent incompatible records from being merged.
\par

A claims table and numeric audit record recomputed main-text values, scopes and source paths. One command regenerates all empirical figures and Tables~\ref{tab:workflow}, \ref{tab:targets} and~\ref{tab:controlled}. It also writes a source-linked manifest of plotted and tabulated values and logs every CNY-to-USD conversion, as detailed in Appendix~\ref{app:G-currency}. Editable draw.io sources and verified exports accompany the method diagrams. Regeneration copies those exports unchanged. Complete CSV files, a table-to-figure map and input hashes preserve the supporting records. Regenerating reported support, precision and cost values requires no GPU computation or judge calls.
\stopcontents[appendix]
\par
\end{document}